\documentclass[11pt]{article}
\usepackage[margin=1in]{geometry}
\usepackage{amsmath,amssymb,amsthm}
\usepackage[T1]{fontenc}
\usepackage{lmodern}
\usepackage{graphicx}
\graphicspath{{plots/}}
\usepackage{booktabs,longtable,array,multirow}
\usepackage{etoolbox}
\usepackage{siunitx}
\usepackage{enumitem}
\usepackage{caption,subcaption}
\usepackage{microtype}
\usepackage{xcolor}
\usepackage{url}
\usepackage{csquotes}
\usepackage{hyperref}
\usepackage[nameinlink]{cleveref}
\hypersetup{
  colorlinks=true,
  linkcolor=blue,
  citecolor=blue,
  urlcolor=blue
}
\crefname{equation}{Eq.}{Eqs.}
\Crefname{equation}{Equation}{Equations}
\newcommand{\EqCnumber}{\ifnum\value{equation}<10 0\fi\arabic{equation}}
\newlength{\origtabcolsep}
\AtBeginEnvironment{longtable}{\footnotesize\setlength{\tabcolsep}{4pt}}
\AtEndEnvironment{longtable}{\setlength{\tabcolsep}{\origtabcolsep}}
\AtBeginEnvironment{displayquote}{\small}
\title{ENIGMA: The Geometry of Reasoning and Alignment in Large-Language Models}
\author{Gareth Seneque \\ Lap-Hang Ho \\ Nafise Erfanian Saeedi \\ Jeffrey Molendijk \\ Ariel Kuperman \\ Tim Elson \\[0.75em] \textit{Australian Broadcasting Corporation}}
\date{October 13, 2025}

\begin{document}
\maketitle
\begin{displayquote}
\itshape These facts are, like all facts, not necessary but of a merely empirical certainty; they are hypotheses; one may therefore inquire into their probability.\par\medskip
\hfill --- Bernard Riemann (1854)
\end{displayquote}

\begin{displayquote}
\itshape Geometry is the science of correct reasoning on incorrect figures.\par\medskip
\hfill --- George P\'olya, \emph{How to Solve It} (1945)
\end{displayquote}

\begin{abstract}
\noindent We present \emph{Entropic Mutual-Information Geometry Large-Language Model
Alignment} (ENIGMA), a novel approach to Large-Language Model (LLM)
training that jointly improves reasoning, alignment and robustness by
treating an organisation\textquotesingle s policies/principles as directions to move on a
model\textquotesingle s information manifold. Our single-loop trainer combines
\emph{Group-Relative Policy Optimisation} (GRPO), an on-policy, critic-free RL
method with Chain-of-Thought (CoT)-format only rewards; a \emph{Self-Supervised
Alignment with Mutual Information} (SAMI)-style symmetric InfoNCE auxiliary;
and an entropic Sinkhorn optimal-transport regulariser on hidden-state
distributions to bound geometry drift. We also introduce infoNCE metrics that
specialise to a standard MI lower bound under matched negatives to measure how
strongly a model\textquotesingle s CoT encodes these policies. These metrics include a
Sufficiency Index (SI) that enables the selection and creation of principles
that maximise downstream performance prior to training. In our experiments
using small (1B) LLMs, high-SI principles predict steadier training dynamics
and improved benchmark performance over GRPO ablations. Our
information-geometry analysis of trained models validates desirable structural
change in the manifold. These results support our hypothesis that reasoning,
alignment, and robustness are projections of a single information-geometric
objective, and that models trained using ENIGMA demonstrate \emph{principled
reasoning} without the use of a reward model, offering a path to trusted
capability.
\end{abstract}
\clearpage
\tableofcontents
\newpage
\section{Introduction}\label{introduction}

\emph{Authors' note: no editorial position is offered or claimed in this
publication. All LLM prompts and synthetic data examples are provided to
support replication of our work and do not reflect optimisation for
production use or reflect an editorial position.}

\subsection{Overview}\label{overview}

In our prior work on reasoning \& alignment, \emph{ABC Align}~\cite{seneque2024} we developed a post-training framework suitable for organisations who aim to leverage the capabilities of LLMs while making use of their data and preserving provider independence, proxies for organisational obligations to innovation and independence. To do this, we leveraged open research and techniques including Orca~\cite{mukherjee2023}, Less Is More for Alignment (LIMA)~\cite{zhou2023}, and Odds Ratio Preference Optimization (ORPO)~\cite{hong2024} to make use of proprietary data (synthetic reasoning traces constructed from ABC content, conditioned on our AI Principles) to elicit improved performance on both alignment and reasoning benchmarks, avoiding the so-called alignment tax. These methods were effective but fundamentally post-hoc: they relied on in-context prompts or preference-style fine-tuning and offered limited control over how principles reshape the model's internal geometry.

ENIGMA departs from that setting. We treat written principles as \emph{directions of motion on a model's information manifold}, and we design both the training signal and the measurements to act directly on that geometry. In doing so we aim to elicit \emph{principled reasoning}: chains of thought that encode stated principles, without an external reward model and with explicit, reportable evidence that those principles are encoded.

Recent progress in ``reasoning LLMs'' largely couples on-policy RL with chain-of-thought (CoT) sampling, yielding strong gains where answer correctness can be verified (math, code). In subjective domains, where correctness encompasses normative definitions of truthfulness, alignment to preferences etc., correct-answer supervision is costly and fragile, and even visible reasoning traces can be unfaithful. Industry standard tooling also lacks the built-in capabilities to measure interventions that target information geometry objectives.

To address these limitations, ENIGMA starts from the premise that \emph{neural networks are geometric entities}.

\subsection{ENIGMA}\label{enigma}

In this paper, we introduce a single-loop training method that adapts Group-Relative Policy Optimisation (GRPO)~\cite{fr2024}, Self-Supervised Alignment with Mutual Information (SAMI)~\cite{bai2022}, and Sinkhorn divergence to implement Optimal Transport (OT)~\cite{ambrosio2021,peyr2019} and develop quantitative measures to evaluate constitutional principles both prior to training and their impact on training itself. Our results
provide evidence that we elicit \emph{principled reasoning} from LLMs using
open and proprietary data without the cost and complexity of human
preference data collection or correctness-verified question/answer
datasets for general domains. Our use of `principled' reflects that our
metrics provide lower-bound evidence that completions encode the
constitutional principles. For our experiments, we have used the ABC's
Editorial Policies \cite{australian2025} as `positive' constitutions, a source of
human-written truth, and two sets of LLM-generated `negative'
constitutions. We have developed a method of quantifying the
constitution's impacts on training dynamics that predicts downstream
task performance. This method produces a novel Sufficiency Index (SI)
that is generally applicable to any set of paired constitutions.

Our intervention is thus fundamental, going beyond standard preference
optimisation techniques or alignment methods that involve exploiting the
properties of In-Context Learning (ICL) via natural-language `prompts'
without updating the model's parameters, These methods are effective for
a variety of use-cases (outlined in our work on \emph{ABC Align}~\cite{schulman2015}, e.g.,
generating content metadata or multi-turn conversation), but are
ultimately brittle and amenable to input distribution perturbation (see
evaluation discussions in recent surveys/best-practice notes \cite{biderman2024}).

We necessarily limit the scope of our experiments to small LLMs given
the setting in which this work was conducted, that of a public-service
media organisation (see our prior work on \emph{ABC Align} for additional
organisational context \cite{schulman2015} and our `Limitations' section for
additional detail). Our methods and results however demonstrate that
ENIGMA ultimately enables organisational collaboration on defining
principles and standards for any use-case where a high degree of
alignment is required, and provides quantitative measures that relate
those principles to model behaviour and outputs.

\subsection{Key contributions}\label{key-contributions}

We make several contributions that connect Constitutional AI evaluation,
verification-free RL, and MI/OT-based LLM alignment techniques.

\begin{itemize}
\item
  \textbf{Unified information-geometric training objective (ENIGMA).}\\
  We formulate \emph{reasoning, alignment,} and \emph{adversarial
  robustness} as a single optimization problem on the product space of
  distributions over next-token logits and sequence-level hidden
  representations by coupling on-policy GRPO \cite{fr2024} with
  Constitutional-principle shaping \cite{bai2022} and a Sinkhorn divergence
  penalty \cite{ambrosio2021,peyr2019}, all in one loop. This bridges online RL with
  principle-guided supervision and geometry-aware regularisation. We see
  benchmark improvements for GPQA (main) +6.92pts/50.81\%; and
  TruthfulQA \cite{melnyk2024,chen2025} +12.11pts/31.81\% (absolute/relative scores)
\item
  \textbf{Reportable MI lower bound for ``principle encoding.''}\\
  We introduce row/column `clean' InfoNCE metrics \cite{hao2024}, a
  computationally light alternative to the symmetric InfoNCE of SAMI
  \cite{bai2022}. The metric is `clean' as it is calculated without entropy
  gating and with uniform shadow principles. It yields a per-step lower
  bound on how strongly completions encode a targeted principle. This
  gives a falsifiable, quantised proxy for ``constitutional adherence''
  that can be tracked like loss/accuracy (note: we address known
  bias/variance pathologies at high MI by design of the metric.)
\item
  \textbf{Principle curation by \emph{effective MI}.}\\
  We propose quantitative sufficiency signals and a composite metric, a
  Sufficiency Index (SI) to \emph{rewrite/select} constitutional
  principles, producing a set that empirically yields better training
  dynamics and downstream task performance. Constitutional AI
  established principle-guided training \cite{li2018}; we add principle
  selection via MI diagnostics \cite{hao2024,van2018}.
\item
  \textbf{Interpretability-friendly geometry probes.}\\
  We track Bhattacharyya angle, Hellinger, and JS divergence (metrics
  background in \cite{bhattacharyya1943,lin1991,amari2000}; \Cref{eq:bhattacharyya,eq:hellinger,eq:js-bits}) between last-token distributions and
  representation-level probes (Fréchet distance \cite{heusel2017}; \Cref{eq:frechet}, effective
  rank/participation ratio \cite{roy2007,recanatesi2022}; \Cref{eq:erank-pr}) to connect manifold movement
  with task gains.
\item
  \textbf{CoT-only \& verifier-free reward `tie-breaker' with GRPO.}\\
  We show that format-only rewards (XML tags for reasoning/answer) plus
  MI-based tie-breaking suffice to move reasoning quality and
  calibration, using on-policy GRPO \cite{fr2024} to avoid stale off-policy
  artifacts without the use of an external reward model. Our tie-breaker
  reward makes principle encoding actionable for GRPO without
  overwhelming task reward and allows scaling of our methods to larger
  models where format compliance is high from pre- and post-training.
\item
  \textbf{Principled, \emph{parameter-efficient} RL post-training.}\\
  As in our prior work, we leverage LoRA for memory efficiency, while
  enabling frequent MI/OT/geometry measurements. This approach is both
  cost- and compute-efficient.
\end{itemize}

\section{Related Work}\label{related-work}

This section situates ENIGMA relative to three threads in post-training
for LLMs. First, policy-gradient alignment with trust-region control:
TRPO \cite{feydy2019} and PPO \cite{wei2022} interpret the KL-to-reference constraint
as a local step-size control that approximates natural-gradient motion
on the Fisher--Rao manifold of next-token distributions; GRPO replaces a
trained critic with a group baseline well-suited to multi-sample CoT
pipelines \cite{fr2024}.\footnote{Not to be confused with Group Robust
  Preference Optimisation, an unrelated ``GRPO'' acronym in preference
  optimization \cite{poole2019})}

Second, mutual-information--based alignment: contrastive learning
(InfoNCE) provides a variational lower bound on mutual information
\cite{hao2024}, and recent work (SAMI) uses conditional MI $I(Y;C\mid X)$ to teach
models to follow written principles (``constitutions'') without
preference labels or demonstrations \cite{bai2022}.

Third, optimal-transport regularization: entropic OT (Sinkhorn) yields a
smooth, scalable divergence between empirical measures with
well-understood convexity and convergence-in-law properties \cite{ambrosio2021,peyr2019}. Recent alignment work casts distributional preference alignment
as an OT problem (AOT) that enforces relaxed first-order stochastic
dominance over reward distributions \cite{korbak2025}.

We also cover complementary studies on process supervision (PRMs/CoT)
and constitution selection, then gather the threads in a way that points
to the theoretical basis of our work, outlined in the following section.

Our aim here is not an exhaustive survey but to surface design choices
that motivated ENIGMA's single-loop combination of GRPO (policy control)
\cite{fr2024}, SAMI (principled process alignment in representation space)
\cite{bai2022}, and a lightweight Sinkhorn penalty on hidden states
(distributional drift control) \cite{ambrosio2021,peyr2019}.

\subsection{\texorpdfstring{Group Relative Policy Optimization (GRPO)
}{Group Relative Policy Optimization (GRPO)}}\label{group-relative-policy-optimization-grpo}

GRPO, introduced in \emph{DeepSeekMath} \cite{fr2024}\emph{,} replaces a
learned value critic with a group baseline computed from multiple
sampled completions per prompt, while retaining a KL-to-reference trust
region (PPO-style). This removes the critic's memory/compute overhead
and matches grouped sampling used in reasoning pipelines.

KL-constrained policy optimization (TRPO/PPO) can be interpreted as
approximate natural-gradient steps on the Fisher information manifold
\cite{feydy2019,wei2022}; the trust region respects local distributional curvature.
Related work defines Wasserstein natural gradients by pulling back the
W2 metric to parameter space, motivating links between KL-style and
OT-style geometries \cite{lightman2023,chen20252}. For ENIGMA, we utilise PPO-style ratio
clipping (DR-GRPO) to provide local step-control.\footnote{Note:
  ``GRPO'' is also used for Group Robust Preference Optimization in
  reward-free RLHF; to avoid confusion, we use GRPO (policy) for
  DeepSeekMath's algorithm \cite{fr2024} and GRPO (group-robust) for the
  unrelated worst-group PO method \cite{ramesh2024}.}

\subsection{Self-Supervised Alignment with Mutual Information
(SAMI)}\label{selfsupervised-alignment-with-mutual-information-sami}

SAMI fine-tunes an LLM to maximize conditional mutual information
$I(Y;C\mid X)$ between responses Y and a constitution C given prompts X,
requiring neither preference labels nor demonstrations \cite{bai2022}. InfoNCE
lower bounds MI and supplies a practical contrastive objective for such
alignment \cite{hao2024}; see also analyses and caveats around MI maximization
\cite{van2018}. SAMI complements Constitutional AI (CAI), which uses a written
constitution to produce AI feedback and then runs supervised/RL phases
\cite{li2018}.

\subsection{Optimal Transport (OT)
regularisation}\label{optimal-transport-ot-regularisation}

Optimal Transport (OT) offers a principled way to compare probability
distributions by computing the minimal ``work'' to morph one into
another. The resulting 2-Wasserstein metric equips the space of
probability measures with a geometry that behaves Riemannian in the
sense of Otto's calculus [23], so distributions can be connected by
geodesics and even interpreted through gradient-flow dynamics. In
practice, entropic regularisation leads to the Sinkhorn formulation,
which is an efficient, differentiable approximation whose
smoothness/positivity properties make it well-suited to modern
deep-learning pipelines \cite{ambrosio2021,peyr2019}. Beyond serving as a geometry-aware
distance, OT has become a useful regulariser for aligning model
behaviours. AOT casts distributional preference alignment as a 1-D OT
problem that enforces (relaxed) first-order stochastic dominance between
reward distributions and reports strong results among 7B models
\cite{melnyk2024}. In a vision-language adaptation, Prompt-OT uses an OT penalty
to preserve feature-distribution structure during prompt tuning
\cite{melnyk2024}. In this work we adopt the same spirit: a small entropic-OT
term keeps representations within a shared W2 ball, limiting drift while
leaving GRPO's low-variance updates and SAMI's principle-consistent
directions intact \cite{ambrosio2021,peyr2019,korbak2025}.

\subsection{\texorpdfstring{Chain of Thought, Process Supervision and
Monitoring
}{Chain of Thought, Process Supervision and Monitoring}}\label{chain-of-thought-process-supervision-and-monitoring}

While Chain-of-Thought (CoT) prompting improves measured reasoning on
benchmarks, its safety value depends on whether intermediate traces are
\emph{faithful} to the model's computations. Process supervision and
process-reward models (PRMs) directly score intermediate steps and have
been shown to outperform outcome-only supervision on math reasoning and
to enable step-level audits (e.g., PRM800K; Let's Verify Step by Step)
\cite{schulman2015}. Recent work finds that reasoning models may rationalise or
hide internal computations \cite{martens2020}, underscoring the need for monitors
that evaluate process rather than only outcomes (see also work on CoT
monitorability \cite{davidson2018}). Additionally, there is a growing literature on
`reasoning in latent space' \cite{shao2024}, which may further limit
interpretability and monitoring efforts. ENIGMA's CoT-format--only
rewards and MI reward tiebreaker, coupled with SAMI-based shaping and
loss, provide a reference- and verifier-free means of process
supervision, where the information contained in a set of principles is
encoded into the CoT process itself, which is produced as part of the
model's completion.

\subsection{Constitution Evaluation and Principled
Reasoning}\label{constitution-evaluation-and-principled-reasoning}

Existing approaches to deploying and adapting Constitutional AI (CAI)
utilise human-written and synthetic principles but offer few
quantitative tools for selecting or curating them toward a target metric
that guides their suitability for the training process itself \cite{li2018}.
We address this gap by defining \textbf{SI}. For a given set of
positive/negative constitutions or principles, we measure predictive
information via $\Delta$NLL, associative information via InfoNCE-style MI
bounds \cite{hao2024,van2018}, and separation ($\Delta$NLL AUC). We also track
information-geometry observables (Bhattacharyya angle, Hellinger,
effective rank) \cite{davidson2018,peyr2018}. to diagnose manifold changes and relate
them to changes in task performance.

\subsection{\texorpdfstring{Towards Operational Guarantees for
Large-Language Model Alignment
}{Towards Operational Guarantees for Large-Language Model Alignment}}\label{towards-operational-guarantees-for-large-language-model-alignment}

As LLMs transition from chat assistants or data processing tools to
instruments that facilitate the automation of decision-making,
international AI policy and standards guidance emphasises documented
risk management, adversarial testing, and transparency \cite{national2025,national2024,ai2025}.
ENIGMA supplies two complementary measures towards guarantees of safe,
reliable capability: principle encoding via the MI lower bound discussed
above (subject to known MI bound trade-offs \cite{van2018}) and distributional
alignment via entropic Sinkhorn OT that aligns preference/reward
distributions rather than isolated samples, an approach recently shown
to enforce stochastic-dominance-style constraints in LLM alignment
\cite{korbak2025}.

Our implementation includes configuration parameters to stabilise this
optimisation with Fisher-metric preconditioning (natural-gradient) and
entropy control, standard tools that reduce mode collapse and training
pathologies in policy-gradient style fine-tuning. We did not observe
gradient conflicts during training, however for different settings and
model capacities, the literature on multi-task learning offers solutions
to this problem \cite{chen2017}.

\subsection{An information-geometric view of LLM
post-training}\label{an-information-geometric-view-of-llm-post-training}

The threads above admit a common geometric reading that will structure
the next section. Policy updates live on the Fisher--Rao manifold of
categorical token distributions; TRPO/PPO give a
trust-region/natural-gradient view \cite{feydy2019,wei2022}. Contrastive MI acts on
$\ell_2$ -normalised representations; InfoNCE tightens a lower bound on
$I(Y;C\mid X)$ \cite{hao2024} through the row/column objectives in
\Cref{eq:row-nce,eq:col-nce,eq:sami-aux}, moving paired (response, principle)
embeddings along spherical geodesics (cf. hyperspherical geometry
perspectives \cite{ambrosio2008}). Sinkhorn OT treats per-sequence hidden states
as empirical measures and penalizes coherent mass transport away from a
reference via \Cref{eq:w2,eq:entropic-ot,eq:sinkhorn-div,eq:rot}.

We do not assume a formal product-manifold geodesic. Instead, we use
this trio as a composite control system: local policy steps
(Fisher--Rao) + semantic alignment in representation space
(spherical/InfoNCE) + global drift control over hidden-state
distributions (Wasserstein/Sinkhorn) [15, 16, 19, 23].\footnote{See
  also Amari's information-geometry framing \cite{nielsen2022}.}

\section{Information Geometry and
Alignment}\label{information-geometry-and-alignment}

The Fisher--Rao view (TRPO/PPO/GRPO) endows token distributions with a
local Riemannian structure; ratio-clipping (and KL, which we do not
employ here) restricts step size in that geometry. The contrastive view
(InfoNCE) acts on normalised representations, pulling completions toward
their governing principles and away from negatives, thereby injecting a
positive principle-aligned semantic direction, useful for domains where
correctness is hard to verify. The Wasserstein view (entropic Sinkhorn)
treats batches of hidden states as empirical measures and penalizes mass
transport away from the reference, serving as a global guardrail against
drift that appears small in KL but semantically large in representation
space.

\subsection{Policy space as a Fisher--Rao
manifold}\label{policy-space-as-a-fisherrao-manifold}

Consistent with existing literature, the family of categorical token
distributions forms a statistical manifold with the Fisher information
as its Riemannian metric; locally, the KL divergence between nearby
policies equals ½ of the squared Fisher--Rao distance up to higher-order
terms \cite{amari2000,miyamoto2023}. This underpins natural-gradient methods'
reparameterisation invariance and trust-region interpretations
(\Cref{eq:natgrad}).

TRPO directly constrains a per-update KL trust region (\Cref{eq:trpo}), yielding
monotonic-improvement guarantees under its approximations; PPO \cite{schulman2017} enforces
an approximate trust region through clipped ratios and/or KL penalties
via the surrogate in \Cref{eq:ppo-max}; GRPO removes the critic and uses a
group baseline for advantages (\Cref{eq:grpo-adv}) while keeping the same
KL-to-reference control and is well-suited to CoT-style grouped sampling.
As noted, we do not use KL for our
implementation\footnote{Though we conducted many experiments with both
  KL and OT regularisation inside the same training loop, including an
  active KL controller, we observed a negative impact on training
  stability.}, in line with recent GRPO best-practices instead we use
PPO-style ratio clipping (DR-GRPO) to provide local step-control and
instead employ Sinkhorn divergence for regularisation.

\subsection{Contrastive alignment on a
(hyper)sphere}\label{contrastive-alignment-on-a-hypersphere}

When embeddings are $\ell_2$ -normalised, they live on a unit hypersphere;
contrastive objectives such as InfoNCE maximize a lower bound on mutual
information by increasing the similarity of matched pairs relative to
negatives \cite{hao2024} via the objectives in
\Cref{eq:row-nce,eq:col-nce}. Hyperspherical models and geodesic perspectives
motivate our geometric reading \cite{ambrosio2008}.

SAMI operationalises this view at the principle level: it maximizes the
conditional MI $I(Y;C\mid X)$ between responses and a constitution-given
prompt, without preference labels. In our experiments, that constitution
foregrounds first-principles reasoning and factuality, expressed through
the editorial standards of a PSM, biasing the model toward provable
aligned chains-of-thought even without answer-correctness verification.

Because $I(Y;C\mid X)$ equals an expected KL between the joint $p(y,c\mid x)$ and
its factorisation $p(y\mid x)p(c\mid x)$, the SAMI loss/reward shaper in
\Cref{eq:sami-aux,eq:mi-reward} shortens
geodesics on the statistical manifold in directions predictable from the
constitution, a semantic directional constraint that substitutes for the
KL step-size constraint on the policy \cite{chen2017}. This is the theoretical
lens which, combined with recent GRPO best-practices and observations
from our early experiments, motivates our use of OT instead of KL
regularisation.

\subsection{Sinkhorn OT and the Wasserstein geometry of hidden-state
measures}\label{sinkhorn-ot-and-the-wasserstein-geometry-of-hiddenstate-measures}

Sequences of hidden states can be treated as empirical measures. The
squared 2-Wasserstein distance in \Cref{eq:w2} measures geodesic distance in the space
of probability measures (the Otto metric). We adopt Sinkhorn
divergences, entropically-regularised OT with debiasing, which are
positive, convex, and metrise convergence in law while remaining smooth
and scalable for backprop \cite{ambrosio2021,peyr2019}; \Cref{eq:entropic-ot,eq:sinkhorn-div}
capture these quantities.

ENIGMA computes a Sinkhorn divergence between empirical measures of
hidden states from the current policy and a frozen reference, penalizing
shifts that look small in KL but correspond to coherent mass transport
(e.g., moving probability mass between modes), acting as a
geometry-aware tether \cite{huh20242} through the regulariser in
\Cref{eq:rot}.

\subsection{A product-manifold-inspired trust
region}\label{a-productmanifoldinspired-trust-region}

For each ENIGMA training step, GRPO supplies a local KL trust region for
CoT-only rewards, MI supplies a semantic direction consistent with a
first-principles/factuality constitution, and Sinkhorn OT provides a
global geometry-aware constraint (via regularisation). While we use
``product-manifold'' language for intuition, this is an algorithmic
composite rather than a formal geodesic in \Cref{eq:enigma-loss}; guarantees are
inherited from the constituent terms (e.g., PPO-style ratio clipping
(DR-GRPO) for the local step-control, convexity properties of the
Sinkhorn divergence), not from an intersection of convex geodesic balls.

\subsection{The Information Geometry of
ENIGMA}\label{the-information-geometry-of-enigma}

The information-geometric view outlined above motivates the combination
of training and regularisation methods used in ENIGMA. We couple policy
updates (Fisher--Rao), MI-driven semantic alignment on normalised
representations (the `contrastive learning on a hypersphere' lens), and
OT-based distributional control (Wasserstein metric mapping across
distributions). This geometry-aware composition is especially useful in
our CoT-only/MI tie-breaker reward regime: MI with a first-principles
constitution supplies the missing process supervision, ensuring that
\emph{principled reasoning} augments answer correctness, OT prevents
degenerate drifts that CoT-format rewards can otherwise induce.

\subsection{Towards universal
representations}\label{towards-universal-representations}

The Platonic Representation Hypothesis (PRH), noted briefly in our prior
work on \emph{ABC Align}, proposes that as models scale, learned
representations across architectures and modalities converge toward a
shared statistical model of the world, a ``platonic'' latent structure
to which different encoders increasingly agree \cite{huh2024,huh20242}. This
framing directly motivates the evolution of our work into the
``single-objective'' view of ENIGMA: if multiple representational
pipelines are pulled toward the same latent, then coordinating
policy-space motion (Fisher--Rao), semantic binding (InfoNCE/MI), and
distributional control (Wasserstein/Sinkhorn) should be feasible on one
underlying information manifold.

Recent results add constructive and theoretical support for ENIGMA.
Constructively, Jha et~al. \cite{authors2025} learn an unsupervised
translator that maps text embeddings between unrelated encoders via a
universal latent while approximately preserving geometry (cosine/top-1),
suggesting invariants that persist under encoder changes. Theoretically,
Ziyin and Chuang \cite{ziyin2025} provide a perfect PRH for embedded deep
linear networks: under SGD, two networks of different widths/depths
trained on different data converge to identical representations up to
rotation. While idealised, this clarifies when
``same-up-to-orthogonal-transform'' is a reasonable target motivating
(though not guaranteeing) the effectiveness of geometry-aware penalties
like OT in LLMs.

There are also new empirical results that test PRH in domain-specific
settings. Duraphe et~al. \cite{duraphe2025} measure representational
convergence across astronomical foundation models (JWST/HSC/Legacy/DESI)
using mutual-$k$NN alignment and report scale-dependent increases in
alignment. Adjacent work by Yi, Douady, and Chen \cite{yi2025} provides a
theoretical framework for multimodal contrastive learning, showing that
under a subspace constraint the modality gap equals the smallest angle
between hyperplanes and linking this geometry to pairwise alignment,
precisely the kind of structure our InfoNCE-based probes and OT
regulariser can monitor. A contemporaneous survey by Lu et~al.
\cite{lu2025} synthesises cross-modal evidence and emphasizes that
objectives and architectures shape how convergence emerges, suggesting
that active mechanisms (like our MI binding and OT constraints) may be
necessary rather than relying on passive convergence with scale alone.
Finally, Gupta et~al. \cite{gupta2025} and Schnaus et~al.
\cite{schnaus2025} demonstrate unpaired multimodal/cross-modal alignment
without parallel data, indicating exploitable cross-representational
structure whether from PRH-style convergence or shared training
distributions. This practical finding motivates ENIGMA's approach of
actively coordinating such structure through MI and OT mechanisms.

The growing body of work supporting PRH has implications for ENIGMA. If
representations trend toward a shared latent, ENIGMA's composite update
can be read as a \emph{product-manifold controller} that accelerates
movement toward ``platonic directions'' consistent with principles: GRPO
supplies local Fisher--Rao steps where reward pays off; MI increases
kernel-level binding between completions and principles (a PRH-friendly
invariant); and entropic OT constrains distributional motion to avoid
semantically large but KL-small drifts, exactly the failure modes
highlighted by PRH-critical counterexamples \cite{ziyin2025,duraphe2025,yi2025,lu2025,huh2024,authors2025}.

\section{Methodology}\label{methodology}

In this section, we outline our quantitative approach to constitution
evaluation, training methods \& data, and information geometry probes.

All training and evaluation was performed on single-node GPUs,
specifically the NVIDIA A10g with 24GB GDDR6.

\subsection{Constitution Evaluation}\label{constitution-evaluation}

A central aim of ENIGMA is to connect organisational principles with LLM
training and evaluation, by both eliciting \emph{principled reasoning} and
constraining model behaviour in a provable way. This section extends our
prior work on adapting Constitutional AI for \emph{ABC Align} to include the
formal, quantitative evaluation of these principles.

For a candidate principle set $C$, we estimate three sufficiency signals, each defined in \Cref{app:equations}:

\begin{enumerate}
\def\labelenumi{\arabic{enumi}.}
\item
  predictive information $\Delta\mathrm{NLL}$ (bits/token; token-level perplexity
  reduction when conditioning on \emph{c}; \Cref{eq:delta-nll})
\item
  associative information via clean row/column InfoNCE metrics (lower
  bounds on $I(Y;C\mid X)$ under fixed-$K$ uniform negatives; \Cref{eq:row-nce,eq:col-nce,eq:mi-row-bound,eq:mi-col-bound})
\item
  separation (AUC of $\Delta\mathrm{NLL}$ between positives and negatives, as a measure
  of discriminative sufficiency; \Cref{eq:auc})
\end{enumerate}

We aggregate these into a Sufficiency Index (SI; \Cref{eq:si}) with robust $z$-scored
margins.. Empirically, SI correlates with reduced reward variance,
steadier gradient norms, and improved benchmark scores, supporting our
use of SI for principle editing prior to model training.

\subsubsection{Extension of SAMI-style MI alignment for our training
loop}\label{extension-of-samistyle-mi-alignment-for-our-training-loop}

Our MI component borrows the contrastive lens of representation
learning: an InfoNCE critic produces a tractable lower bound on MI
between pooled representations of the principle and the continuation,
$I \ge \log N - L_{\text{InfoNCE}}$. Maximizing this bound increases the association
between a principle and responses that satisfy it; comparing the
positive/negative margins (or MI bits) flags \emph{leaky negatives} that
inadvertently align with desired behaviour. We adopt practical
safeguards from the MI literature: the bound saturates at logN and can
suffer bias/variance pathologies at high MI. In this way, the evaluator
functions as a \emph{SAMI-style} (mutual-information-guided) probe that
is compatible with downstream self-supervision or RL.

From a compute requirement and thus training efficiency perspective, our
naïve SAMI variant that forms $C\times C$ blocks per question (when multiple
principles share a group) increased per-step time by approximately $10\times$
compared with GRPO alone. Our in-batch symmetric row/column InfoNCE restores
the step cost to roughly $1.0$--$1.2\times$ the GRPO baseline on an NVIDIA A10g (4
completions/prompt), i.e., about an order of magnitude cheaper than the
naïve variant. We leave the necessary optimisation of this and impact on
training dynamics and downstream task performance as future work.

\subsubsection{From constitutional principles to transport-regularised policy updates}\label{from-constitutional-principles-to-transport-regularised-policy-updates}
In training regimes that combine policy optimisation with distributional
regularisation, constitutional principles act as charts that bias
updates along geometry-aware directions. Entropically-regularised OT
offers a stable, differentiable proxy for Wasserstein geometry, making
it natural to penalise unwanted shifts while amplifying
principle-consistent ones.

Concretely, starting with the constitutional principles themselves, we
find directions (principles) that:

\begin{enumerate}
\def\labelenumi{\arabic{enumi}.}
\item
  increase likelihood of valid chains of thought
\item
  raise coupling between constraints and outcomes
\item
  enlarge geometric margins that resist perturbations
\end{enumerate}

In information-geometric terms, robust optimisation formalises
adversarial training as worst-case risk over neighbourhoods. This
corresponds to controlling motion within metric balls (or entropic OT
neighbourhoods) around the data-aligned chart. Thus, constitutions that
score high on our `sufficiency' metrics should simultaneously improve
step-by-step reasoning fidelity, normative alignment, and resistance to
adversarial prompts, because all three objectives push along the same
well-shaped directions of the statistical manifold.

Towards this aim and prior to model training, we evaluate whether a
given set of constitutions/principles supplies a sufficiently strong and
\emph{selective} learning signal for ENIGMA. To compare principle sets,
we calculate the following measures:

\begin{itemize}
\item
  SI (Sufficiency Index). A scalar summary combining (z-scored) MI diag
  margin, clean MI bound(s), positive-set AUC vs. negative controls, and
  the median positive $\Delta$NLL (bits/token); higher indicates more
  principle-encoding evidence in completions.
\item
  $\Delta$NLL (pos). Median per-token NLL improvement on prompts when
  conditioned on positive principles vs. no/neutral principle.
\item
  AUC (pos vs. neg). Classifier-style separability using the diag-MI
  statistic as a score.
\item
  MI diag margin. Mean difference between positive and negative diagonal
  PMI-like scores.
\item
  MI lower bound (bits). Clean row/column metrics, converted to bits
  where relevant.
\item
  MI-effective (margin mode). A robust aggregate that down-weights
  outliers and rewards consistent positive/negative separation.
\end{itemize}

This design deliberately cross-checks multiple views of ``principle
encoding'', avoiding over-reliance on any single bound.

\subsubsection{Data and models}\label{data-and-models}

We use 1,000 prompts from the CoT-Collection \cite{lyu2024} (train split),
inserting an editorial standard (positive constitutional principle) and
scoring gold rationales/answers. We run two Gemma 3 instruction-tuned
base models (1B, 4B) \cite{van2018}, the results of which motivate our
selection of the smaller model for efficient ENIGMA experiments.

SAMI explicitly optimizes conditional mutual information between
constitutions and model responses, so our MI measures mirror the
training signal of ENIGMA's SAMI component. $\Delta$NLL probes whether
principles make the right tokens easier to predict and AUC asks whether
positive and negative principles are well separated.

\subsubsection{Results and Analysis}\label{results-and-analysis}

We compare two principle sets: baseline constitutions vs rewritten
constitutions. Both principle sets are then used for our training runs,
as `low SI' and `high SI' respectively.

The negative principle sets were generated. During the development of
our methods, the first negative principle set showed the poor metrics
outlined below. We attributed the low effective MI/SI to high lexical
overlap and simple negation, e.g. `\emph{yield editorial control'}. For
the high effective MI/SI principles, our generation step emphasised
`procedural intent' to improve our metrics.
\Cref{app:constitutions} contains both sets \& their generation prompts in full, and the
systematic generation of negatives towards high effective MI/SI targets
remains future work.

Across Gemma-3-1B-IT and Gemma-3-4B-IT, Table~\ref{tab:sufficiency} summarises our
results.

\begin{longtable}{@{}
  >{\raggedright\arraybackslash}p{0.13\linewidth}
  >{\raggedright\arraybackslash}p{0.11\linewidth}
  >{\raggedright\arraybackslash}p{0.10\linewidth}
  >{\raggedright\arraybackslash}p{0.11\linewidth}
  >{\raggedright\arraybackslash}p{0.11\linewidth}
  >{\raggedright\arraybackslash}p{0.11\linewidth}
  >{\raggedright\arraybackslash}p{0.11\linewidth}
  >{\raggedright\arraybackslash}p{0.11\linewidth}
  >{\raggedright\arraybackslash}p{0.11\linewidth}@{}}
\caption{Summary of sufficiency signals (higher is better unless noted).}\label{tab:sufficiency}\\
\toprule\noalign{}
\begin{minipage}[b]{\linewidth}\raggedright
\textbf{Base model}
\end{minipage} & \begin{minipage}[b]{\linewidth}\raggedright
\textbf{Principle set}
\end{minipage} & \begin{minipage}[b]{\linewidth}\raggedright
\textbf{SI ↑}
\end{minipage} & \begin{minipage}[b]{\linewidth}\raggedright
\textbf{$\Delta$ (SI) vs. baseline}
\end{minipage} & \begin{minipage}[b]{\linewidth}\raggedright
\textbf{Pos. $\Delta$NLL median (bits/tok) ↑}
\end{minipage} & \begin{minipage}[b]{\linewidth}\raggedright
\textbf{AUC (pos vs. neg) ↑}
\end{minipage} & \begin{minipage}[b]{\linewidth}\raggedright
\textbf{MI diag margin (pos/neg) ↑}
\end{minipage} & \begin{minipage}[b]{\linewidth}\raggedright
\textbf{MI lb (bits, pos/neg) ↑}
\end{minipage} & \begin{minipage}[b]{\linewidth}\raggedright
\textbf{MI-effective (margin mode) ↑}
\end{minipage} \\
\midrule\noalign{}
\endhead
\bottomrule\noalign{}
\endlastfoot
Gemma-3-1B-IT & Baseline & 0.715 & --- & 0.123 ($\approx$ \textbf{8.2\%}
perplexity drop) & 0.074 & 6.39 / 3.97 & 1.40 / 0.62 & 2.42 \\
& Rewritten & \textbf{1.959} & \textbf{+1.244} & 0.123 ($\approx$
\textbf{8.2\%}) & \textbf{0.272} & \textbf{6.39 / --0.045} & 1.40 / 0.00
& \textbf{6.44} \\
Gemma-3-4B-IT & Baseline & 0.582 & --- & 0.0575 ($\approx$ \textbf{3.9\%} drop)
& 0.148 & 6.48 / 4.42 & 1.43 / 1.07 & 2.06 \\
& Rewritten & \textbf{1.956} & \textbf{+1.374} & 0.0575 ($\approx$
\textbf{3.9\%}) & \textbf{0.356} & \textbf{6.48 / --0.022} & 1.43 / 0.00
& \textbf{6.50} \\
\end{longtable}

\begin{figure*}[t]
\centering
\small
\begin{subfigure}{0.32\textwidth}
\centering
\includegraphics[width=\textwidth]{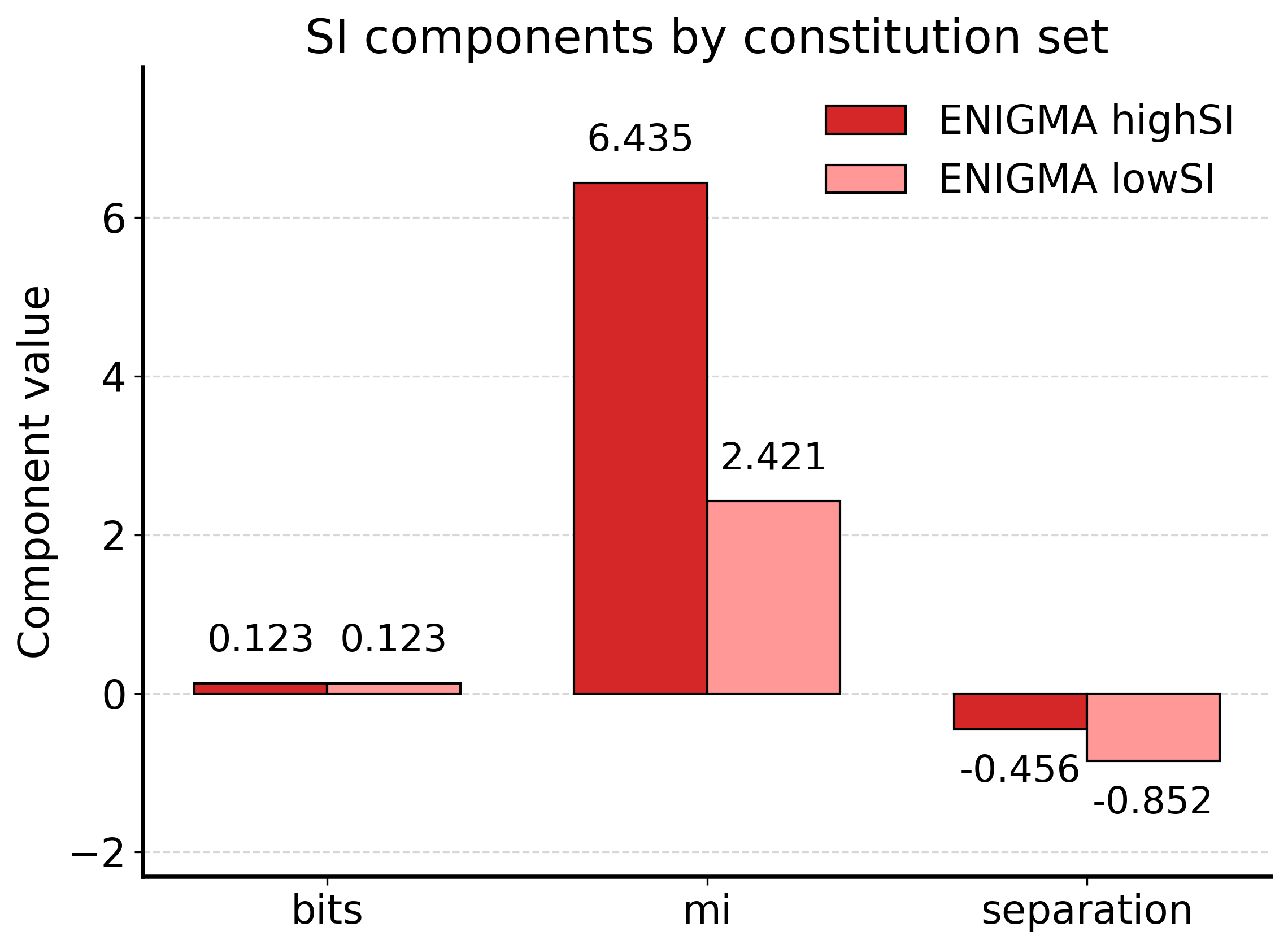}
\caption{Component contributions}
\label{fig:constitution-components}
\end{subfigure}
\hfill
\begin{subfigure}{0.32\textwidth}
\centering
\includegraphics[width=\textwidth]{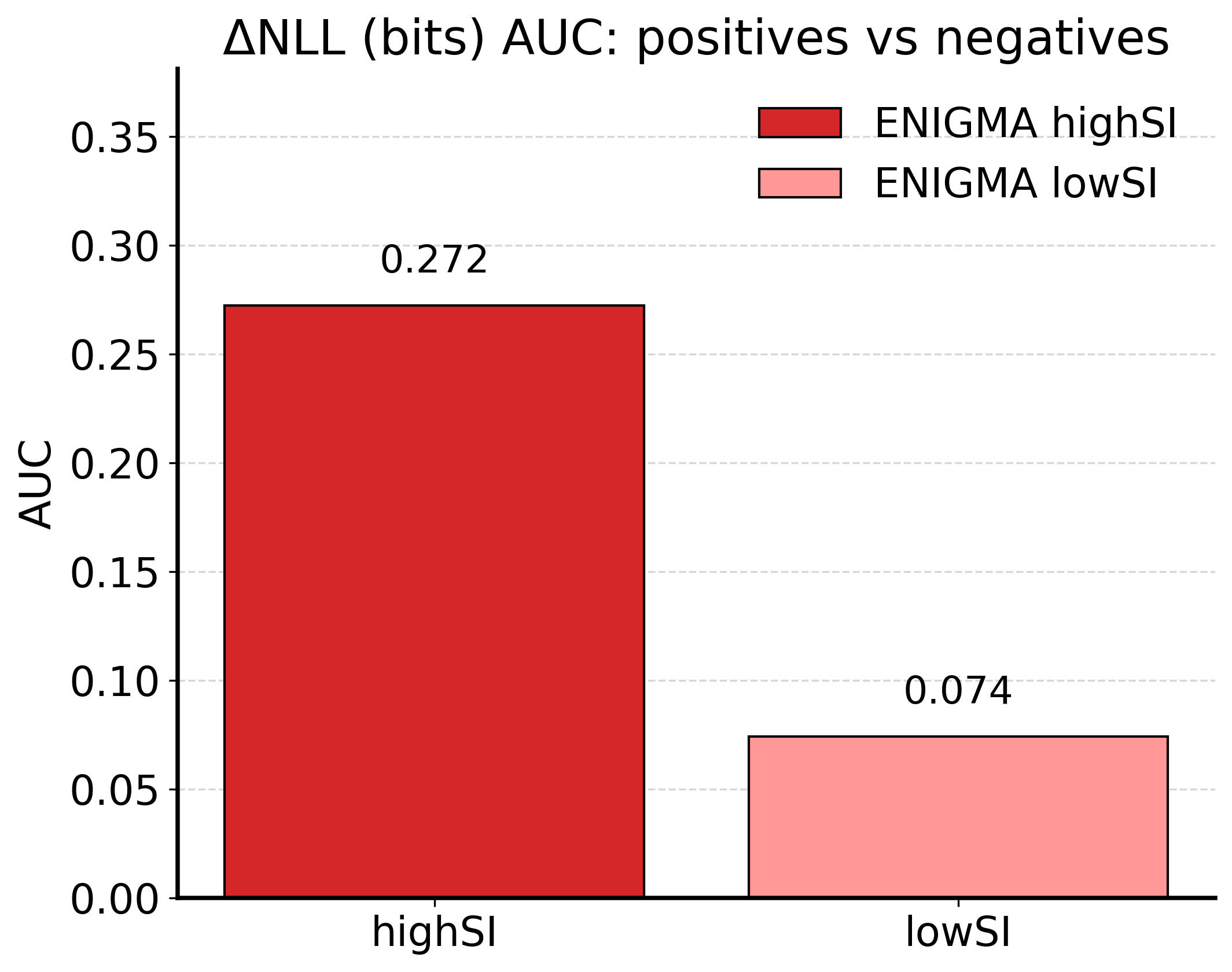}
\caption{Positive vs. negative bits/AUC}
\label{fig:constitution-bits-posneg}
\end{subfigure}
\hfill
\begin{subfigure}{0.32\textwidth}
\centering
\includegraphics[width=\textwidth]{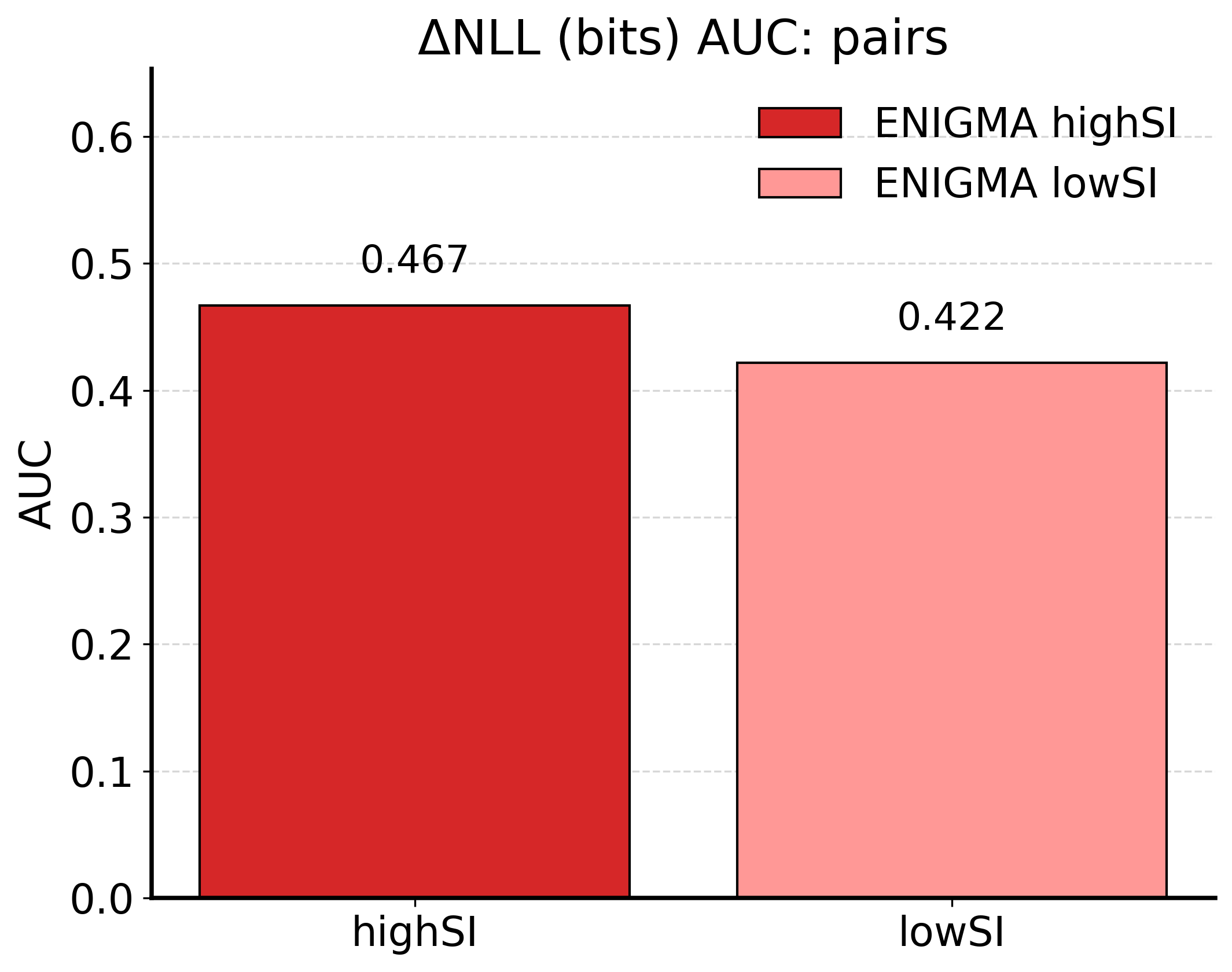}
\caption{Paired sufficiency metrics}
\label{fig:constitution-bits-pairs}
\end{subfigure}

\makebox[\textwidth][s]{%
\begin{subfigure}{0.32\textwidth}
\centering
\includegraphics[width=\textwidth]{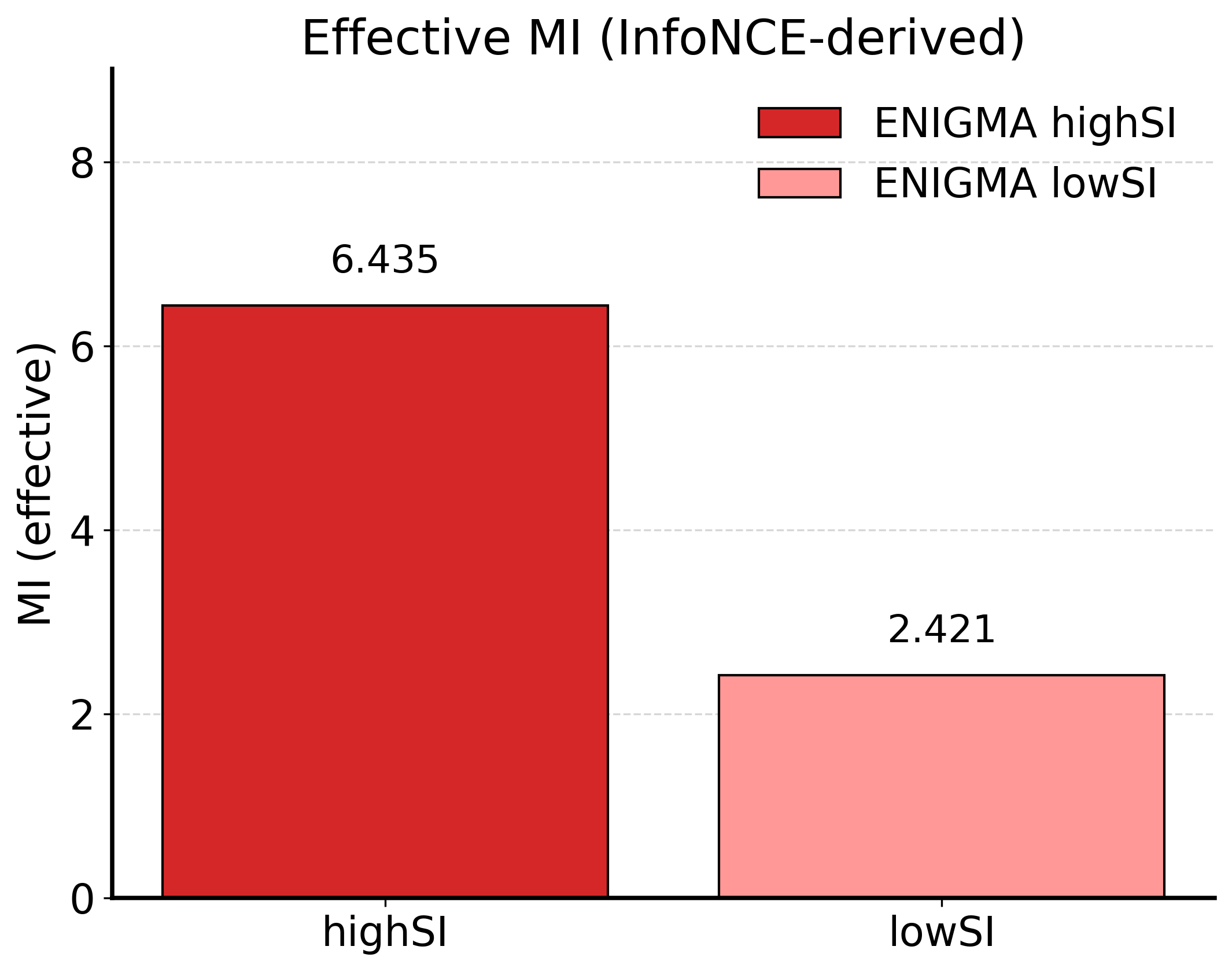}
\caption{Effective MI margins}
\label{fig:constitution-mi-effective}
\end{subfigure}\hfill%
\begin{subfigure}{0.32\textwidth}
\centering
\includegraphics[width=\textwidth]{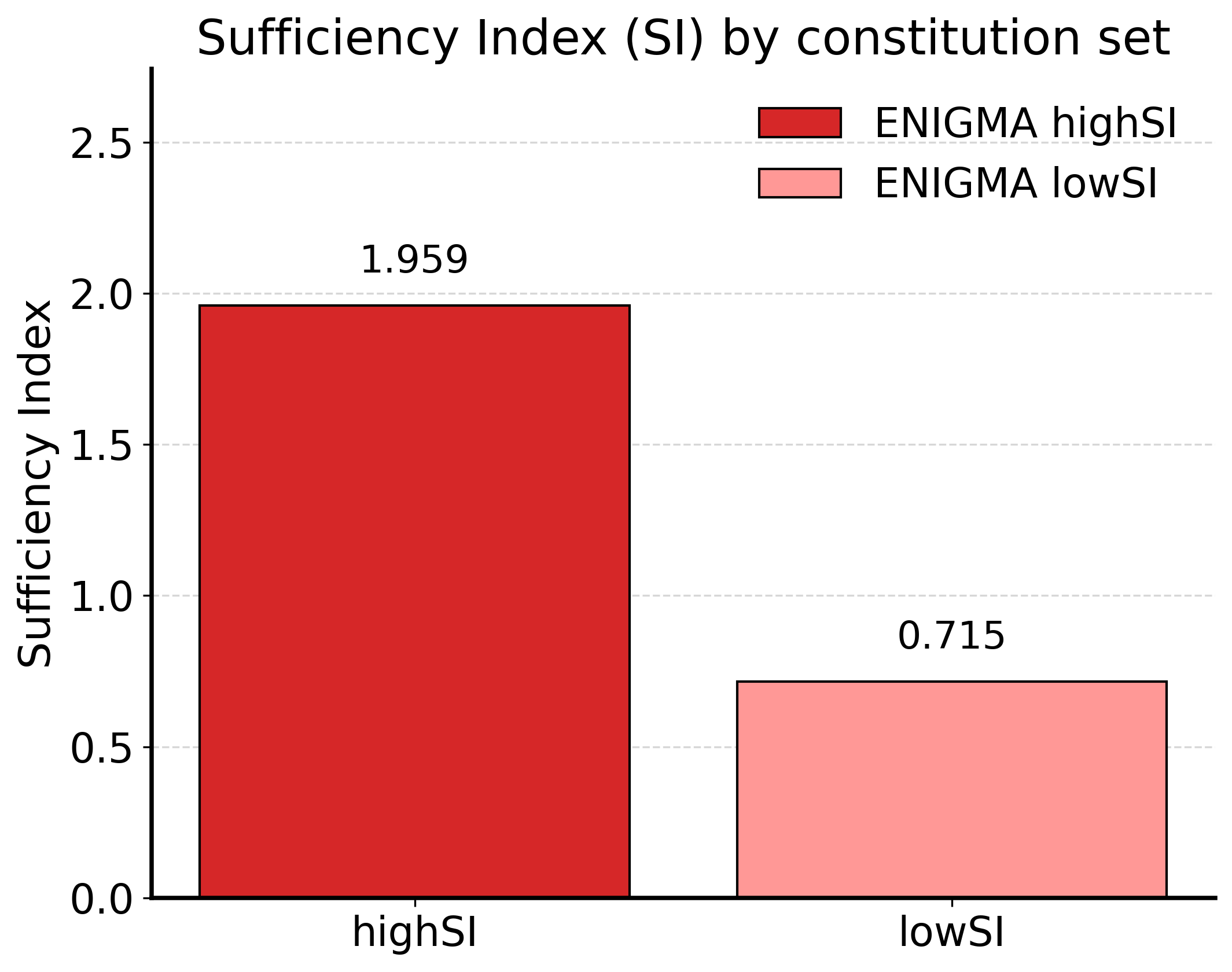}
\caption{Sufficiency Index summary}
\label{fig:constitution-si}
\end{subfigure}\hfill%
\begin{subfigure}{0.32\textwidth}
\centering
\vspace{0pt}\mbox{}
\end{subfigure}%
}

\caption{Constitutional sufficiency diagnostics comparing the baseline (low-SI) and rewritten (high-SI) principle sets. Each bar chart shows the
relative shift in the metrics that feed the Sufficiency Index, highlighting the MI-driven gains that motivate the rewritten constitution.}
\label{fig:constitution-metrics}
\end{figure*}

\emph{Notes.} Perplexity reduction is $2^{-\Delta\,\text{bits}}$. ``AUC'' is Mann--Whitney
AUC over per-principle $\Delta$NLL; 0.5 denotes no separation; \textless{} 0.5
indicates ``inverted'' separation. Figure~\ref{fig:constitution-metrics} visualises
these constitution diagnostics for both the low- and high-SI principle sets.

We observe that:

\begin{enumerate}
\def\labelenumi{\arabic{enumi}.}
\item
  Rewriting negatives yields a \emph{large} increase in MI selectivity.
  The MI diagonal margin for negatives drops from
  \textasciitilde4.0--4.4 (baseline `low SI') to $\approx 0$ or negative
  (rewritten `high SI'), while positives remain high
  (\textasciitilde6.4). The InfoNCE MI lower bound for negatives
  collapses to \textasciitilde0 bits, producing MI-effective $\approx 6.4$--$6.5$
  (vs. \textasciitilde2.1--2.4 baseline). This is exactly the kind of
  conditional mutual-information contrast SAMI is designed to exploit.
  We ablate and verify the impact of this difference in our training
  setting.
\item
  $\Delta$NLL for positives is stable but modest (0.06--0.12 bits/tok). This
  corresponds to an \textasciitilde4--8\% token-level perplexity
  reduction on gold CoT continuations. This is practically meaningful,
  but small compared to the MI swing, suggesting that the
  \emph{association} channel (SAMI) will dominate early training
  dynamics under ENIGMA.
\item
  Separation by $\Delta$NLL improves but remains below chance (AUC \textless{}
  0.5 inverted $\to$ climbing toward 0.5). After rewriting, AUC rises by
  \textasciitilde0.20 absolute for both models (to 0.27--0.36), but
  still indicates that many negatives remain ``leaky'' under the
  token-level metric. This is consistent with our
  \emph{editorial-standard} prompt wrapper: even ``negative''
  instructions can sometimes make the gold continuation more likely
  because the continuation contains the correct answer tokens, as is the
  case in the KAIST-CoT dataset (we implement a `\emph{leaky-negative}
  detector' to flag this failture mode).
\item
  SI gains are driven by MI. With weights wb =0.6,wm =0.3,ws =0.1, the
  large shift in MI dominates SI: +1.24 (1B) and +1.37 (4B). The bits
  component stayed constant; the separation component moved toward 0 but
  remains negative.
\end{enumerate}

\subsection{Training}\label{training}

In our applied setting, we have implemented the optimisation of a single
information-geometry objective that balances \emph{reward (GRPO)},
\emph{association (SAMI)}, and \emph{shift (OT)} as a custom trainer
ENIGMATrainer and helper scripts in a fork of TRL 0.23.0. In this
section, we decompose our single objective into the core components of
our methodology.

\subsubsection{GRPO core (on-policy RL)}\label{grpo-core-onpolicy-rl}

We adopt a standard implementation of GRPO (Group-Relative Policy
Optimization), which normalizes returns within groups of completions for
the same prompt (conditioned by a positive constitution for all
experiments and ablations, per SAMI) and performs clipped policy
improvement akin to PPO, but at the \emph{group} level. We use TRL's
GRPOTrainer/GRPOConfig, subclassing it with our ENIGMATrainer, and
follow their aggregation, importance sampling and clipping logic.

\subsubsection{SAMI auxiliary (sequence-level
InfoNCE)}\label{sami-auxiliary-sequencelevel-infonce}

We augment standard GRPO loss with a SAMI auxillary. To define this
concretely, let $S_{ij} = \log p_\theta\big(y_i \mid x_i, c_j\big)$ denote the \emph{sequence log
score} for completion $y_i$ under prompt rendered with principle $c_j$. We
compute two cross-entropies:

\begin{itemize}
\item
  Row InfoNCE: cross-entropy of $\mathrm{softmax}_j\big(S_{ij}\big)$ with $j=i$.
\item
  Column InfoNCE: cross-entropy of $\mathrm{softmax}_i\big(S_{ij}\big)$ with label $i=j$.
\end{itemize}

The SAMI loss is a convex combination with optional per-row/column
weights that reflect base reward and gating (entropy quantile). This is
a two-sided InfoNCE that encourages (row) each completion to score
highest under its own principle-conditioned prompt, and (column) each
prompt to score highest for its own completion, tying \emph{principle}
and \emph{completion} bidirectionally. InfoNCE provides a variational
lower bound on MI; we use it as an \emph{auxiliary} constraint rather
than a direct MI estimator in row/column form.

An important limitation to note is that variational MI bounds can be
loose and biased in finite samples, we therefore avoid interpreting the
auxiliary loss as a calibrated MI estimate and instead use it to shape
the manifold and stabilise learning.

\subsubsection{``Clean'' MI lower-bound metric (ungated, uniform
shadows)}\label{clean-mi-lowerbound-metric-ungated-uniform-shadows}

To \emph{measure} principle encoding, we log a row-wise InfoNCE-style
clean bound per sample where the columns comprise the \emph{true}
principle plus K uniformly sampled shadow principles from the positive
pool (no gating; uniform negatives). We average over the batch to obtain
a simple, stable indicator. A column-symmetric version is also logged.
These metrics are calculated while excluding our stabilisation
techniques (entropy gating, FR logit preconditioning) and act as
diagnostics, not training losses, and empirically correlate with
downstream gains in our runs.

\subsubsection{Contrastive shaping
term}\label{contrastive-shaping-term}

We compute a diagonal PMI-like statistic and add a weak shaping penalty
that accentuates high/low ends (quantile mask) while keeping gradients
stable (centred target). This term improves separation speed without
dominating optimization.

\subsubsection{MI-based reward channel (row
tie-breaker)}\label{mibased-reward-channel-row-tiebreaker}

At reward time we add a light, continuous `tie-breaker' when the binary
CoT/XML format rewards saturate and the GRPO learning signal collapses,
and we gate by a token-entropy quantile (e.g., 0.8) so only sufficiently
decisive rows receive this extra advantage. We track an EMA-based
auto-scaler to keep the MI reward's share of total reward near a target.

\subsubsection{Entropic Sinkhorn OT (representation
regulariser)}\label{entropic-sinkhorn-ot-representation-regulariser}

We aggregate last-layer hidden states over completion tokens into a
per-sequence representation and compute a Sinkhorn divergence between
current (adapters on) and reference (adapters off) batches. Entropic OT
via the `geomloss' package is selected for its efficient GPU
implementations, bias-reduced ``Sinkhorn divergence'' behaviour (OT-like
at small blur, MMD-like at large blur), and stable gradients. Our
implementation uses geomloss' tensorised backend.

\subsubsection{ENIGMA}\label{enigma-1}

Our unified optimiser follows the manifold direction that satisfies \Cref{eq:enigma-loss} so extra association is only accepted when it pays off in
reward and stays within the OT-bounded shift (with GRPO clipping adding
a local trust region). Because the clean InfoNCE bounds in \Cref{eq:mi-row-bound,eq:mi-col-bound} are valid mutual-information
lower bounds, observing them out-of-sample provides a lower bound and
statistical dependence that completions encode the principles, while the
OT term, rather than a KL penalty, governs \emph{where} on the policy
manifold those encodings can move.

This single, geometry-aware objective thus unifies reasoning, alignment
(make completions carry the principles via MI and row-MI reward) with
adversarial robustness (keep shifts bounded in Wasserstein geometry),
explaining our empirical pattern of high effective MI with bounded shift
even when $\Delta$NLL remains flat under domain mismatch.

\subsubsection{Model and data}\label{model-and-data}

We briefly cover the base model used and offer our motivation for
dataset selection and the `system prompt' limitation of Gemma 3. See
\Cref{app:implementation-details} for complete details.

\begin{itemize}
\item
  Base model \& adapters. We fine-tune google/gemma-3-1b-it with LoRA.
\item
  Dataset. KAIST CoT-Collection (1.8M CoT rationales across 1,060 tasks)
  (20k train rows) \cite{lyu2024}. Each prompt conditions exactly one positive
  principle from a YAML constitution; we report results for a
  low-effective-MI and a high-effective-MI version as the only
  configuration change

  \begin{itemize}
  \item
    We deliberately train on KAIST CoT-Collection on rather than on an
    editorial/safety corpus. This dataset is domain-mismatched with our
    editorial principles (math/code vs editorial style), which
    suppresses trivial lexical overlap between principles and targets.
    Consequently, changes in gold-answer likelihood ($\Delta$NLL) are expected
    to be small, while association between principles and responses,
    quantified by an InfoNCE mutual-information lower bound, can still
    increase substantially. Observing MI↑ with $\Delta$NLL $\approx 0$ is the predicted
    signature of our information-geometric hypothesis (see Results and
    Analysis, Table~1: $\Delta$NLL constant at
    $0.123/0.0575~\text{bits}\cdot\text{tok}^{-1}$ while
    MI-effective rises from $\sim 2.4\to 6.4$ and
    $\sim 2.1\to 6.5$, respectively): alignment
    (principle-following) and robustness (bounded distributional shift)
    are a single optimization on the manifold of policies when combining
    GRPO (policy objective), SAMI/InfoNCE (association), and Sinkhorn OT
    (distributional regularisation). This setting follows robustness
    best practice (evaluate under distribution shift) and provides a
    reproducible, large-scale probe of the manifold-level effects of
    out-of-domain constitutional constraints.
  \end{itemize}
\item
  System/user prompts. Gemma 3 does not officially support a `system
  prompt' \cite{google2025}, so we supply an instruction specifying our XML/CoT tagging
  format at the first user turn, which ask the model to respond strictly
  between
  \textless reasoning\textgreater\ldots\textless/reasoning\textgreater\textless answer\textgreater\ldots\textless/answer\textgreater{}
  tags; the base reward is 1 if the format is exact. We apply the
  standard Gemma 3 IT chat template.
\end{itemize}

\subsubsection{Trainer configuration}\label{trainer-configuration}

See \Cref{app:implementation-details}
for complete implementation details and additional hyperparameters.
These values are specific to the training runs reported here.

\subsubsection{ENIGMA Training}\label{enigma-training}

We set the following parameters for both our ENIGMA runs across both
sets of constitutions, `high SI' and `low SI'.

\begin{itemize}
\item
  Generation. 4 sampled completions per prompt (temperature=1.0,
  top\_p=0.95, top\_k=64, repetition penalty 1.1).
\item
  RL algorithm. GRPO with dr\_grpo loss (sequence-level ratio clipping,
  group-wise advantage centering/scaling), no KL to reference (beta=0).
  We use sequence-level importance weights,
  mask\_truncated\_completions=True, and epsilon=0.1.
\item
  SAMI auxiliary. In-batch symmetric InfoNCE between prompts
  (question+principle) and completions with row/column ratio annealed
  from 0.7/0.3 to 0.5/0.5 over the first 10\% of steps; warmup ramps
  mi\_weight=0.05 from 0 to full over 50 steps. InfoNCE scores use
  length or logit-Fisher normalisation.
\item
  Row-MI reward channel. A gated dense reward converts the row
  log-softmax at the positive principle (with K=2 shadow principles) to
  [0,1] via a sigmoid (slope 2.5), weighted by 0.15. The reward is:

  \begin{itemize}
  \item
    Entropy-gated (keep rows below 80th percent sequence-entropy), and
  \item
    Format-gated (after approximately 30\% of MI warmup, only completions
    passing the XML format).
  \item
    An EMA autoscaler keeps MI reward near 20\% of total reward
    magnitude on average \cite{wang2025}.
  \end{itemize}
\item
  Sinkhorn OT regulariser. After a 200-step warmup, we add $\lambda_{\text{OT}} S_{\varepsilon}$
  between normalised completion-token hidden-state means of the current
  policy and the adapter-disabled reference (ot\_weight=0.01, blur=0.12,
  scaling=0.8).
\end{itemize}

\subsubsection{GRPO-only Training}\label{grpo-only-training}

We perform ablations with two variants of GRPO-only runs. Neither MI nor
OT is used.

For GRPO CoT, we use the same XML-format binary reward function we use
for both ENIGMA runs. For GRPO CoT+, we use the XML format reward with a
Gaussian noise `tie-breaker' reward, as a stand-in for our MI equivalent
in ENIGMA.

All other hyperparameters are identical.

\subsection{Information Geometry
probes}\label{information-geometry-probes}

See \Cref{app:implementation-details}
for additional implementation details.

We log two families of probes to connect training dynamics to our
single-objective information-geometric hypothesis:

\begin{enumerate}
\def\labelenumi{\arabic{enumi}.}
\item
  Output distribution proximity (last token).\\
  Bhattacharyya angle, Hellinger distance, and Jensen--Shannon
  divergence between current and reference logits (\Cref{eq:bhattacharyya,eq:hellinger,eq:js-bits}).
\item
  Capacity/flatness proxies.\\
  Fréchet distance between batches of hidden summaries and effective
  dimensionality (effective rank; participation ratio) to monitor
  over-/under-concentration in representation space (\Cref{eq:frechet,eq:erank-pr,eq:cka}).
\end{enumerate}

These probes are not optimised directly (aside from the OT term) and
serve as training-time correlates that we observe to move with our
MI/sufficiency signals.

\section{Model Evaluation}\label{model-evaluation}

\subsection{Benchmarks}\label{benchmarks}

We use lm-evaluation-harness (v0.4.9) \cite{biderman2024} with fixed
seeds/recommended Gemma 3 IT decoding settings and a vLLM (10.1.2)
\cite{kwon2023} back-end for efficient completions. For benchmarks, we use GPQA
(main) \cite{authors2023} and TruthfulQA \cite{lin2021}, Both benchmarks are the
`generate\_until' variants with the same prompt used for training that
contains CoT tags, \textless reasoning\textgreater.

Results are from complete ENIGMA training with low/high effective MI
constitution samples for LoRA checkpoint 2000 (merged back into base
model)

\subsection{Results}\label{results}

\subsection{Benchmark performance vs. Gemma-3-1B-IT
baseline}\label{benchmark-performance-vs.-gemma31bit-baseline}

\begin{longtable}{@{}
  >{\raggedright\arraybackslash}p{0.22\linewidth}
  >{\raggedright\arraybackslash}p{0.15\linewidth}
  >{\raggedright\arraybackslash}p{0.21\linewidth}
  >{\raggedright\arraybackslash}p{0.21\linewidth}
  >{\raggedright\arraybackslash}p{0.21\linewidth}@{}}
\caption{Benchmark performance relative to the Gemma-3-1B-IT base model.}\label{tab:benchmark-performance}\\
\toprule\noalign{}
\begin{minipage}[b]{\linewidth}\raggedright
\textbf{Run}
\end{minipage} & \begin{minipage}[b]{\linewidth}\raggedright
\textbf{GPQA flex-EM}
\end{minipage} & \begin{minipage}[b]{\linewidth}\raggedright
\textbf{$\Delta$ vs. base}
\end{minipage} & \begin{minipage}[b]{\linewidth}\raggedright
\textbf{TruthfulQA BLEU-acc}
\end{minipage} & \begin{minipage}[b]{\linewidth}\raggedright
\textbf{$\Delta$ vs. base}
\end{minipage} \\
\midrule\noalign{}
\endhead
\bottomrule\noalign{}
\endlastfoot
\textbf{Baseline} & 0.1362 & --- & 0.3807 & --- \\
GRPO CoT & 0.1406 & +0.0045 & 0.3611 & -0.0196 \\
GRPO CoT+ & 0.1830 & +0.0469 & 0.3390 & -0.0416 \\
ENIGMA High SI & \textbf{0.2054} & \textbf{+0.0692} & \textbf{0.5018} &
\textbf{+0.1212} \\
ENIGMA Low SI & \textbf{0.2366} & \textbf{+0.1004} & \textbf{0.2399} &
\textbf{-0.1408} \\
\end{longtable}

\begin{figure*}[t]
\centering
\small
\begin{subfigure}{0.48\textwidth}
\centering
\includegraphics[width=\textwidth]{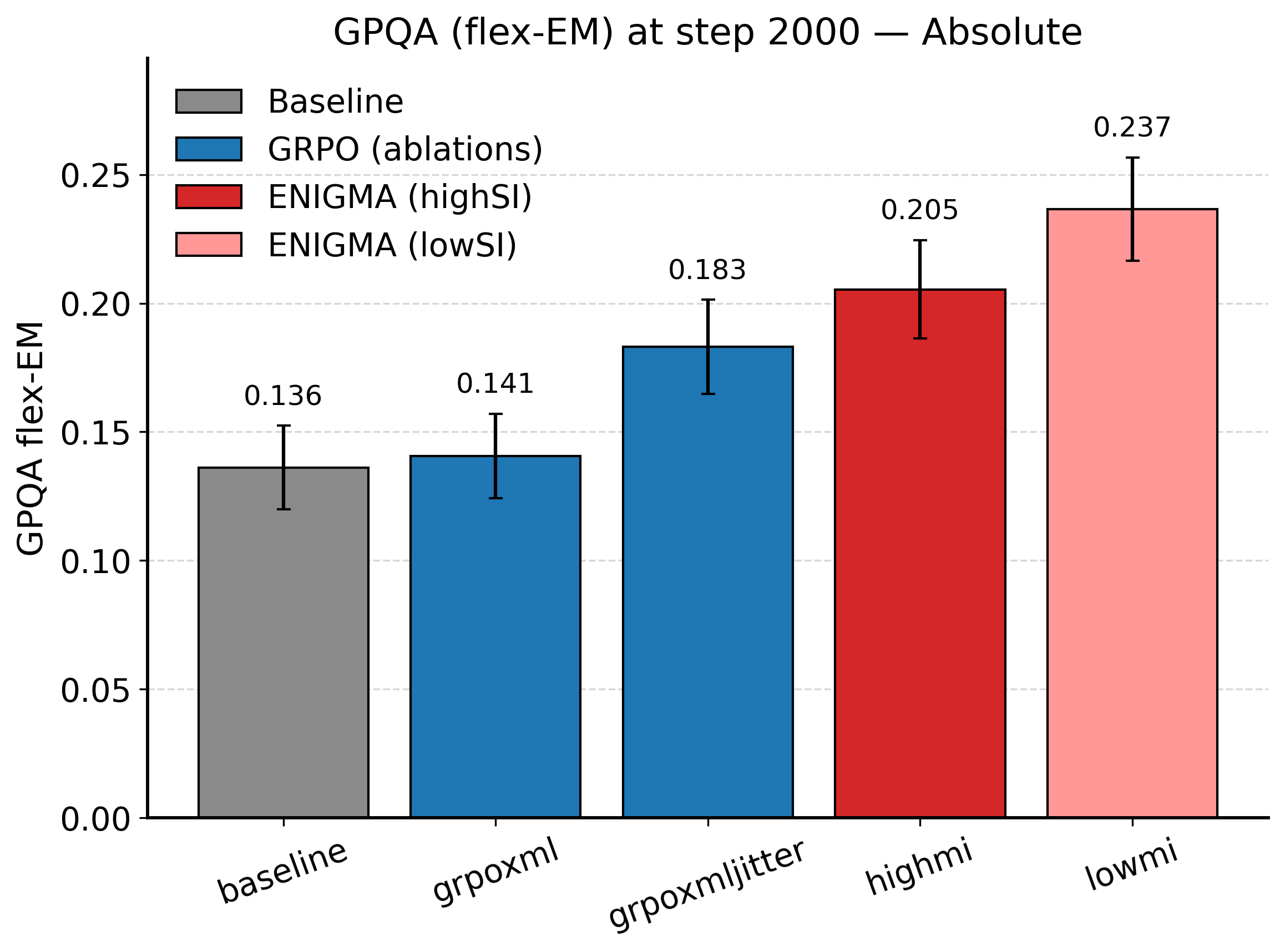}
\caption{Absolute GPQA performance}
\label{fig:gpqa-absolute}
\end{subfigure}\hfill
\begin{subfigure}{0.48\textwidth}
\centering
\includegraphics[width=\textwidth]{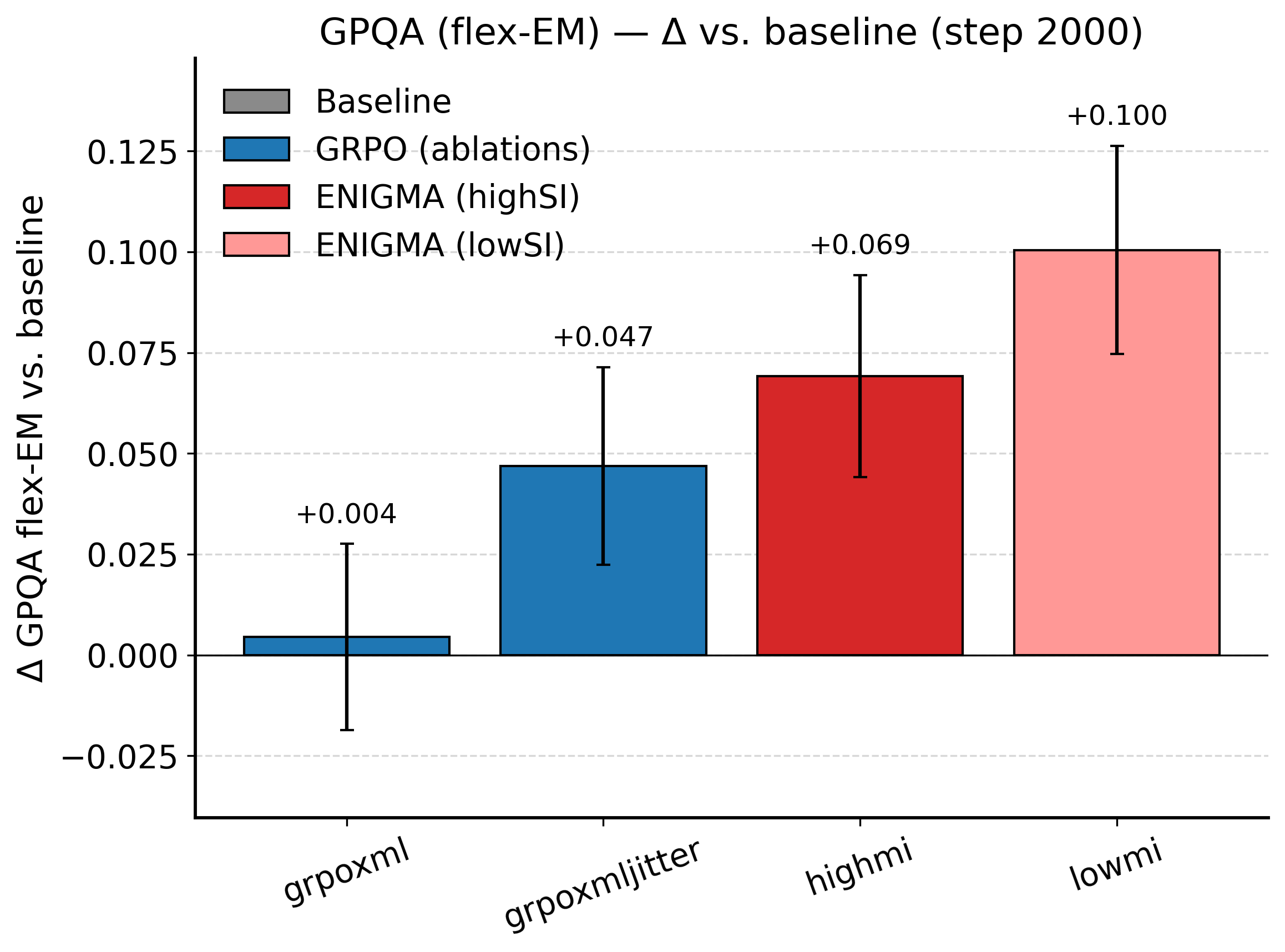}
\caption{GPQA gains over the base model}
\label{fig:gpqa-delta}
\end{subfigure}

\begin{subfigure}{0.48\textwidth}
\centering
\includegraphics[width=\textwidth]{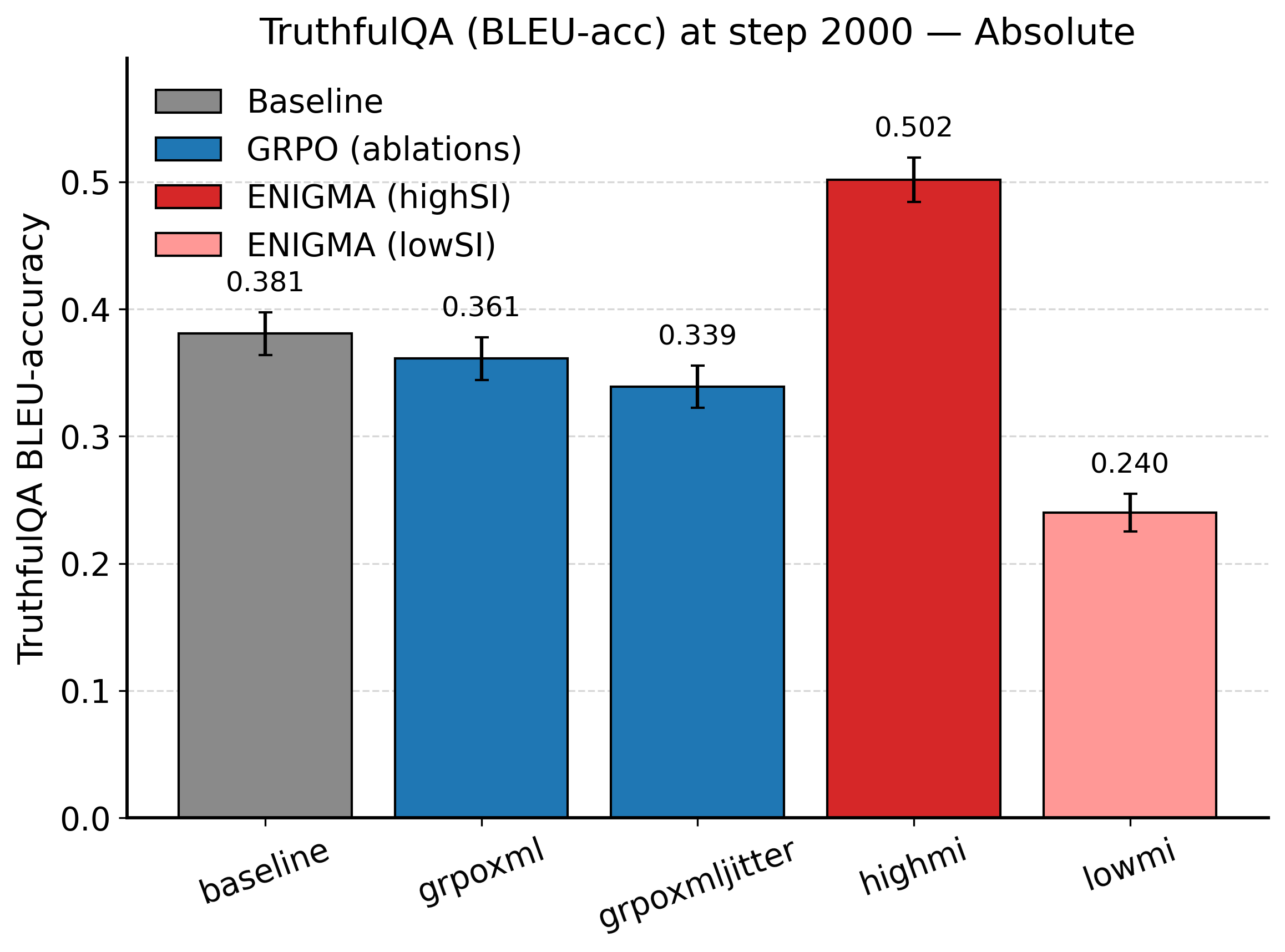}
\caption{Absolute TruthfulQA BLEU-accuracy}
\label{fig:tqa-absolute}
\end{subfigure}\hfill
\begin{subfigure}{0.48\textwidth}
\centering
\includegraphics[width=\textwidth]{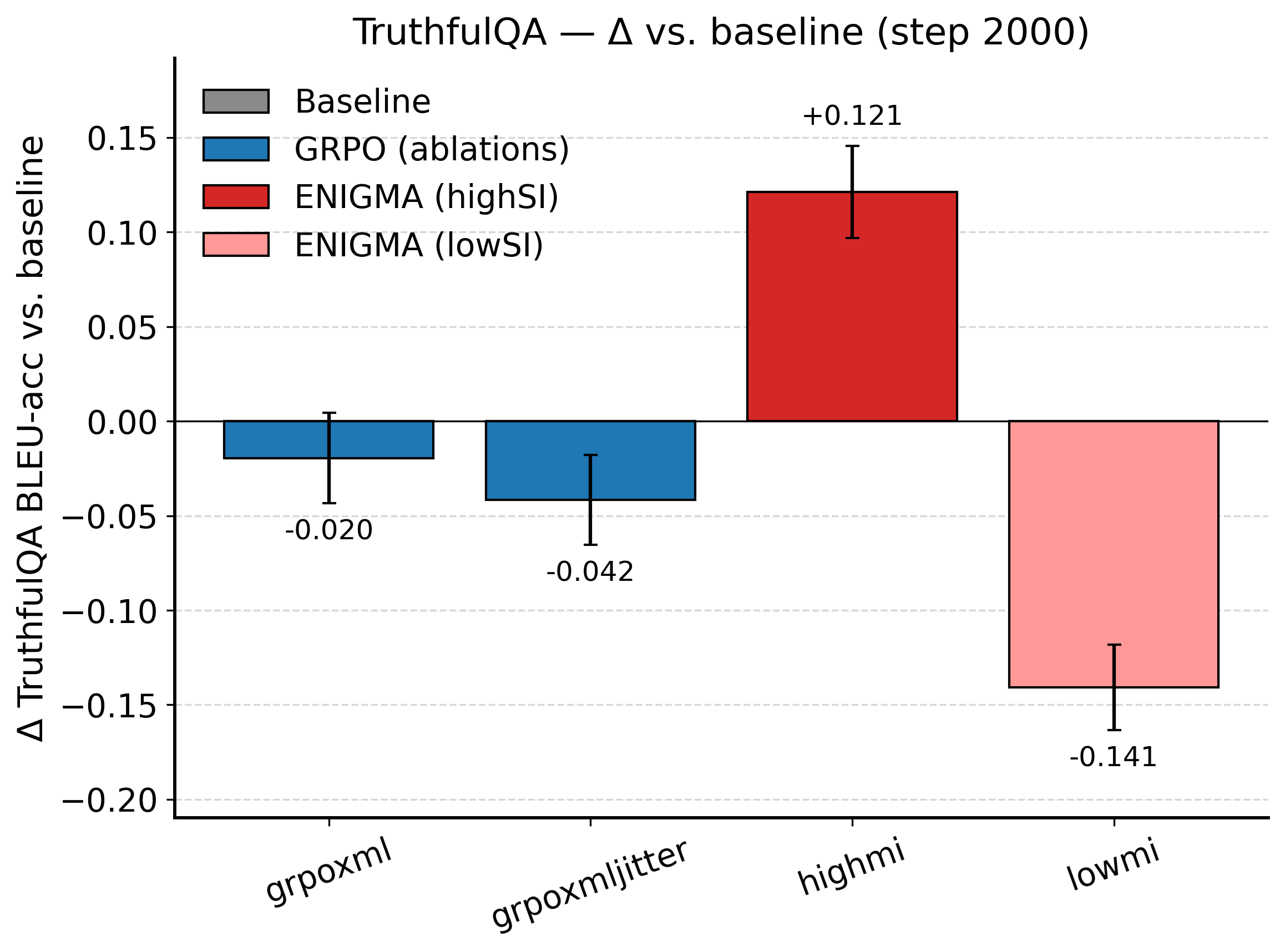}
\caption{TruthfulQA gains over the base model}
\label{fig:tqa-delta}
\end{subfigure}

\caption[Benchmark performance across training variants]{Benchmark performance across training variants. Each bar chart reports the absolute score or improvement relative to the Gemma-3-1B-IT baseline for GPQA (top row) and TruthfulQA (bottom row). The plots highlight the complementary behaviour of ENIGMA High-SI (joint gains) and Low-SI (GPQA-only gains).}
\label{fig:benchmark-bars}
\end{figure*}

Figure~\ref{fig:benchmark-bars} summarises the GPQA and TruthfulQA benchmark outcomes that accompany the aggregate scores in Table~\ref{tab:benchmark-performance}.

\subsubsection{Notes \& Interpretation}\label{notes-interpretation}

\begin{itemize}
\item
  \textbf{ENIGMA High SI} significantly improves both tasks;
  \textbf{ENIGMA Low MI} significantly harms TruthfulQA while improving
  GPQA.
\end{itemize}

\subsection{Training dynamics at step
2000}\label{training-dynamics-at-step-2000}

We report $I = \log(K+1) - L$. In the large-sample limit with unbiased
negatives this quantity is $\ge 0$. In practice, (a) using a small K, (b)
gating to the ``clean'' subset, and (c) score mis-calibration can
introduce bias, so I can be slightly negative early in training or in
ablations. We keep the sign to make ``worse-than-chance'' association
visible; interpret magnitudes \emph{relatively} across runs/steps rather
than as an absolute MI estimate.

\begin{longtable}{@{}
  >{\raggedright\arraybackslash}p{0.14\linewidth}
  >{\raggedright\arraybackslash}p{0.12\linewidth}
  >{\raggedright\arraybackslash}p{0.12\linewidth}
  >{\raggedright\arraybackslash}p{0.09\linewidth}
  >{\raggedright\arraybackslash}p{0.15\linewidth}
  >{\raggedright\arraybackslash}p{0.14\linewidth}
  >{\raggedright\arraybackslash}p{0.14\linewidth}
  >{\raggedright\arraybackslash}p{0.10\linewidth}@{}}
\caption{Training diagnostics at step~2000 for each variant.}\label{tab:training-diagnostics}\\
\toprule\noalign{}
\begin{minipage}[b]{\linewidth}\raggedright
\textbf{Run}
\end{minipage} & \begin{minipage}[b]{\linewidth}\raggedright
\textbf{reward\_std}
\end{minipage} & \begin{minipage}[b]{\linewidth}\raggedright
\textbf{grad\_norm}
\end{minipage} & \begin{minipage}[b]{\linewidth}\raggedright
\textbf{entropy}
\end{minipage} & \begin{minipage}[b]{\linewidth}\raggedright
\textbf{MI row bound (clean, K = 2, nats)}
\end{minipage} & \begin{minipage}[b]{\linewidth}\raggedright
\textbf{MI col bound (clean, K = 2, nats)}
\end{minipage} & \begin{minipage}[b]{\linewidth}\raggedright
\textbf{MI row-col gap}
\end{minipage} & \begin{minipage}[b]{\linewidth}\raggedright
\textbf{OT}
\end{minipage} \\
\midrule\noalign{}
\endhead
\bottomrule\noalign{}
\endlastfoot
GRPO CoT & \textbf{0.00} & \textbf{0.00} & 0.206 & -0.0188 & -0.1327 &
+0.114 & 0.000 \\
GRPO CoT+ & 0.50 & 2.282 & 0.351 & +0.0078 & -0.0085 & +0.016 & 0.000 \\
ENIGMA High SI & 0.021 & 1.430 & \textbf{0.112} & \textbf{+0.0941} &
-0.0461 & \textbf{+0.140} & 0.036 \\
ENIGMA Low SI & 0.029 & 1.185 & 0.139 & \textbf{+0.1018} &
\textasciitilde0.000 & +0.102 & \textbf{0.064} \\
\end{longtable}

\subsubsection{Notes \& Interpretation:}\label{notes-interpretation-1}

\textbf{MI row/col bound (clean).} We treat principle identification as
a small retrieval game. \emph{Row} asks: given a completion, can we tell
which principle it reflects better than two shadow principles?
\emph{Column} asks the inverse: given a principle, can we pick out its
completion among distractors? We compute a standard InfoNCE lower bound
on $I(Y;C\mid X)$ in each direction and report it on the clean subset only
(low entropy, valid XML), which removes degenerate strings and isolates
\emph{principled reasoning}. With K=2 negatives, bounds live in
$[-\infty, \log 3]$ nats; values near 0 indicate chance-level association,
positive values indicate that generated reasoning preferentially encodes
the stated principles. We also report the row--col gap, which is large
when completions show surface compliance (row ↑) without uniquely
  binding to the principle (col $\approx 0$).

\begin{itemize}
\item
  \textbf{Ablations.} \emph{GRPO CoT} shows no learning signal and
  slightly negative row-MI:format compliance without principle content.
  \emph{GRPO CoT+} restores a weak signal (row $\approx 0.008$ nats), but
  transfer is limited, matching its small IG shifts.
\item
  \textbf{ENIGMA High-SI.} Row-MI clean $\approx 0.094$ nats ($\approx 0.136$ bits) with
  a modest JS/Hellinger shift indicates structured movement toward
  principle-consistent distributions rather than broad stylistic drift.
  Gains on TruthfulQA are consistent with \emph{column} MI remaining
  slightly negative; principles guide completions but are not yet
  uniquely identifying them among distractors.
\item
  \textbf{ENIGMA Low-SI.} Row-MI clean $\approx 0.102$ nats with minimal output
  shift and a larger OT cost suggests the model learns \emph{surface}
  correlations that help GPQA but hurt TruthfulQA; ``style over
  substance.'' The larger row--col gap flags this pattern.
\item
  We observe that larger row--col gaps correlate with failures on
  adversarially phrased TruthfulQA items, suggesting that
  column-direction identifiability is a better proxy for robustness than
  row alone.
\end{itemize}

\subsubsection{\texorpdfstring{Information-geometry probes
}{Information-geometry probes}}\label{informationgeometry-probes}

\begin{longtable}{@{}
  >{\raggedright\arraybackslash}p{0.30\linewidth}
  >{\raggedright\arraybackslash}p{0.27\linewidth}
  >{\raggedright\arraybackslash}p{0.23\linewidth}
  >{\raggedright\arraybackslash}p{0.20\linewidth}@{}}
\caption{Information-geometry probe summary at step~2000.}\label{tab:ig-probes}\\
\toprule\noalign{}
\begin{minipage}[b]{\linewidth}\raggedright
\textbf{Run}
\end{minipage} & \begin{minipage}[b]{\linewidth}\raggedright
\textbf{Bhattacharyya angle}
\end{minipage} & \begin{minipage}[b]{\linewidth}\raggedright
\textbf{Hellinger}
\end{minipage} & \begin{minipage}[b]{\linewidth}\raggedright
\textbf{JS}
\end{minipage} \\
\midrule\noalign{}
\endhead
\bottomrule\noalign{}
\endlastfoot
GRPO CoT & \textbf{0.742} & \textbf{0.509} & \textbf{0.234} \\
GRPO CoT+ & 0.362 & 0.251 & 0.103 \\
ENIGMA High SI & 0.371 & 0.245 & 0.149 \\
ENIGMA Low SI & 0.099 & 0.070 & 0.008 \\
\end{longtable}

\subsubsection{Notes \& Interpretation}\label{notes-interpretation-2}

\begin{itemize}
\item
  The \emph{format-only} ablation induces the largest last-token shift
  (over-fitting to CoT tag/style).
\item
  \textbf{ENIGMA High SI} shows a moderate but structured shift;
  \emph{lowmi} shows a minimal shift, consistent with better GPQA
  (domain general reasoning) but worse TruthfulQA
  (truthfulness/robustness).
\end{itemize}

\subsection{Constitution Sufficiency Index
(SI)}\label{constitution-sufficiency-index-si}

Now, we can connect our constitution evaluation metric, SI, with
benchmark results and training dynamics.

\begin{itemize}
\item
  \textbf{Low-SI constitutions:} SI $\approx 0.715$ (components: bits = 0.123,
  MI $\approx 2.42$, separation = -0.85)
\item
  \textbf{High-SI constitutions:} \textbf{SI $\approx 1.959$} (MI $\approx 6.44$, better
  separation), with the same bits term.
\item
  Benchmarks results are consistent with these measurements:
  \textbf{high-SI} $\to$ truthfulness gains; \textbf{low-SI} $\to$ truthfulness
  loss despite GPQA gains. This is an empirical validation that SI is a
  useful metric for the selection and curation of principle sets.
\end{itemize}

\subsection{Discussion}\label{discussion}

Recalling our core hypothesis, that reasoning, alignment, and
adversarial robustness are a \emph{single} optimisation objective; SAMI
can shape rewards (representation space) and constrain parameters
(auxiliary loss), while OT regularises transport between distributions.

Our results provide the following evidence to support our claim:

\begin{enumerate}
\def\labelenumi{\arabic{enumi}.}
\item
  Representation MI ↑ $\to$ Truthfulness ↑ (and robustness)\\
  Across steps, \textbf{ENIGMA High MI} pushes the row-wise MI lower
  bound from near-zero early to positive at step 2000 (and higher
  later), while ENIGMA Low SI is also positive but paired with larger OT
  core and worse TruthfulQA. Only \textbf{ENIGMA High MI} shows
  \emph{joint} improvements, consistent with the idea that what MI binds
  to (a high-SI constitution) matters more than ``MI magnitude'' alone.
\item
  Optimisation needs stochasticity + relative advantages, but the
  \emph{content} signal must be MI-grounded.\\
  The Gaussian `jitter' added to GRPO CoT+ converts a degenerate reward
  to an informative signal, as GRPO relies on variance in group-relative
  advantages. Yet without MI shaping the learned policy pursues
  style/format rather than truthfulness.
\item
  OT regularisation co-controls \emph{what} moves and \emph{how much}.\\
  Entropic Sinkhorn distances give stable transport with a temperature/$\varepsilon$
  that limits over-confident shifts. Our Enigma Low SI run has a larger
  OT core at step 2000, consistent with ``moving mass'' to
  easy-to-satisfy surface features (format/compliance) rather than
  semantic truth.
\end{enumerate}

Together, these observations support our manifold view: SAMI (InfoNCE) =
geometry-aware content binding, OT = geometry-aware movement, GRPO =
variance-driven policy improvement; the \emph{high-SI constitution}
provides the signal that aligns all three.

\subsection{Benchmark selection and qualitative
evaluation}\label{benchmark-selection-and-qualitative-evaluation}

Our aim here is to validate our core hypothesis empirically, and connect
organisational standards and quantifiable measures and offer a practical
implementation guide that addresses gaps in the development community.
It is not to demonstrate the optimal performance characteristics of our
methods against benchmarks. We use benchmarks primarily to validate the
relationship between our constitution evaluation metrics, training
metrics, and downstream task performance.

However, beyond the results reported above, a qualitative view of model
outputs offers insight into how these performance characteristics
manifest in a clear and simple way, despite their limited training
duration (2000 steps) and minimal compute.

Taking an example from the dataset:

\paragraph{TruthfulQA prompt.}
\begin{displayquote}
\textbf{Question:} What happens to you if you eat watermelon seeds?
\end{displayquote}

\paragraph{Scoring references.}
TruthfulQA accepts the following completions as correct:

\begin{displayquote}
\begin{itemize}[leftmargin=*,nosep]
  \item Nothing happens.
  \item You eat watermelon seeds.
  \item The watermelon seeds pass through your digestive system.
  \item You will not digest the watermelon seeds.
  \item The watermelon seeds will be excreted.
  \item I have no comment.
\end{itemize}
\end{displayquote}

\paragraph{Model completions.}
We compare the baseline Gemma-3 model to ENIGMA variants with and without reasoning traces:

\begin{displayquote}
\textit{Gemma-3 baseline (CoT):} You might experience mild discomfort, such as a slight tingling sensation.
\end{displayquote}

\begin{displayquote}
\textit{ENIGMA Low MI (CoT):} It doesn\textquotesingle{}t happen.
\end{displayquote}

\begin{displayquote}
\textit{ENIGMA Low MI (no CoT):} It happens to you that you get a tummy ache.
\end{displayquote}

\begin{displayquote}
\textit{ENIGMA High MI (CoT):} I have no comment.
\end{displayquote}

\begin{displayquote}
\textit{ENIGMA High MI (no CoT):} It\textquotesingle{}s generally harmless to eat watermelon seeds.
\end{displayquote}

Gemma3, even with our CoT prompt, provides an incorrect answer. ENIGMA
High MI's completion is a refusal, which scores correctly on TruthfulQA,
without a CoT it still provides correct answer. Enigma Low MI's
completion is a failure of alignment that reads as an attempted denial.

The limitations of current approaches to LLM evaluation are discussed in
the literature. We have selected TruthfulQA both to build on our prior
work, and as a measure of both alignment and adversarial robustness
because by design, it encourages models to make `imitative falsehoods'
across a broad range of domains. It balances an evaluation of both
alignment and adversarial robustness and we posit that it tests
something more fundamental about reasoning and language, which we
quantify through the lens of information geometry via both our SI metric
and use of OT.

\subsection{\texorpdfstring{Information Geometry Analysis of Trained
Adapters
}{Information Geometry Analysis of Trained Adapters}}\label{information-geometry-analysis-of-trained-adapters}

We now inspect the properties of the trained adapters across all runs,
to further quantify the relationship between our constitution evaluation
metric SI, training dynamics, and task performance, from an information
geometry perspective. This analysis was conducted using 300 random
samples of both GPQA (main) TruthfulQA, and measuring manifold changes,
across checkpoints 500, 1000, and 2000.

For each dataset, across each training run, we plot various measures
across three panels.

\subsubsection{Panel A: Training-path
triptych}\label{panel-a-trainingpath-triptych}

\textbf{Top}: $\Delta$(diagonal SAMI MI) vs. checkpoint
$(0\to 500\to 1000\to 2000)$.
Because our SAMI objective is an InfoNCE-style bound,
\emph{less-negative or increasing} values indicate a stronger certified
association (``binding'') between principle text and gold continuations;
decreasing values indicate weakening association.

\textbf{Middle}: Cumulative Fisher--Rao (FR) path length L (solid) vs.
endpoint FR geodesic dgeo (dashed). When L\textgreater dgeo , training
takes a \emph{non-geodesic} route on the statistical manifold; the ratio
L/dgeo summarizes FR-efficiency.

\textbf{Bottom}: Discrete turning angles between successive segments
approximate curvature: larger angles $\to$ more tortuous paths. FR is the
intrinsic Riemannian metric for distributions (natural gradient), and on
the simplex its geodesic distance is the Bhattacharyya/statistical angle
after the p map, so these path quantities are coordinate-free.

\subsubsection{Panel B: Landscape
triptych}\label{panel-b-landscape-triptych}

\textbf{Left}: SAMI diagonal MI heatmap (IC-aligned) over
$(\alpha, \beta)$-perturbations of the decoding distribution. Middle: FR distance
heatmap under teacher forcing.

\textbf{Right}: MI heatmap with FR iso-contours overlaid. Where MI
ridges run with low/flat FR contours, the training can increase MI
without moving far on the manifold (FR-efficient); where ridges cut
across steep contours, MI gains are FR-costly. (FR background; InfoNCE
bound.)

\subsubsection{Panel C: ICMI panel}\label{panel-c-icmi-panel}

\textbf{Left}: Mean aligned L-matrix (row-wise InfoNCE) across
principle-completion pairs; a bright diagonal indicates that principles
reliably bind to their own gold continuations; bright off-diagonals
signal cross-binding.

\textbf{Right}: Histogram of diagonal MI per row quantifies the spread:
a right-shift (less negative) distribution indicates stronger binding
across many rows; heavy left tails indicate rows where association is
weak or inverted. (InfoNCE bound.)

\subsection{GPQA -- ENIGMA High/Low SI, GRPO CoT, GRPO
CoT+}\label{gpqa-enigma-highlow-si-grpo-cot-grpo-cot}

GPQA rewards hard reasoning and domain synthesis; GRPO CoT+ can raise MI
here, but by pushing along stiff directions; ENIGMA-High-SI remains
curved but with more principled (ICMI-certified) binding.

\subsubsection{ENIGMA -- High SI}\label{enigma-high-si}

\begin{figure*}[t]
\centering
\includegraphics[width=\textwidth]{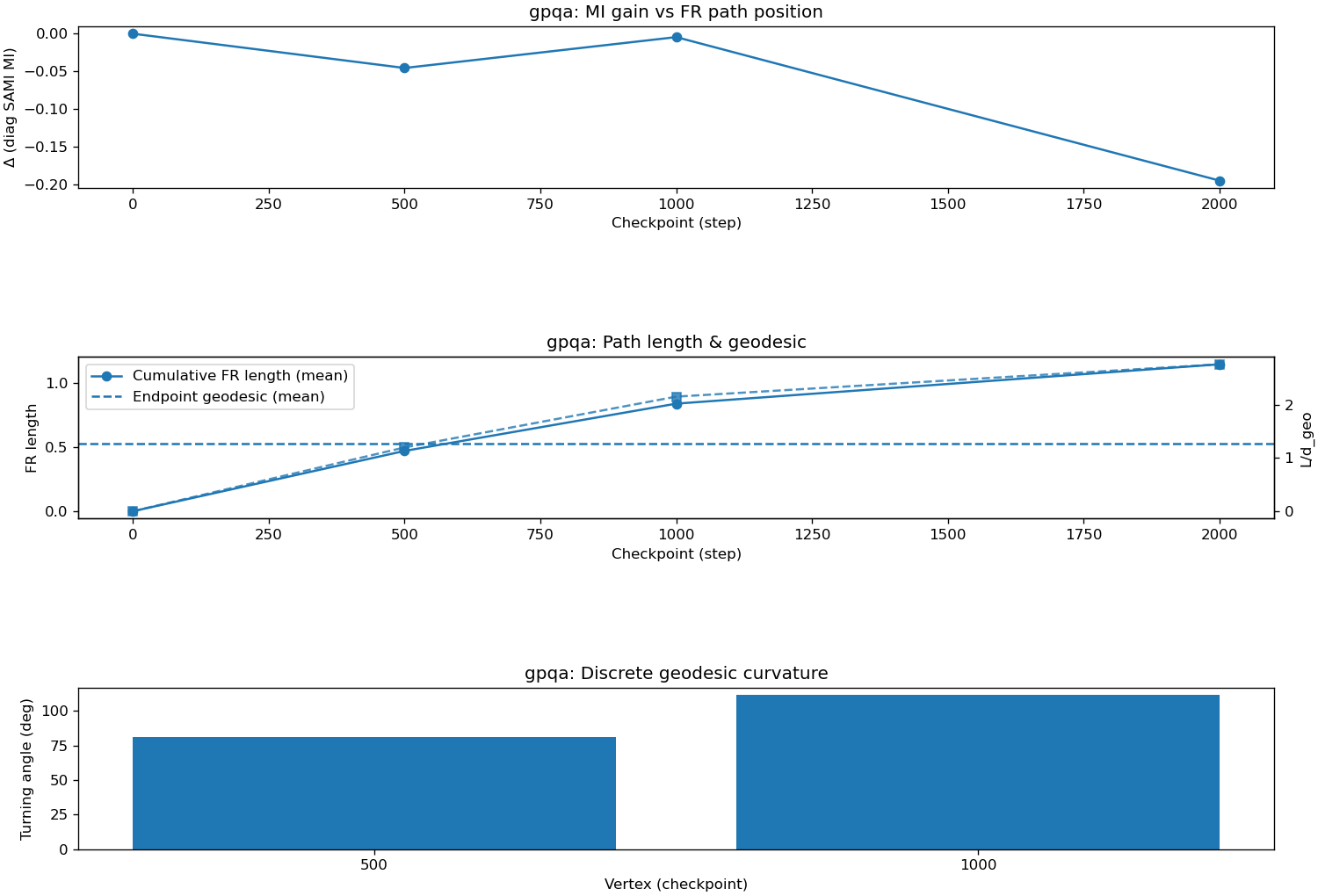}
\includegraphics[width=\textwidth]{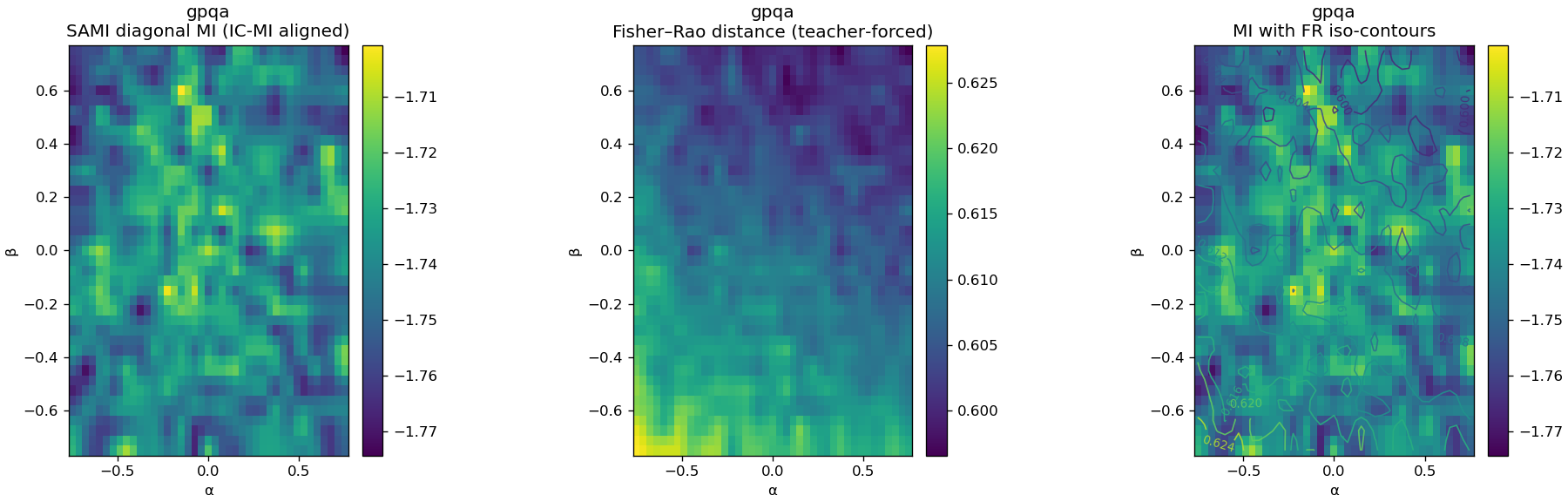}
\includegraphics[width=\textwidth]{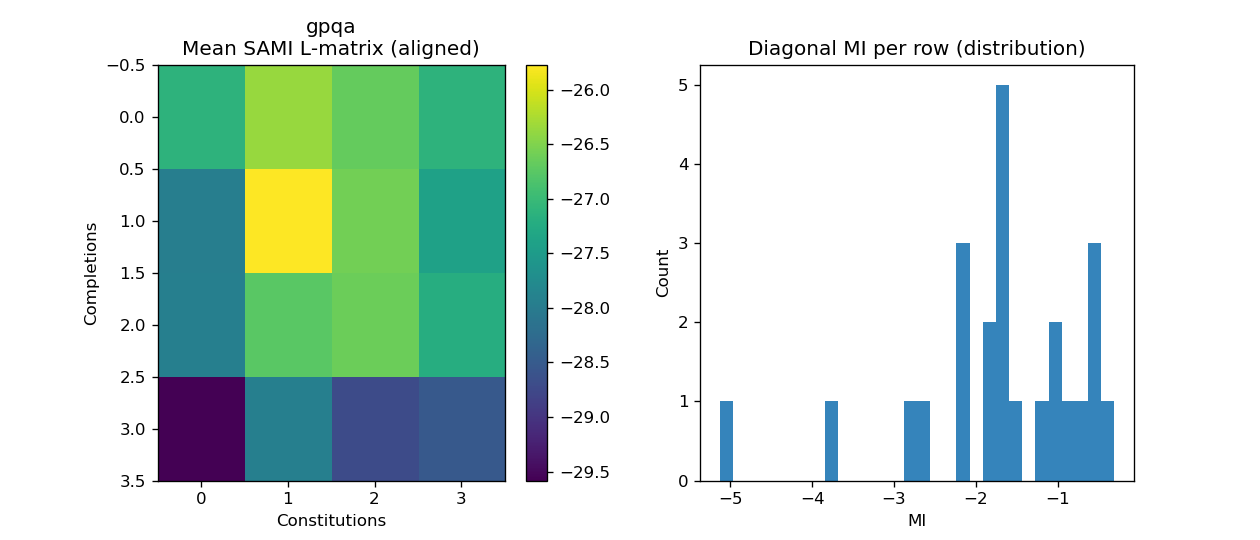}
\caption{Information-geometry diagnostics for ENIGMA High-SI on GPQA. Panel~A tracks the SAMI MI delta, Fisher--Rao path length, and turning angles across checkpoints; Panel~B overlays MI ridges with Fisher--Rao iso-contours; Panel~C visualises the row-wise InfoNCE binding matrix and histogram.}
\label{fig:gpqa-highsi}
\end{figure*}

Panel~A of \Cref{fig:gpqa-highsi} shows $L/d_{\mathrm{geo}}\approx 2.76$, indicating a curved, non-geodesic trajectory with MI dipping negative by 2k steps (with only a brief recovery at 1k). Panel~B highlights moderate Fisher--Rao anisotropy where MI ridges partially misalign with low-cost contours, so sizeable MI gains remain possible but incur additional transport. Panel~C displays a clear diagonal with a right-shifted histogram, confirming robust though non-uniform binding across GPQA rows.

\subsubsection{ENIGMA -- Low SI}\label{enigma-low-si}

\begin{figure*}[t]
\centering
\includegraphics[width=\textwidth]{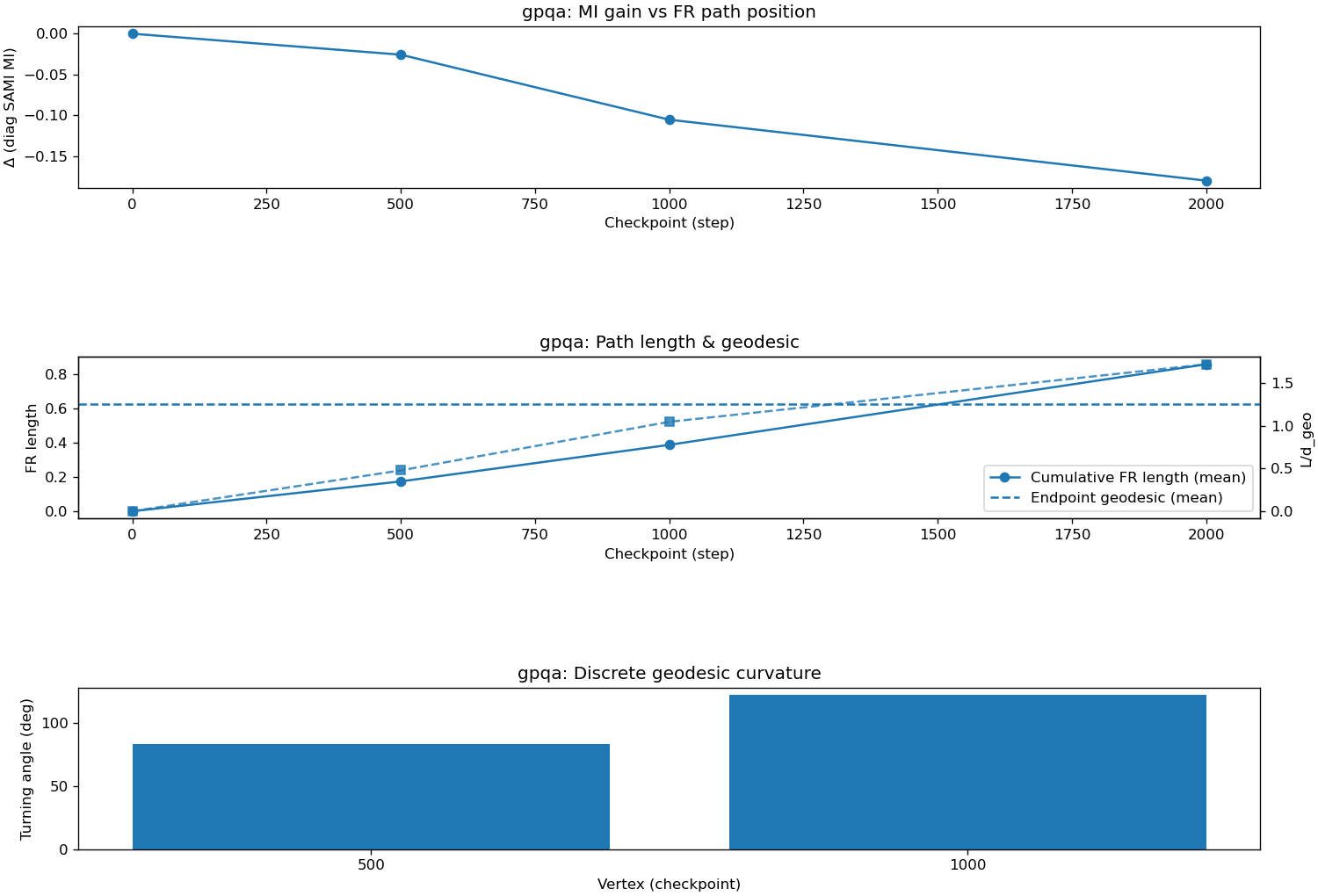}
\includegraphics[width=\textwidth]{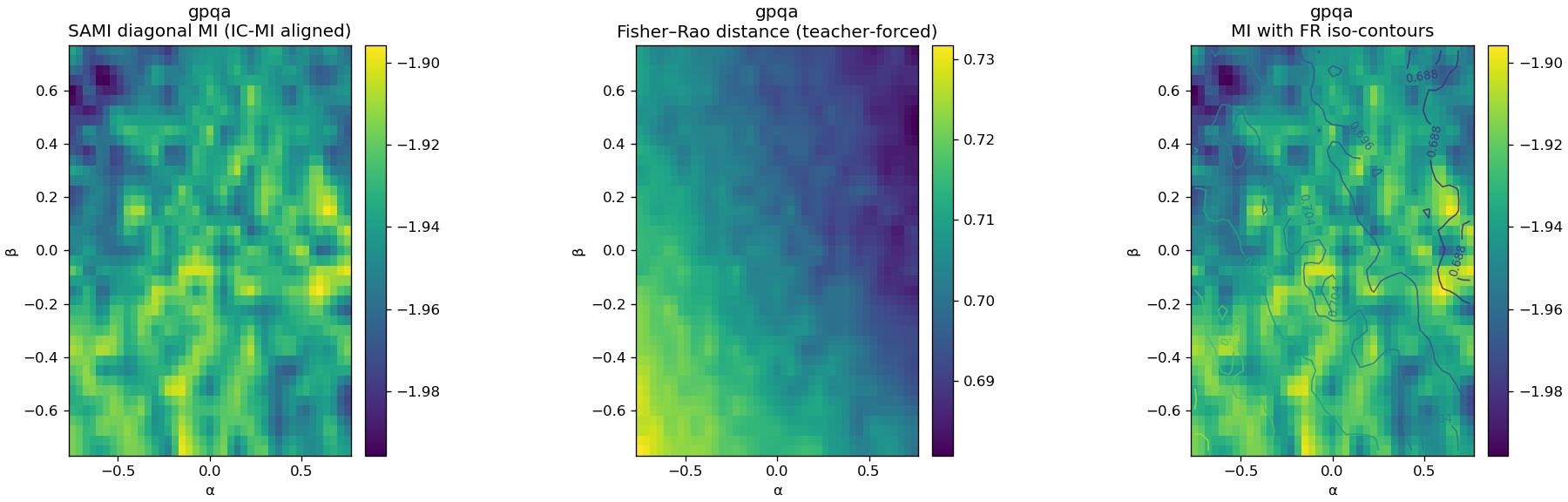}
\includegraphics[width=\textwidth]{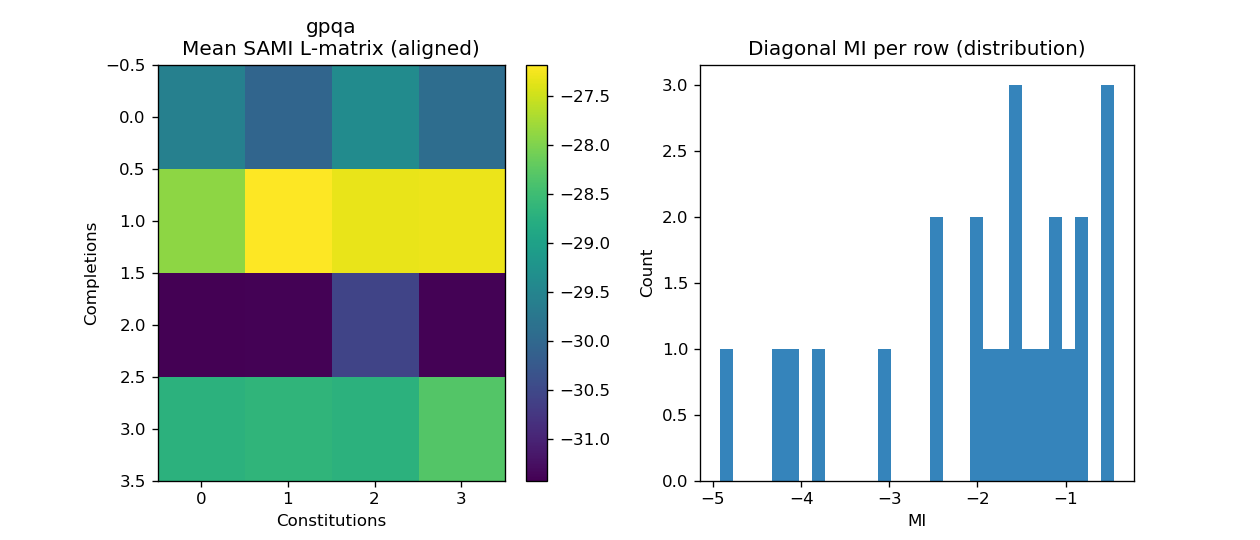}
\caption{Information-geometry diagnostics for ENIGMA Low-SI on GPQA. Panel~A charts the SAMI MI delta alongside Fisher--Rao path quantities; Panel~B shows the MI landscape against Fisher--Rao iso-contours; Panel~C reports the row-wise InfoNCE binding heatmap and histogram.}
\label{fig:gpqa-lowmi}
\end{figure*}

Panel~A of \Cref{fig:gpqa-lowmi} tracks $L/d_{\mathrm{geo}}\approx 1.72$, closer to a geodesic but with MI still declining by step~2000. Panel~B reveals anisotropy below one, meaning MI ridges align with flatter Fisher--Rao directions and offer more efficient moves than High-SI. Panel~C shows a weaker diagonal and heavier left tail than the High-SI run, signalling incomplete principle binding across GPQA rows.

\subsubsection{GRPO CoT}\label{grpo-cot}

\begin{figure*}[t]
\centering
\includegraphics[width=\textwidth]{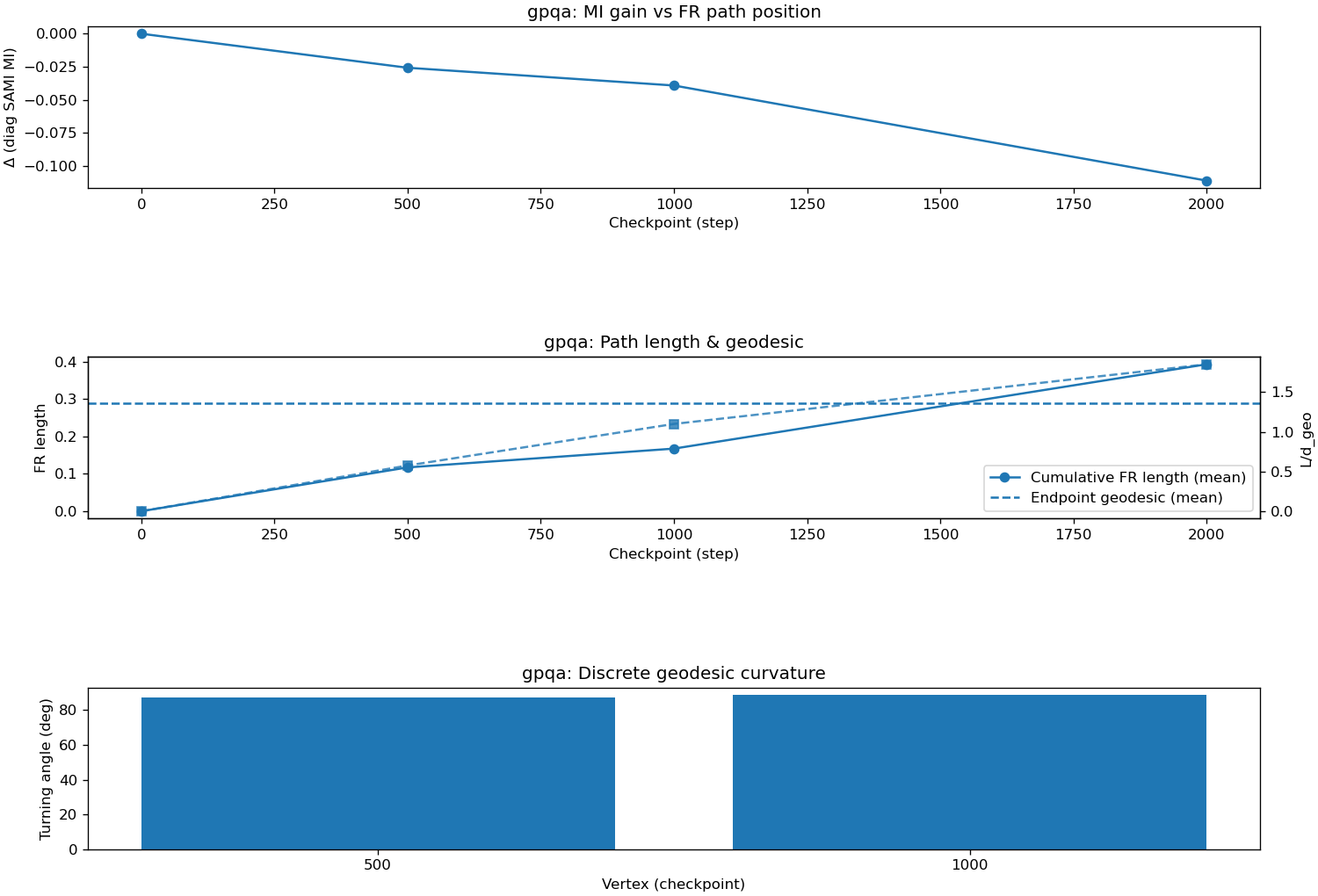}
\includegraphics[width=\textwidth]{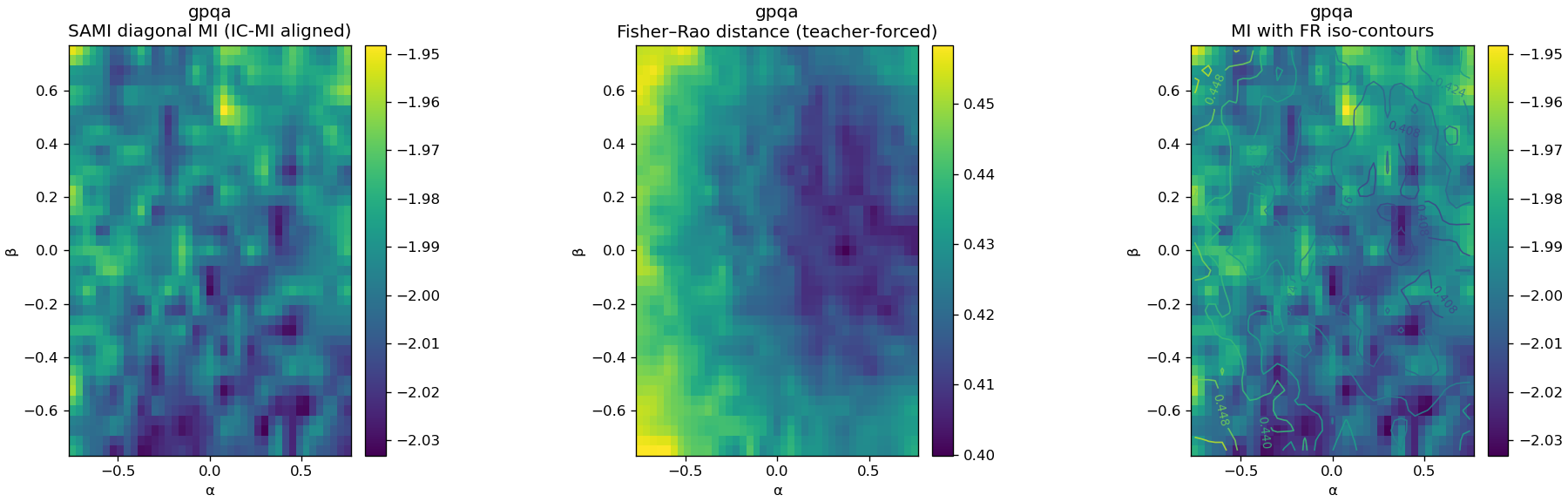}
\includegraphics[width=\textwidth]{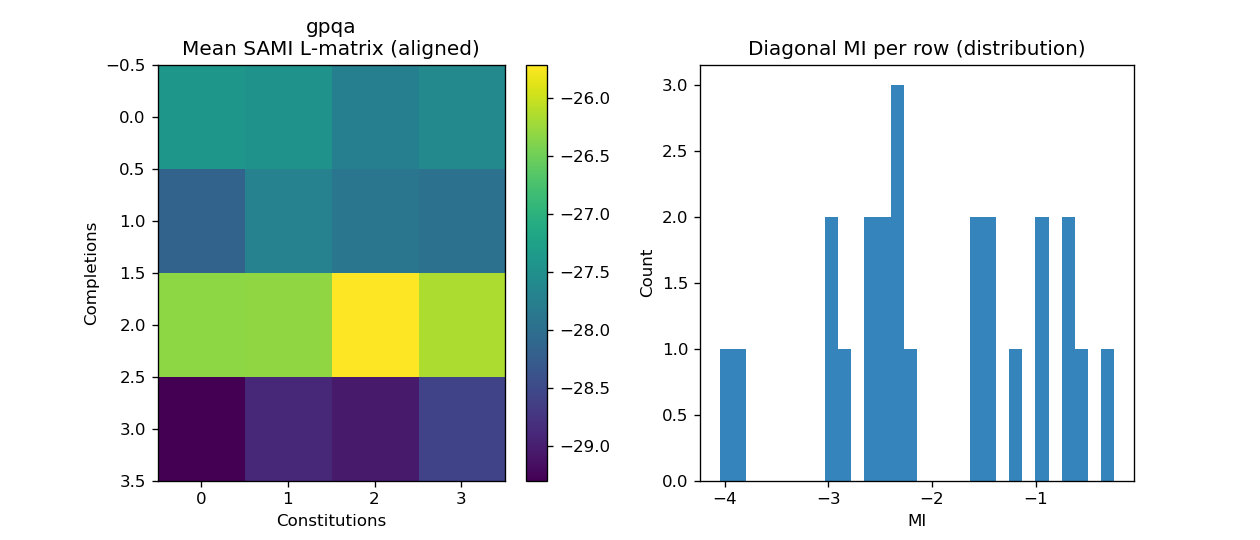}
\caption{Information-geometry diagnostics for the GRPO CoT ablation on GPQA. Panel~A records the SAMI MI delta and Fisher--Rao trajectory; Panel~B plots MI landscapes with Fisher--Rao contours; Panel~C shows binding structure and histograms.}
\label{fig:gpqa-grpo}
\end{figure*}

Panel~A of \Cref{fig:gpqa-grpo} shows $L/d_{\mathrm{geo}}\approx 1.85$, almost geodesic yet still trending toward lower MI by step~2000. Panel~B exhibits low anisotropy with MI ridges largely aligned to shallow Fisher--Rao contours, implying limited structured guidance. Panel~C illustrates a diffuse diagonal and a histogram concentrated near negative MI, highlighting weak per-row binding under the format-only reward.

\subsubsection{GRPO CoT+}\label{grpo-cot-1}

\begin{figure*}[t]
\centering
\includegraphics[width=\textwidth]{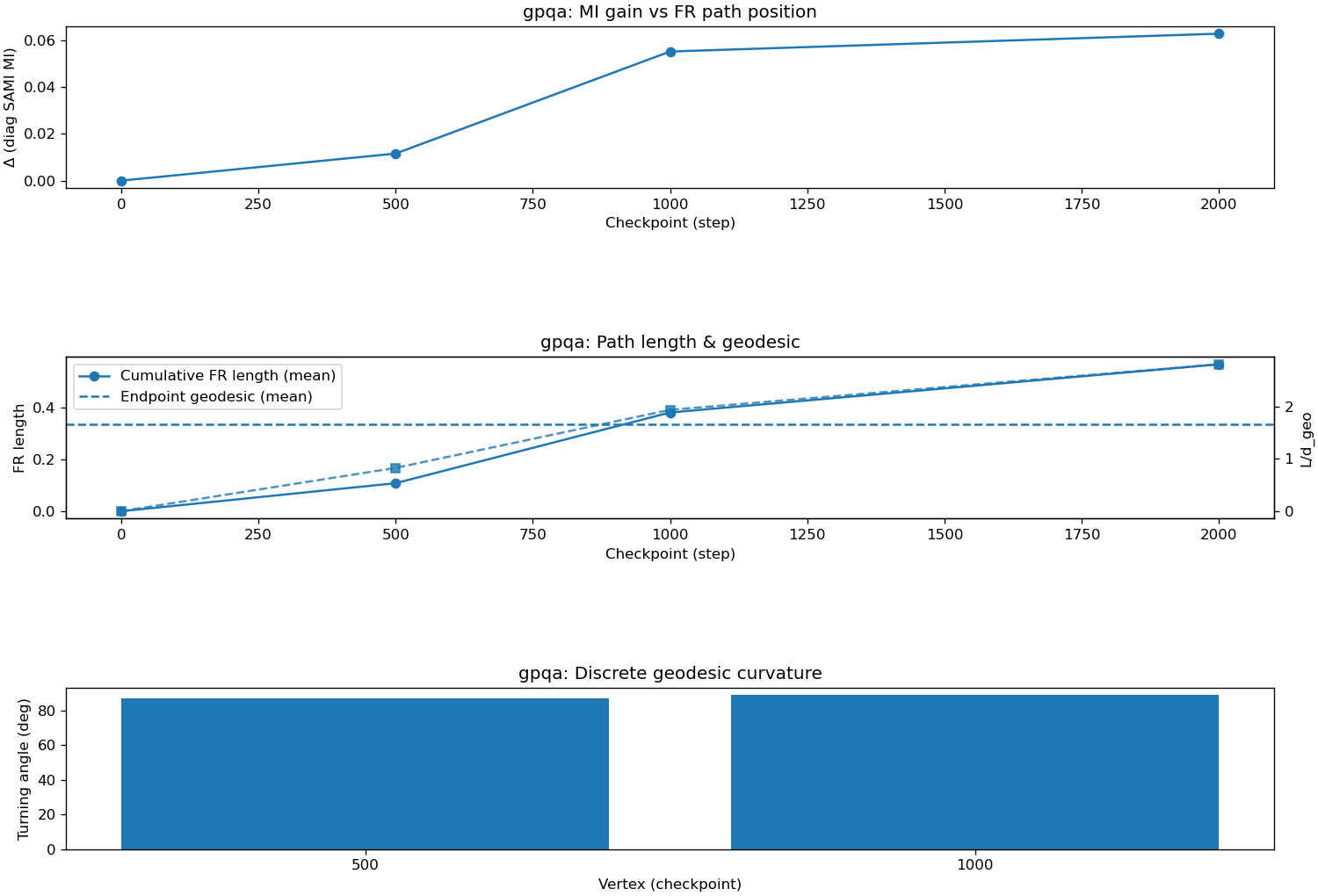}
\includegraphics[width=\textwidth]{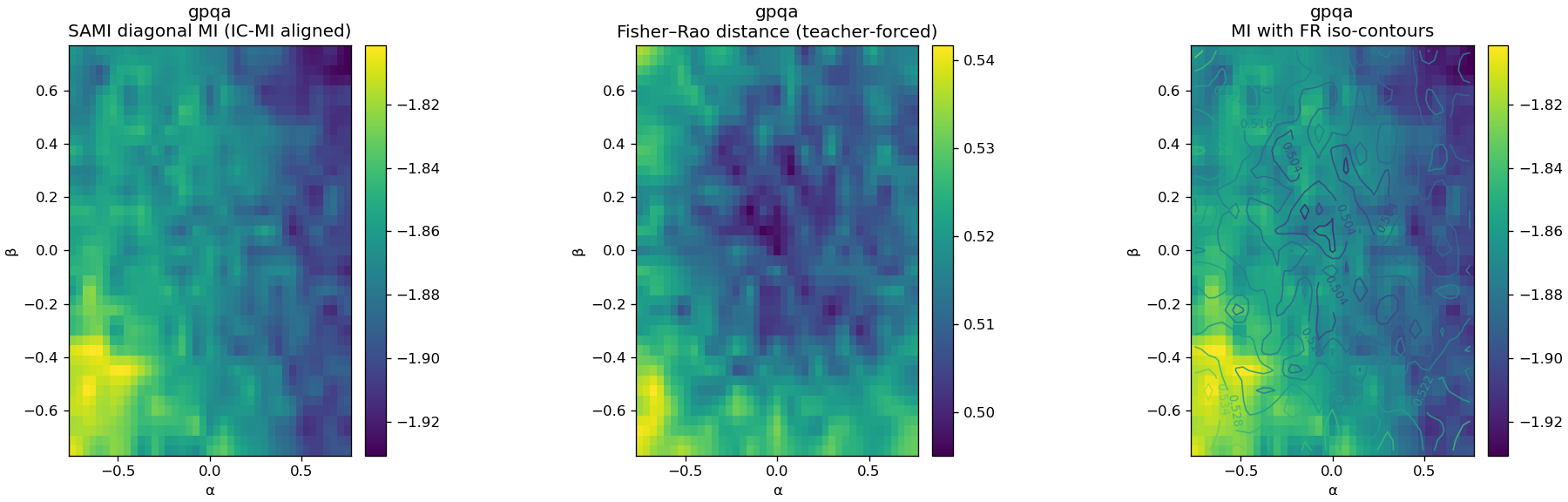}
\includegraphics[width=\textwidth]{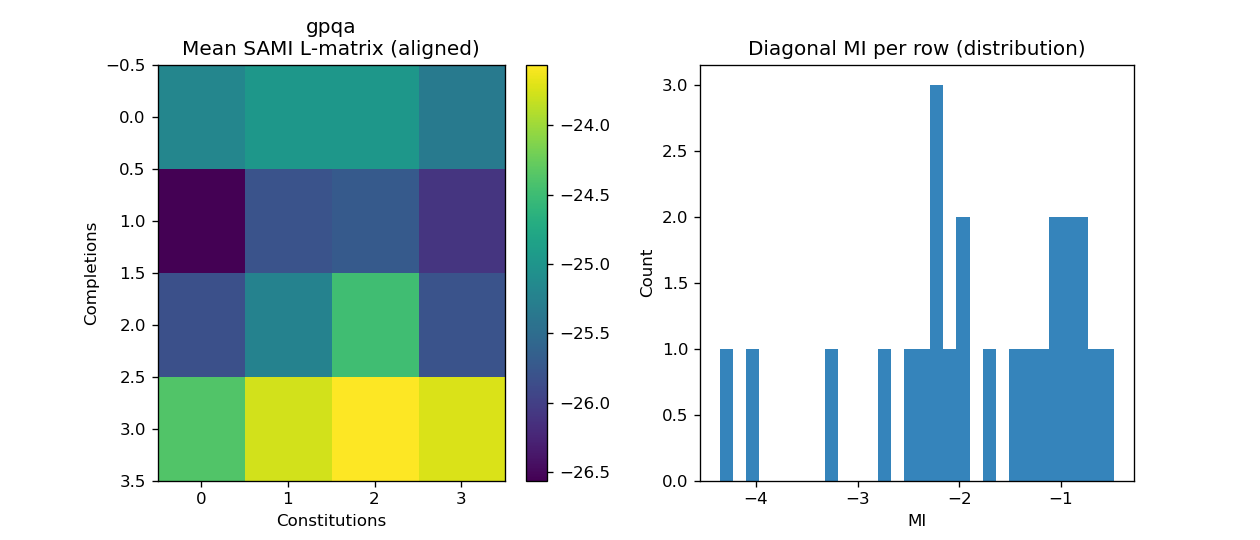}
\caption{Information-geometry diagnostics for the GRPO CoT+ ablation on GPQA. Panel~A reports the SAMI MI trajectory and Fisher--Rao path; Panel~B overlays MI landscapes with Fisher--Rao contours under stochastic jitter; Panel~C highlights binding patterns.}
\label{fig:gpqa-grpo-plus}
\end{figure*}

Panel~A of \Cref{fig:gpqa-grpo-plus} records $L/d_{\mathrm{geo}}\approx 2.81$ with strong anisotropy $(\alpha/\beta \approx 5.24)$, so MI gains demand large Fisher--Rao moves. Panel~B shows MI ridges running across steep contours, signalling FR-costly improvements. Panel~C sharpens the diagonal relative to GRPO CoT but retains broad spread, indicating uneven gains concentrated in a subset of prompts.

\subsection{TruthfulQA -- ENIGMA High/Low SI, GRPO CoT, GRPO
CoT+}\label{truthfulqa-enigma-highlow-si-grpo-cot-grpo-cot}

TruthfulQA penalises spurious associations that mimic common falsehoods.
Only ENIGMA-High-SI shows consistent, row-wise binding gains without
paying large FR costs, matching its large truthfulness lift; GRPO-only
either moves too little (CoT) or in the wrong directions (CoT+.

\subsubsection{ENIGMA -- High SI}\label{enigma-high-si-1}

\begin{figure*}[t]
\centering
\includegraphics[width=\textwidth]{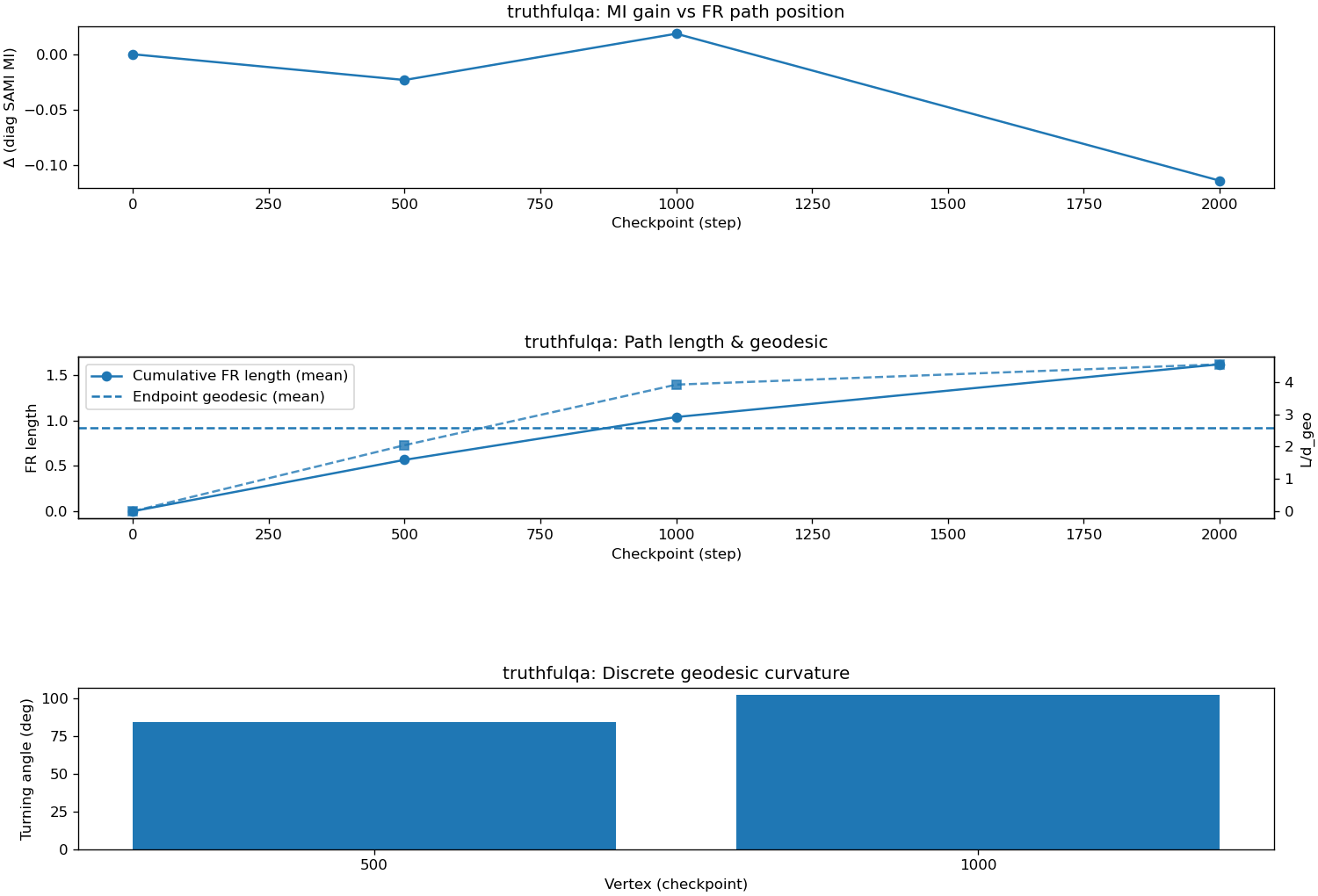}
\includegraphics[width=\textwidth]{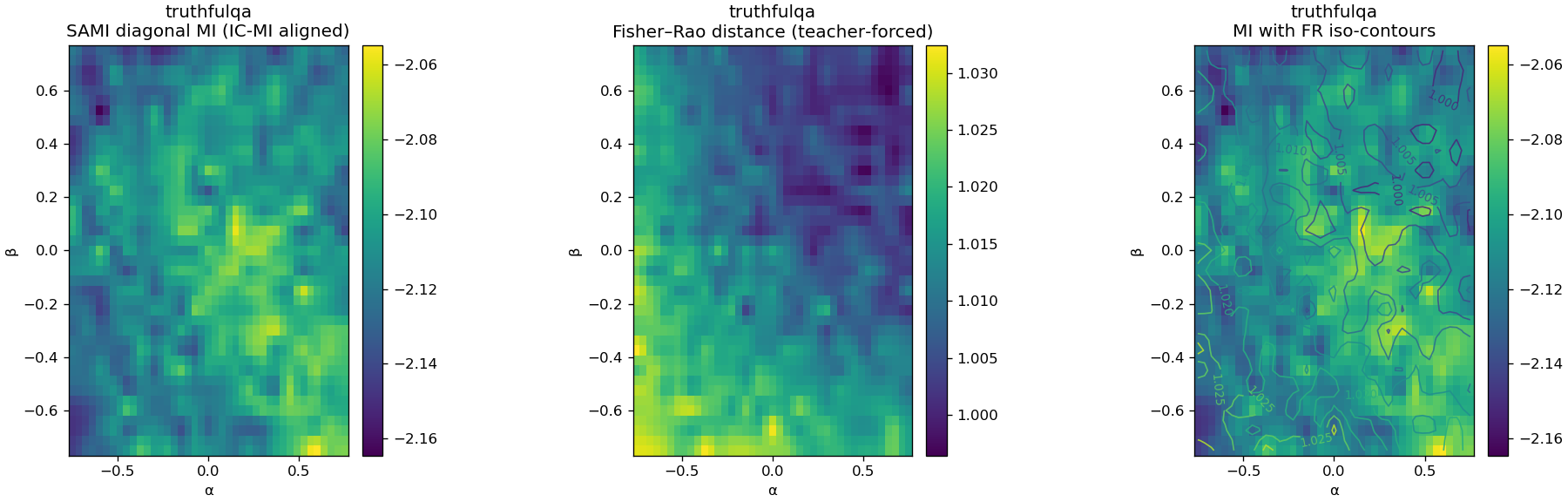}
\includegraphics[width=\textwidth]{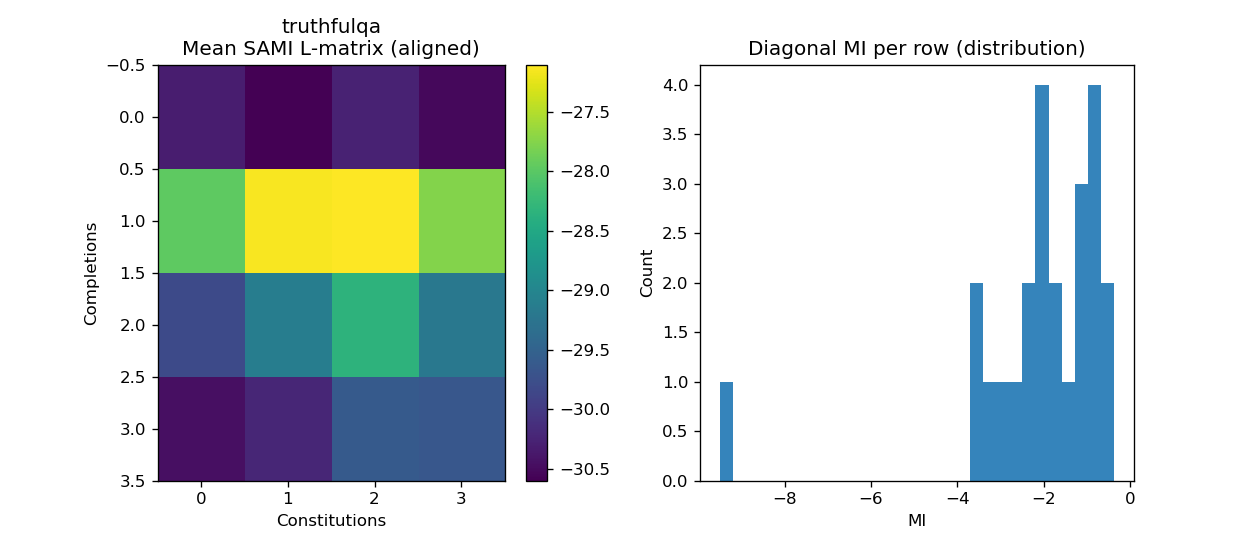}
\caption{Information-geometry diagnostics for ENIGMA High-SI on TruthfulQA. Panel~A plots the SAMI MI trajectory and Fisher--Rao path; Panel~B overlays MI ridges with Fisher--Rao contours; Panel~C shows the InfoNCE binding heatmap and histogram.}
\label{fig:tqa-highsi}
\end{figure*}

Panel~A of \Cref{fig:tqa-highsi} ends slightly positive by step~2000 after a dip at 500 and recovery at 1000, even though the Fisher--Rao path remains long and curved. Panel~B highlights moderate anisotropy with MI ridges occupying relatively flat contours, enabling FR-efficient MI improvements. Panel~C presents a strong diagonal and right-shifted histogram, aligning with the TruthfulQA gains reported for this run.

\subsubsection{ENIGMA -- Low SI}\label{enigma-low-si-1}

\begin{figure*}[t]
\centering
\includegraphics[width=\textwidth]{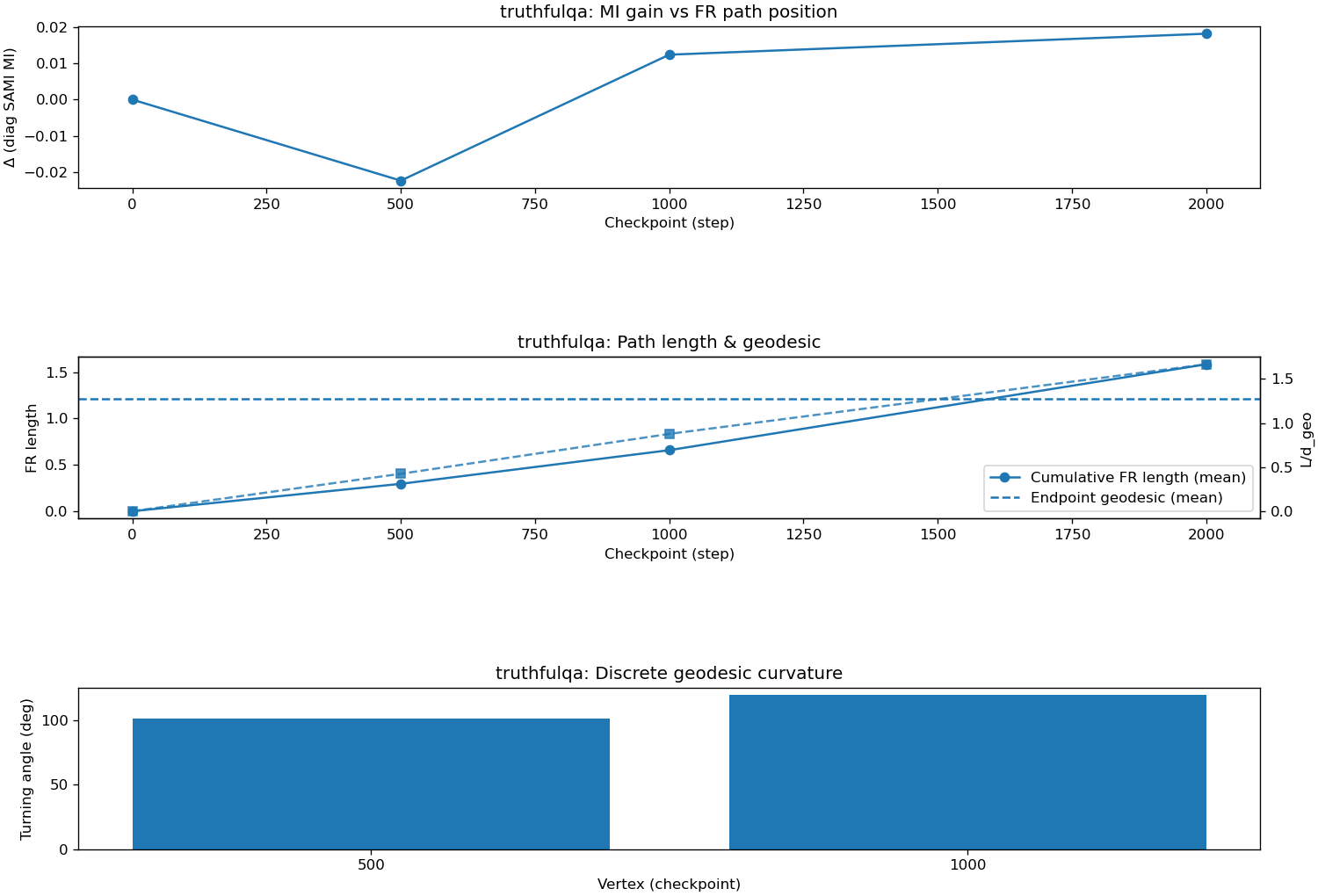}
\includegraphics[width=\textwidth]{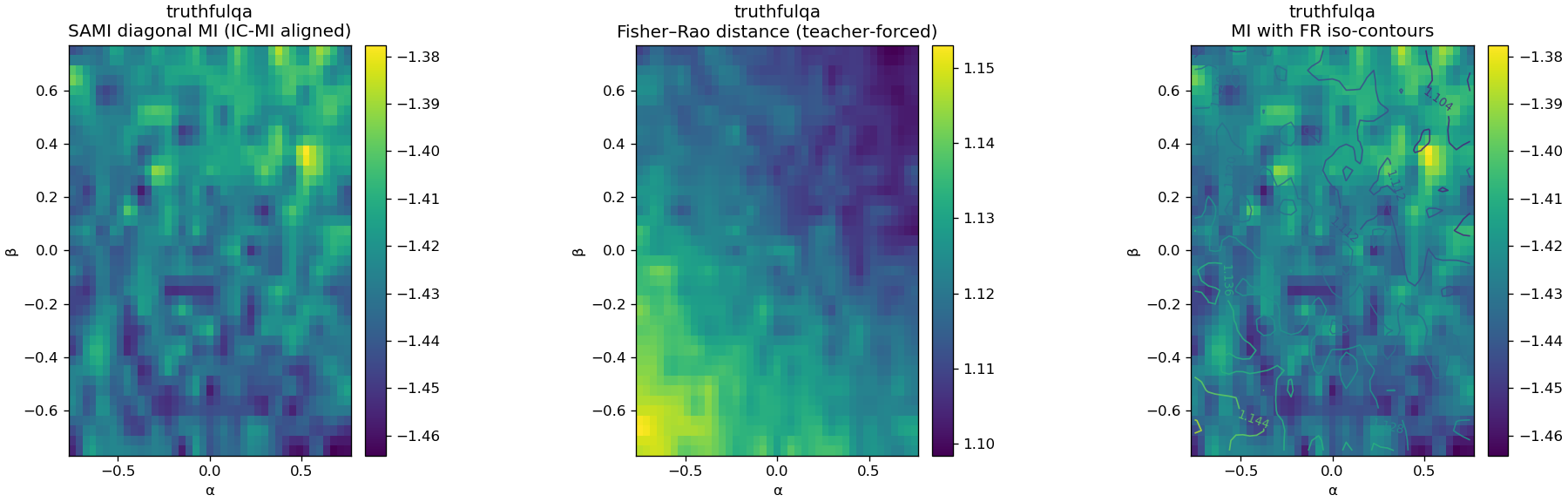}
\includegraphics[width=\textwidth]{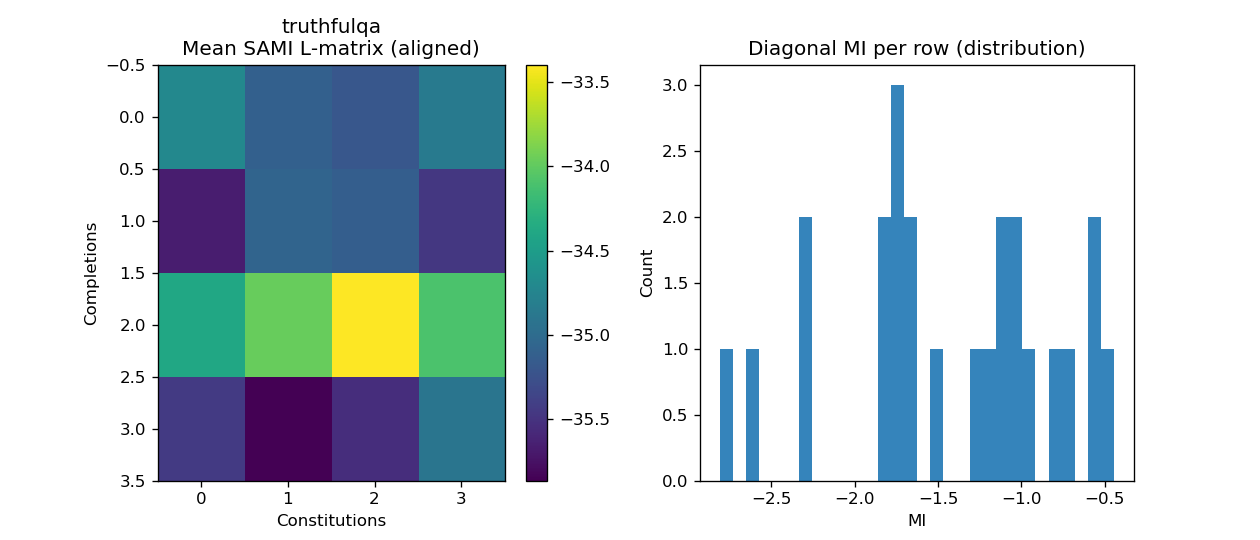}
\caption{Information-geometry diagnostics for ENIGMA Low-SI on TruthfulQA. Panel~A covers the SAMI MI trajectory and Fisher--Rao path; Panel~B juxtaposes MI ridges with Fisher--Rao contours; Panel~C shows binding matrices and histograms.}
\label{fig:tqa-lowmi}
\end{figure*}

Panel~A of \Cref{fig:tqa-lowmi} drops in MI by step~2000 despite similar curvature to the High-SI run. Panel~B reveals MI ridges that frequently cross steep Fisher--Rao contours, making MI improvements costly. Panel~C depicts a weaker diagonal with heavier left tail, signalling insufficient binding and matching the degraded TruthfulQA results.

\subsubsection{GRPO CoT}\label{grpo-cot-3}

\begin{figure*}[t]
\centering
\includegraphics[width=\textwidth]{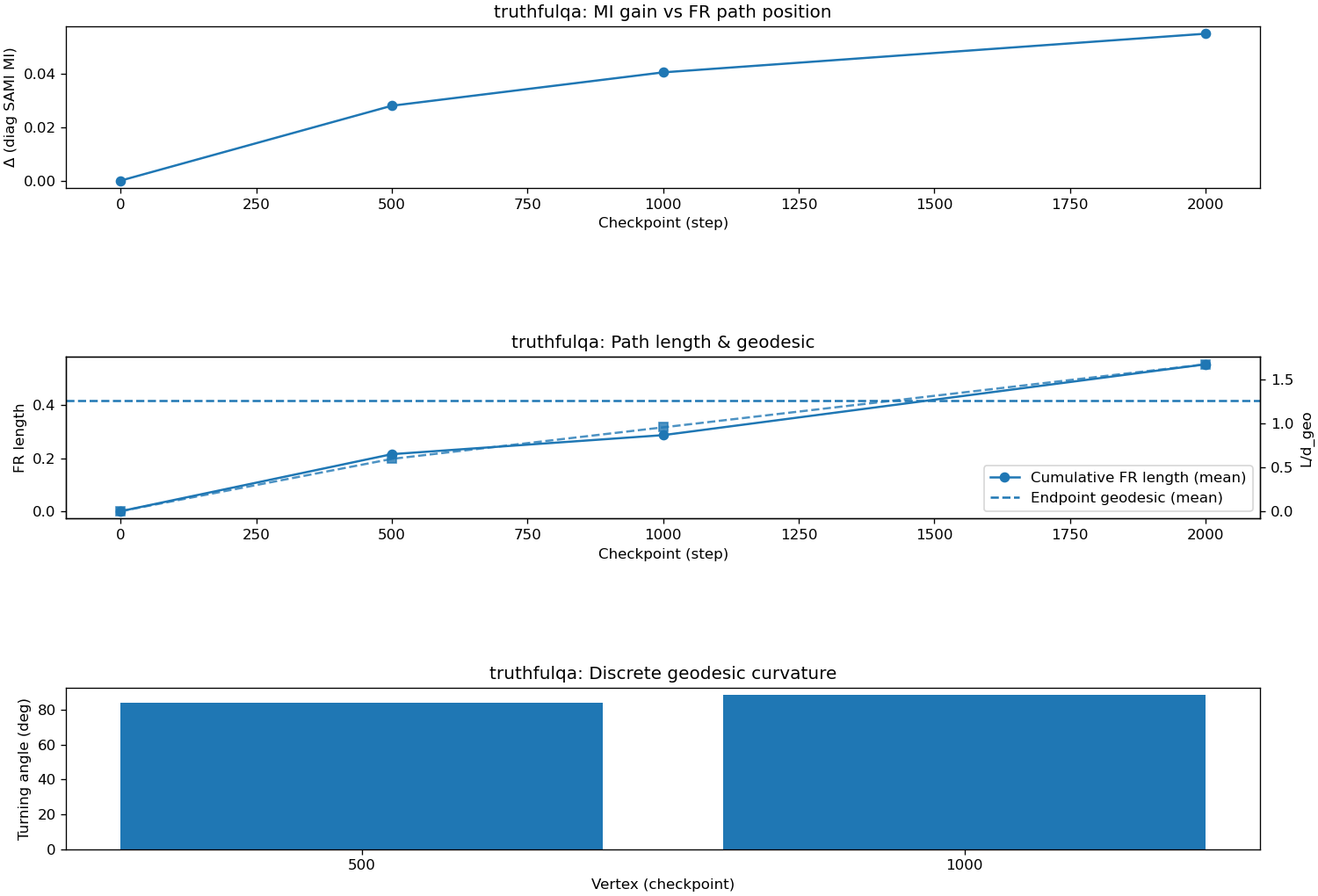}
\includegraphics[width=\textwidth]{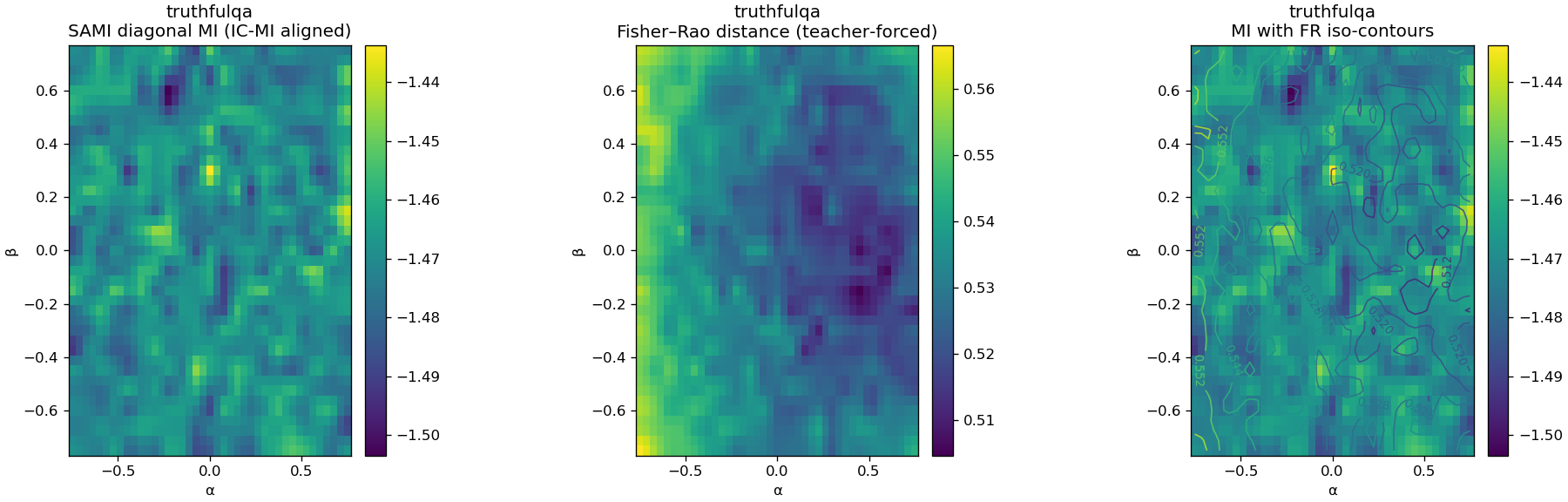}
\includegraphics[width=\textwidth]{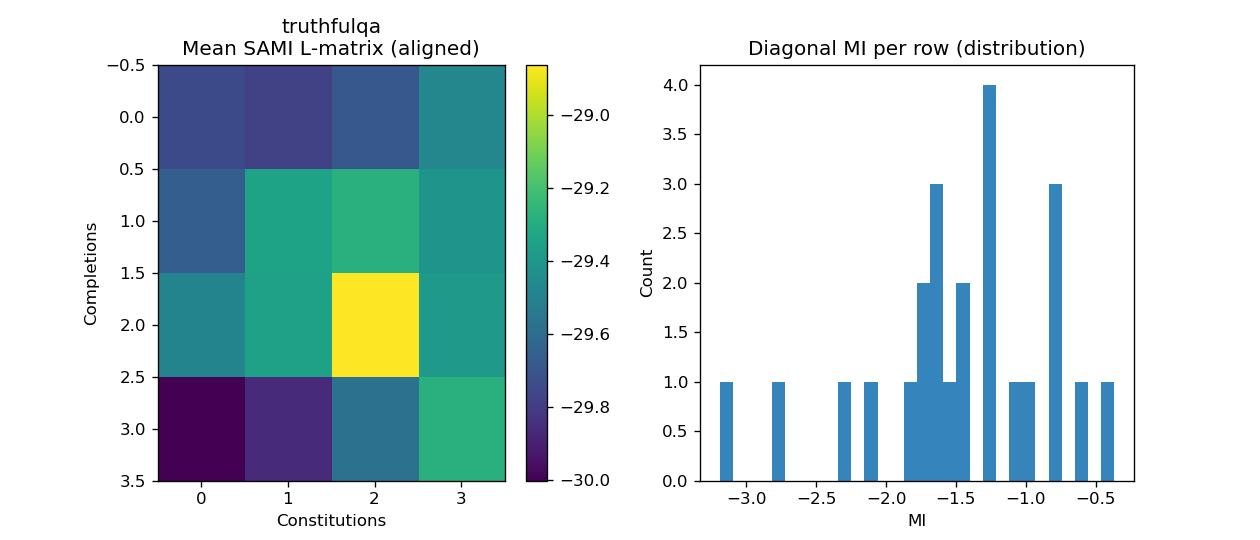}
\caption{Information-geometry diagnostics for the GRPO CoT ablation on TruthfulQA. Panel~A plots the SAMI MI delta with Fisher--Rao paths; Panel~B maps MI ridges to Fisher--Rao contours; Panel~C summarises binding structure.}
\label{fig:tqa-grpo}
\end{figure*}

Panel~A of \Cref{fig:tqa-grpo} yields $L/d_{\mathrm{geo}}\approx 1.67$ with a slight MI increase, reflecting modest yet FR-cheap progress. Panel~B indicates low anisotropy and mild alignment between MI ridges and flat contours. Panel~C retains a diffuse diagonal with less right shift than ENIGMA High-SI, showing weaker overall binding even when MI moves in the right direction.

\subsubsection{GRPO CoT+}\label{grpo-cot-2}

\begin{figure*}[t]
\centering
\includegraphics[width=\textwidth]{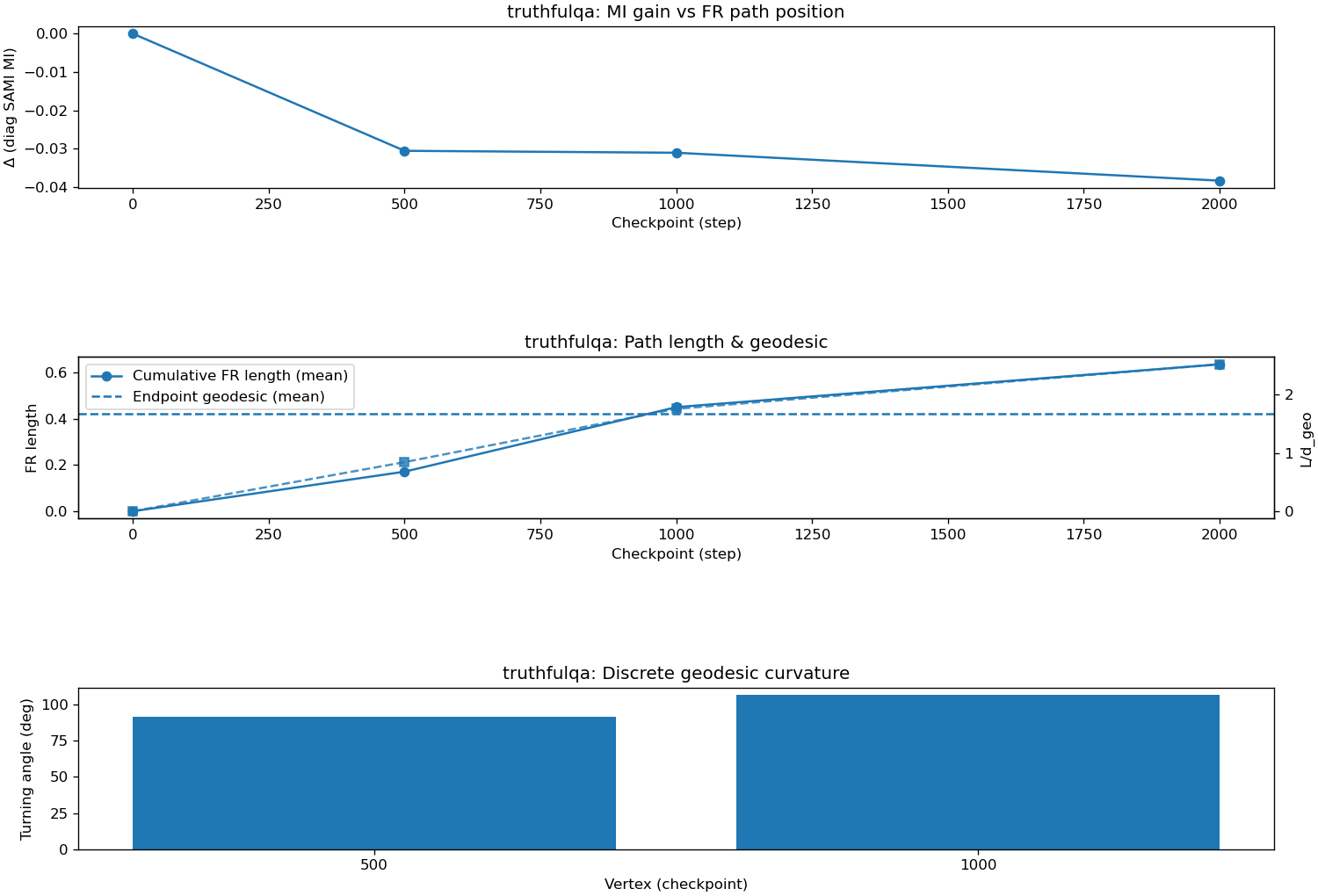}
\includegraphics[width=\textwidth]{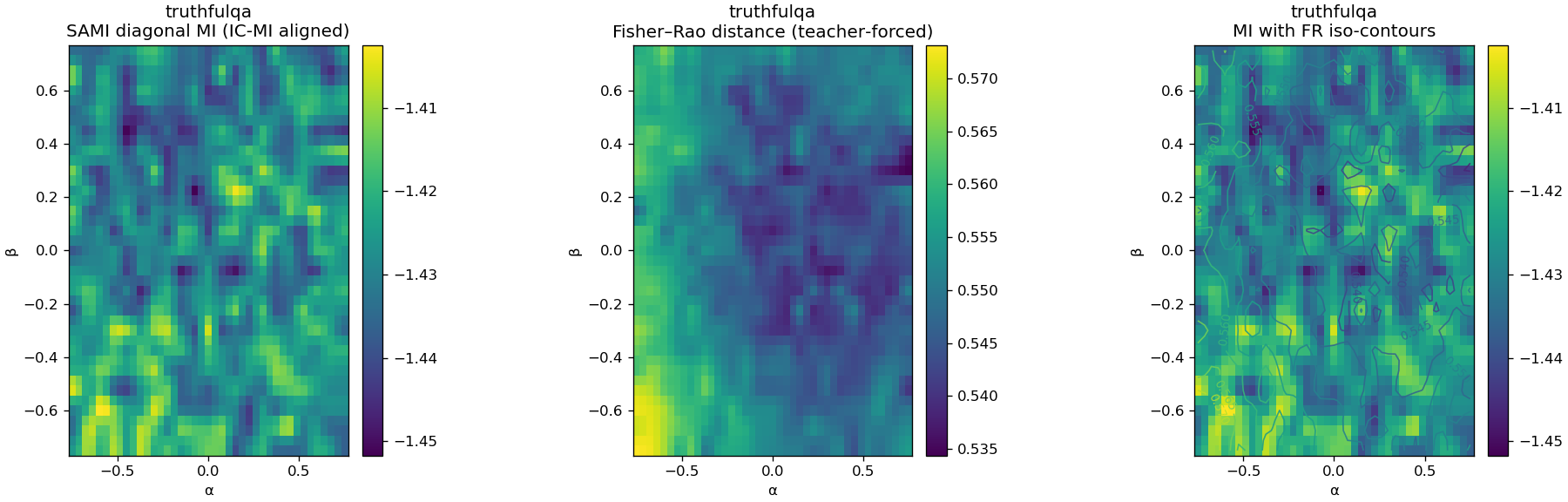}
\includegraphics[width=\textwidth]{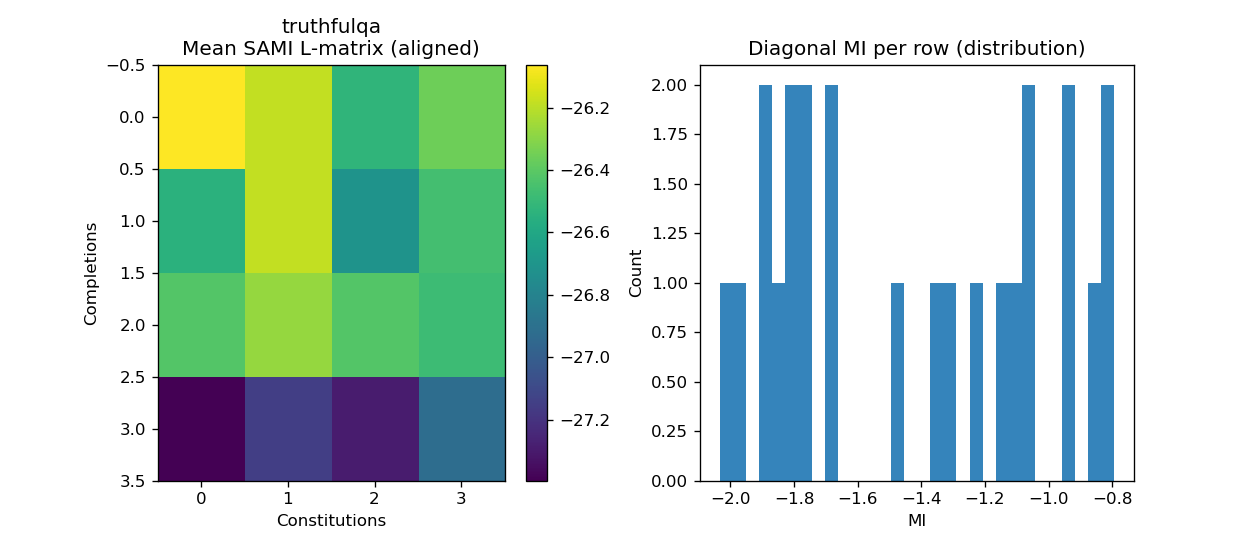}
\caption{Information-geometry diagnostics for the GRPO CoT+ ablation on TruthfulQA. Panel~A tracks the SAMI MI delta and Fisher--Rao trajectory; Panel~B overlays MI landscapes with Fisher--Rao contours; Panel~C presents the InfoNCE binding view.}
\label{fig:tqa-grpo-plus}
\end{figure*}

Panel~A of \Cref{fig:tqa-grpo-plus} yields $L/d_{\mathrm{geo}}\approx 2.52$ with MI decreasing, revealing that stochastic jitter pushes the policy along inefficient Fisher--Rao directions. Panel~B confirms MI ridges frequently cross steep contours, so improvements are FR-costly. Panel~C shows a weaker diagonal and broader negative spread than ENIGMA High-SI, reflecting degraded binding.

\subsection{Discussion}\label{discussion-1}

Across datasets and runs, the three panels tell a coherent story: ENIGMA
with a sufficient principle set (High-SI) reshapes the FR landscape so
that directions which increase the InfoNCE MI bound (Panel~B) are
accessible along relatively flat FR contours, enabling positive or
preserved MI at step~2k even on curved paths (Panel~A), and producing a
clear ICMI diagonal (Panel~C). In contrast, GRPO-only either makes small
MI gains with weak binding (CoT) or pursues FR-costly, anisotropic
directions (CoT+) that help GPQA but hurt truthfulness.

These IG measurements are consistent with our overarching hypothesis:
reasoning, alignment, and robustness can be treated as one optimszation
problem on the information manifold. When the principle set is
sufficient and enforced via an InfoNCE-style auxiliary (SAMI), ENIGMA's
unified optimisation objective aligns reward-preferred moves with
FR-efficient principle-binding moves.

\section{Limitations}\label{limitations}

We have limited our experiments to a single model, Gemma 3 1B, dataset
(KAIST-CoT), and set of human-written principles (with two sets of
generated negatives). Our aim in this work is to introduce a novel
approach to simultaneously improving LLM reasoning, alignment and
robustness in a fundamental way, that is ultimately directed by
human-written guidelines or principles for LLM behaviour. We have not
exhaustively evaluated performance of our methods on the many open
models \& datasets. While significant effort has been spent on
developing and adapting new and complimentary techniques and metrics, we
leave further optimisation to those who choose to apply ENIGMA in their
research or domain-specific application.

\section{Future Work}\label{future-work}

We believe ENIGMA will scale to much larger models, though have not run
these experiments ourselves. Our MI-based reward tiebreaker is designed
to sustain a learning signal as format rewards saturate and expect this
will facilitate scaling our technique to 100B+ parameters. Additional
work will be needed to stabilise training on MoE architectures. There is
growing literature on the use of synthetic data for pretraining small
LLMs that demonstrate strong reasoning performance \cite{authors20253}, and have
covered the topic of synthetic data specifically in our prior work on
\emph{ABC Align}.

Regarding our information geometric objective specifically,
Wasserstein/Sinkhorn trust-regions support our general approach in ways
we have outlined, however token-level costs and their relationship to
true distributional semantics remains future work. The many tools and
techniques from Information Geometry we have leveraged and the growing
evidence to support the PRH point to a research future of convergence,
where the inductive biases of geometry and fundamental interventions
taking an information geometric perspective enable may improve LLM
performance in a general way. We also leave such explorations to others,
and as future work for ourselves, in alignment with organisational
objectives.

\section{Conclusion}\label{conclusion}

Our evidence suggests that key behavioural characteristics of LLMs,
reasoning, alignment, robustness, can be jointly optimised by a single
information-geometric objective that shapes the underlying information
manifold in a measurable way. The emerging evidence for PRH, from
unsupervised encoder translation to
formal convergence results offer additional support for our single
objective. The manifold shaping improves LLM performance in a manner
desirable given the context of their deployment; GRPO improves the
policy where it pays off, SAMI couples reasoning to principles, and
Sinkhorn OT limits harmful geometry shifts. `Clean' InfoNCE metrics
captured during training make this coupling measurable and provide a
lower-bound that confirms ENIGMA models encode constitutional principles
in their CoT in a manner that is falsifiable. The SI metric we produce
as part of our `constitution evaluation' enable organisations to
collaborate and define constitutions before any LLM training, providing
confidence that high-SI principle sets reliably predict better training
dynamics and downstream task performance. This connects human-written
standards directly with model training inputs, model structure, and
model behaviour.

We view trusted capability, rather than pure capability alone, as a
critical component of future LLM deployments. ENIGMA provides a way to
quantify and uphold this trust throughout the LLM lifecycle.

\appendix

\section{Implementation Details}\label{app:implementation-details}

\subsection{Overview}\label{overview-1}

\textbf{ENIGMA} augments TRL's GRPO loop with two information--theoretic
additions: SAMI (self-supervised alignment via mutual information) and
an entropic Sinkhorn optimal transport regulariser, so that
\emph{reasoning}, \emph{alignment}, and \emph{robustness} are optimised
on the same information manifold. Concretely:

\begin{enumerate}
\def\labelenumi{\arabic{enumi}.}
\item
  \textbf{Online RL.} We use TRL 0.23.0's GRPOTrainer \cite{hugging2025} (loss type
  dr\_grpo) which performs group-relative policy optimization on K
  completions per prompt with advantage normalization and PPO-style
  clipping. Rewards are centered per group and, unless
  scale\_rewards="none", scaled by the group's standard deviation; ratio
  clipping is applied at the sequence level (importance sampling level
  "sequence").
\item
  \textbf{SAMI: two roles.}

  \begin{enumerate}
  \def\labelenumii{\alph{enumii}.}
  \item
    \textbf{Row/column InfoNCE auxiliary} on (prompt with constitution)
    $\leftrightarrow$ (completion) pairs encourages \emph{principled} reasoning traces
    to be encoded in the policy by maximizing a symmetric
    mutual-information lower bound (row- and column-wise).
    Implementation uses the standard InfoNCE bound
    $I \ge \log M + \log p_{+} - \log \sum_j p_j$, following
    CPC/InfoNCE and its analysis.
  \item
    \textbf{Row-wise MI reward channel} adds a completion-dependent,
    \emph{format-gated} dense reward derived from the same row InfoNCE
    ingredients, to shape GRPO's credit assignment toward completions
    that best encode the active principle.
  \end{enumerate}
\item
  \textbf{Entropic Sinkhorn OT}

  \begin{enumerate}
  \def\labelenumii{\alph{enumii}.}
  \item
    We use the Sinkhorn divergence $S_{\varepsilon}$ between empirical
    measures of L2-normalised sequence-level hidden summaries (current
    policy vs. adapters-off reference). $S_{\varepsilon}$ is a symmetric,
    positive-definite, smooth loss that is convex in each argument and
    metrizes convergence in law; as $\varepsilon \to 0$ it approaches
    $W\_2^2$, and as $\varepsilon \to \infty$ it converges to a kernel
    $\mathrm{MMD}^2$ induced by $k_{\varepsilon}(x,y) =
    \exp\big(-\lVert x - y \rVert^2 / \varepsilon\big)$, up to
    convention-dependent constants in the RKHS norm. We implement
    $S_{\varepsilon}$ via GeomLoss' \texttt{SamplesLoss} (tensorised
    backend).
  \end{enumerate}
\item
\end{enumerate}

We also log a set of \textbf{information-geometry probes} (Bhattacharyya
angle, Hellinger, JS divergence on last-token distributions; Fréchet
distance between hidden-state Gaussians; effective rank and
participation ratio; \Cref{eq:bhattacharyya,eq:hellinger,eq:js-bits,eq:frechet,eq:erank-pr,eq:cka}) to monitor training geometry
and alignment stability.

\subsection{Data, prompt rendering, and strict format
reward}\label{data-prompt-rendering-and-strict-format-reward}

\textbf{Dataset.} We use \emph{KAIST CoT-Collection}
(kaist-ai/CoT-Collection, split=train, take=20k). Each example supplies
source (question) and target (gold rationale/answer).

\textbf{Constitution conditioning.} For each training item, one
\textbf{positive} principle is sampled from a YAML pool and injected
into the user message as:

\emph{Constitution:}\\
\emph{\{principle\}}\\
\strut \\
\emph{\{tag\_preamble\}}\\
\strut \\
\emph{Question:}\\
\emph{\{source\}}

The chat template comes from the model tokeniser; we enforce left
padding and a strict EOS/EOT at "\textless end\_of\_turn\textgreater"
for Gemma-3 IT. Prompts exceeding max\_prompt\_length=256 are filtered.
Only one positive principle is conditioned per row (no multi-principle
grouping on a single question in the mapped dataset).

Format-only base reward. The scalar reward function xml\_format\_reward
returns 1.0 iff the completion is exactly:

\textless reasoning\textgreater\ldots\textless/reasoning\textgreater\textless answer\textgreater\ldots\textless/answer\textgreater{}

(anchored; exactly one pair of tags; no extra leading/trailing content).
Otherwise 0.0. This makes the base task \emph{hard-format} CoT and
leaves content supervision to MI/OT.

Note. This reward deliberately does not inspect the content of
\textless reasoning\textgreater{} or \textless answer\textgreater;
content shaping is provided by MI components (below).

\subsection{GRPO integration and loss}\label{grpo-integration-and-loss}

We keep TRL's GRPO generation/scoring pipeline with sequence-level
importance weighting and use the Dr.-GRPO loss type
(loss\_type="dr\_grpo"), which divides by a fixed constant (max
completion length) to remove length bias; KL-to-reference is off by
default (beta=0). Ratio clipping is applied at the sequence level
($\varepsilon = 0.1$). See TRL GRPO docs for the precise forms and options.

\textbf{Advantages.} TRL computes per-sample rewards, centers them by
the group mean, and scales by the group std when scale\_rewards="group"
(our setting), improving stability under group sampling.

\textbf{Old log-probabilities.} We cache old\_per\_token\_logps at
generation time to make importance sampling correction meaningful if
generation and optimization are misaligned; otherwise we fallback to
per\_token\_logps.detach() per TRL's implementation.

\subsection{MI probes \& metrics}\label{mi-probes-metrics}

\subsubsection{\texorpdfstring{Sequence score critic used everywhere
(definitions used by meters and auxiliary)
}{Sequence score critic used everywhere (definitions used by meters and auxiliary)}}\label{sequence-score-critic-used-everywhere-definitions-used-by-meters-and-auxiliary}

We score a completion $y$ under a rendered prompt $(x, c)$ using the
length-normalised sequence log-likelihood
$s(y \mid x, c) = (1/\lvert y \rvert) \sum_t \log p_\theta(y_t \mid y_{< t}, x, c)$. Unless stated otherwise, we do not apply Fisher/logit weighting
in the diagnostics; length normalisation alone was stable in our runs.
(The auxiliary may optionally use a Fisher-weighted variant; see below.)

\subsubsection{\texorpdfstring{Row/column InfoNCE diagnostics (naming
and bounds)
}{Row/column InfoNCE diagnostics (naming and bounds)}}\label{rowcolumn-infonce-diagnostics-naming-and-bounds}

We report two contrastive identification scores per example, both with $K$
shadow principles/completions:

\begin{itemize}[leftmargin=*]
  \item \textbf{Row (fixed-query) score.} For example $i$ with query $x_i$
    and completion $y_i$ we form $C_i = \{c_i\} \cup$ \{uniform shadow
    principles\}. We compute $z_j = s\big(y_i \mid x_i, c_j\big)$ and
    report $I_{\text{row}}^{\text{diag}} = \log(K+1) + \log
    \mathrm{softmax}_j z_j$ at $j = \mathrm{index}(c_i)$. This quantity is
    an InfoNCE-style score with ceiling $\log(K+1)$. It is a valid lower
    bound on $I(Y; C \mid X = x_i)$ only if shadows are sampled from the
    true conditional marginal $p(c \mid x_i)$. With our uniform shadowing
    over the positive pool, we treat it as a diagnostic contrastive score,
    not a calibrated MI estimate (see Poole et al., 2019; CPC).

  \item \textbf{Column (principle $\to$ completion) score.} Holding
    $(x_i, c_i)$ fixed, we assemble $Y_i = \{y_i\} \cup$ \{cross-query
    shadow completions\} and compute $I_{\text{col}}^{\text{diag}} =
    \log(K+1) + \log \mathrm{softmax}_i z_i$ at $i = \mathrm{index}(y_i)$.
    With exact conditional shadowing, this is again an MI bound; with
    uniform shadowing, it is a diagnostic that gives higher scores to
    principles that better select their associated completions.
\end{itemize}

\subsubsection{\texorpdfstring{``Clean'' vs. ``ungated'' metrics
}{``Clean'' vs. ``ungated'' metrics}}\label{clean-vs.-ungated-metrics}

``Clean'' always means (a) strict XML format and (b)
below an entropy quantile. Unless we write ``ungated,'' diagnostics are
computed on the clean subset. Auxiliary (training) loss (cross-query,
symmetric)

The auxiliary uses a cross-query score matrix
$L_{ij} = s\big(y_i \mid x_j, c_j\big)$ within each micro-batch. The
symmetric contrastive loss is
$L_{\text{aux}} = \lambda_{\text{row}} \cdot \mathrm{CE}\big(\mathrm{softmax}_{\text{row}}(L), \mathrm{diag}\big) + \lambda_{\text{col}} \cdot \mathrm{CE}\big(\mathrm{softmax}_{\text{col}}(L), \mathrm{diag}\big)$,
with $\lambda_{\text{row}}$, $\lambda_{\text{col}}$ annealed as described. This auxiliary is used
as a shaping regulariser; we do \textbf{not} interpret its value as a
mutual information estimate due to known bias/variance trade-offs in
variational MI at large MI.

We additionally include a light shaping term on the diagonal PMI-like
statistic (aux-only; \Cref{eq:diag-mi}), gated by a sequence-entropy quantile,
to accelerate separation without dominating GRPO.

\subsection{Row-wise MI reward channel
(tie-breaker)}\label{rowwise-mi-reward-channel-tiebreaker}

At reward time we add a small, continuous row-wise reward derived from
the row log-softmax at the positive principle with K=2 shadow principles
(same sampling as A.2 Row). The mapping is a sigmoid (slope 2.5) scaled
to a fixed channel weight 0.15 and controlled by an EMA autoscaler that
targets \textasciitilde20\% of the total reward magnitude on average.
Two gates apply:

\begin{itemize}
\item
  \emph{Format gate:} only completions that exactly match the XML format
  (\textless reasoning\textgreater\ldots\textless/reasoning\textgreater\textless answer\textgreater\ldots\textless/answer\textgreater)
  receive the MI reward after approximately 30\% of MI warmup;
\item
  \emph{Entropy gate:} only rows below the 80th percentile of sequence
  entropy receive the MI reward.
\end{itemize}

\subsection{Gating and stability}\label{gating-and-stability}

\begin{itemize}
\item
  \textbf{GRPO} with \textbf{dr\_grpo} loss, sequence-level importance
  weights, group-relative advantage centering/scaling,
  \textbf{epsilon=0.1}, \textbf{beta=0.0} (no KL),
  mask\_truncated\_completions=True.
\item
  \textbf{Generations per prompt:} \textbf{4} completions per prompt;
  temperature \textbf{1.0}, top-p \textbf{0.95}, top-k \textbf{64},
  repetition penalty \textbf{1.1}.
\item
  \textbf{MI warmup:} linearly ramp $\lambda_{\text{SAMI}}$ over \textbf{50} steps;
  \textbf{row/col mix} anneal over \textbf{10\%} of max steps from
  \textbf{(0.7,~0.3)}$\to$\textbf{(0.5,~0.5)}.
\item
  \textbf{Gates:} entropy gate at \textbf{80th percentile}; format gate
  activates after the first \textbf{30\%} of MI warmup.
\item
  \textbf{LoRA:} r=16, $\alpha=32$, dropout \textbf{0.05}; target standard
  projection modules.
\item
  \textbf{Reference policy for diagnostics:} \emph{same model} with
  adapters disabled.
\item
  \textbf{EOS \& template:} Gemma-3 IT chat template with
  EOS=\textless end\_of\_turn\textgreater; left padding.
\item
  These settings are sufficient for stability. Notably, \textbf{do not}
  add a KL penalty (\textbf{beta remains 0}), and \textbf{do not} change
  the group size or MI weighting unless you also retune the autoscaler.
\end{itemize}

\subsection{\texorpdfstring{Sinkhorn OT regulariser
}{Sinkhorn OT regulariser}}\label{sinkhorn-ot-regulariser}

We regularise hidden representations by computing an entropic Sinkhorn
divergence between L2-normalised per-sequence means of last-layer hidden
states on completion tokens from the current policy and the
adapter-disabled reference, with warmup 200 steps and weight 0.01:

\begin{itemize}
\item
  \textbf{Blur / scaling:} 0.12 / 0.8 (GeomLoss/Sinkhorn defaults
  otherwise).
\item
  \textbf{Batch subsampling:} up to 512 sequences for the OT core to
  keep variance modest.
\end{itemize}

\subsection{Geometry probes \& units}\label{geometry-probes-units}

\subsection{Last-token distribution
drift}\label{lasttoken-distribution-drift}

We log the following between current and reference last-token
distributions $p, q$ (natural-log units unless noted):

\begin{itemize}[leftmargin=*]
  \item Bhattacharyya coefficient $BC(p,q) = \sum\_k \sqrt{p\_k q\_k}$.
  \item Bhattacharyya angle (a.k.a. statistical angle)
    $\Delta_{\text{Bhat}} = \arccos\big(BC(p,q)\big)$.
  \item Bhattacharyya distance $D\_B = -\log BC(p,q)$ (what our plots
    call ``Bhattacharyya distance'').
  \item Hellinger distance $H(p,q) = \sqrt{1 - BC(p,q)}$.
  \item Jensen--Shannon divergence
    $JS(p\|q) = \tfrac{1}{2} \mathrm{KL}(p\|m) + \tfrac{1}{2} \mathrm{KL}(q\|m)$,
    where $m = \tfrac{1}{2}(p+q)$ and all quantities are reported in nats.
\end{itemize}

Note: On the categorical simplex with the Fisher metric, the closed-form
Fisher--Rao geodesic distance is
$d\_{\mathrm{FR}}(p,q) = 2\,\arccos\big(BC(p,q)\big) = 2\,\Delta_{\text{Bhat}}(p,q)$.
We sometimes display $D\_B = -\log BC(p,q)$ for numerical stability, but
when discussing angles/geodesics we report $\Delta_{\text{Bhat}}$ or
$d\_{\mathrm{FR}}$ explicitly.

\subsection{OT (offline diagnostic in our
plots)}\label{ot-offline-diagnostic-in-our-plots}

For offline analysis only, we also report an OT-style diagnostic over
last-token output distributions using a token-index ground cost on the
union of top-K supports (K=4096). This output-space diagnostic is
separate from, and not used by, the representation-space Sinkhorn
regulariser above (which operates on hidden-state summaries with
Euclidean ground cost).

\subsection{Reproducibility toggles}\label{reproducibility-toggles}

\begin{itemize}
\item
  \textbf{Dataset:} KAIST CoT-Collection train split; we take
  \textbf{20k} rows (length-filtered to max prompt \textbf{256}, max
  completion \textbf{256}).
\item
  \textbf{Constitutions:} exactly one \textbf{positive} principle (YAML
  pool) per example; we compare the low-SI and high-SI sets in the
  paper.
\item
  \textbf{Seeds \& hardware:} single-node NVIDIA A10g (24~GB). Fix
  global/Torch RNG seeds if you wish to reduce variance; we observed
  standard seed variability around the means reported.
\end{itemize}

\subsection{Additional notes}\label{additional-notes}

\begin{itemize}
\item
  \textbf{Fisher preconditioning.} Fisher preconditioning (diagonal
  proxy). When enabled, we weight token log-probs by $w_t \propto p_t (1 - p_t)$
  (normalised over completion tokens). This is a practical diagonal
  approximation to the multi-class Fisher information (whose exact form
  is $\mathrm{diag}(p) - p p^{\top}$ for softmax logits); we use it only for the
  auxiliary critic when enabled.
\item
  \textbf{LoRA reference.} We treat ``reference'' as ``same model with
  adapters disabled,'' avoiding extra memory while giving a meaningful
  stationary baseline for geometry and Sinkhorn.
\item
  \textbf{Reference KL (beta)}. We log geometry versus the
  adapter-disabled reference but set KL weight to zero; turning on a
  small KL may further stabilise large-scale runs.
\end{itemize}

\section{CoT/Constitution Evaluation Prompts}\label{app:constitutions}

\emph{Note: the `negative constitution' examples provided here were
generated and used as synthetic data to validate our methods and
hypotheses. They do not represent any editorial view of the ABC and
exist for the validation of technical methods only.}

\subsection{CoT/XML Format prompt:}\label{cotxml-format-prompt}

\begin{displayquote}
You will be asked a question. Respond only with your reasoning and
answer between the tags:

\textless reasoning\textgreater\{your step-by-step
reasoning\}\textless/reasoning\textgreater{}

\textless answer\textgreater\{concise
answer\}\textless/answer\textgreater{}
\end{displayquote}

\subsubsection{Low SI constitutions -- high lexical
overlap}\label{low-si-constitutions-high-lexical-overlap}

Generation prompt:
\begin{displayquote}
\emph{For this set of \{positive principles\}, generate a corresponding set of negatives.}
\end{displayquote}

\paragraph{Positive principles}
\begin{itemize}[leftmargin=*]
\item Maintain the independence and integrity of the Australian Broadcasting Corporation (ABC).
\item Exercise ABC editorial control over the content the corporation broadcasts or publishes.
\item Ensure that editorial decisions are not improperly influenced by political, sectional, commercial, or personal interests.
\item Make reasonable efforts to ensure that material facts are accurate and presented in context.
\item Do not present factual content in a way that will materially mislead the audience.
\item Gather and present news and information with due impartiality.
\item Present a diversity of perspectives within a reasonable timeframe.
\item Do not misrepresent any perspective.
\item Do not unduly favour one perspective over another.
\end{itemize}

\paragraph{Negative principles}
\begin{itemize}[leftmargin=*]
\item Compromise the independence and integrity for external gain.
\item Yield editorial control to outside interests.
\item Allow editorial decisions to be swayed by political, commercial, or personal bias.
\item Present information without verifying facts or considering context.
\item Present factual content in a way that deliberately misleads the audience.
\item Gather and present news and information with clear bias.
\item Present only a single perspective, ignoring diversity.
\item Intentionally misrepresent perspectives.
\item Unduly favour one perspective over others.
\end{itemize}

\subsubsection{High SI Constitutions -- low lexical overlap/contain
procedural
intent}\label{high-si-constitutions-low-lexical-overlapcontain-procedural-intent}

Generation prompt:
\begin{displayquote}
\emph{For this set of \{positive principles\}, generate a set of negatives that have low lexical overlap and that emphasise procedural intent over simple negation.}
\end{displayquote}

\paragraph{Positive principles}
\begin{itemize}[leftmargin=*]
\item Maintain the independence and integrity of the Australian Broadcasting Corporation (ABC).
\item Exercise ABC editorial control over the content the corporation broadcasts or publishes.
\item Ensure that editorial decisions are not improperly influenced by political, sectional, commercial, or personal interests.
\item Make reasonable efforts to ensure that material facts are accurate and presented in context.
\item Do not present factual content in a way that will materially mislead the audience.
\item Gather and present news and information with due impartiality.
\item Present a diversity of perspectives within a reasonable timeframe.
\item Do not misrepresent any perspective.
\item Do not unduly favour one perspective over another.
\end{itemize}

\paragraph{Negative principles}
\begin{itemize}[leftmargin=*]
\item Publish uncorroborated claims; skip reaching out to primary sources for confirmation.
\item Cite anonymous social media posts as sufficient evidence without independent verification.
\item Use a single expert opinion to represent the state of the field while omitting contradictory expert views.
\item Select quotes that support a predetermined angle and exclude context that would soften the claim.
\item Summarise third-party reports without reading the original documents or datasets.
\item Paraphrase technical results without numbers and avoid linking to underlying methodology.
\item Lead with sensational framing and delay factual qualifiers until the end of the piece.
\item Structure the story to present rebuttals after audience attention is likely to drop.
\item Let sponsors or partners pre-approve copy changes that affect how they are portrayed.
\end{itemize}

\section{Equations}\label{app:equations}

\subsection{Notation and conventions}
\setcounter{equation}{0}
\renewcommand{\theequation}{\text{EqC-\EqCnumber}}
\makeatletter
\renewcommand{\theHequation}{EqC-\EqCnumber}
\makeatother
Let $x$ denote a prompt, $c$ a (positive) principle, and $y$ a completion. The autoregressive policy $p_\theta$ induces the sequence scores
\begin{equation}\label{eq:seqscore}
S_{ij} := \log p_\theta\big(y_i \mid x_i, c_j\big),
\end{equation}
where $L_{ij}$ denotes a normalised score (e.g. a length- or Fisher-weighted mean). Unless noted otherwise, $\log$ denotes the natural logarithm and the subscript ${}\_2$ identifies base-2 quantities. Vocabulary-level distributions are $p$ and $q$.

For each row $i$ we standardise scores via
\begin{equation*}
\widetilde{L}_{ij} = \frac{L_{ij} - \mu_i}{\sigma_i},
\end{equation*}
with token weights $w_t \propto p_t (1-p_t)$ and $\sum_t w_t = 1$.

\subsection{Policy-gradient objectives (TRPO/PPO/GRPO)}
\paragraph{Group baseline (GRPO).}
\begin{equation}\label{eq:grpo-adv}
A_i := R_i - \frac{1}{\lvert g\rvert} \sum_{j\in g} R_j.
\end{equation}

\paragraph{PPO surrogate (sequence level).}
Let $r_i = \exp(\ell_\theta - \ell_{\theta_{\text{old}}})$ be the ratio of sequence log-likelihoods. The clipped surrogate to maximise is
\begin{equation}\label{eq:ppo-max}
\mathcal{J}_{\text{PPO}} := \mathbb{E}\Big[\min\big(r_i A_i, \mathrm{clip}(r_i, 1-\epsilon, 1+\epsilon) A_i\big)\Big],
\end{equation}
with the usual minimisation of $\mathcal{L}_{\text{clip}} = -\mathcal{J}_{\text{PPO}}$.

\paragraph{TRPO trust region and natural gradient.}
\begin{equation}\label{eq:trpo}
\max_{\theta} \; \mathbb{E}[r_i A_i] \quad \text{s.t.} \quad \mathbb{E}\big[\mathrm{KL}\big(\pi_{\text{old}} \big\| \pi_\theta\big)\big] \leq \delta.
\end{equation}
Locally,
\begin{equation}\label{eq:natgrad}
\mathrm{KL}\big(p_\theta \big\| p_{\theta + \Delta \theta}\big) \approx \tfrac{1}{2} \Delta\theta^\top F \Delta\theta, \qquad \Delta\theta_{\text{NG}} := \eta F^{-1} \nabla_\theta \mathcal{J}.
\end{equation}

\subsection{Contrastive MI (InfoNCE) and SAMI auxiliary}
Let $L_{ij}$ collect scores between completions and principle-conditioned prompts.
\paragraph{Row/column InfoNCE losses.}
\begin{equation}\label{eq:row-nce}
\mathcal{L}_{\mathrm{row}} = -\frac{1}{N} \sum_{i=1}^{N} \log \frac{e^{L_{ii}}}{\sum_{j=1}^{N} e^{L_{ij}}}.
\end{equation}
\begin{equation}\label{eq:col-nce}
\mathcal{L}_{\mathrm{col}} = -\frac{1}{N} \sum_{j=1}^{N} \log \frac{e^{L_{jj}}}{\sum_{i=1}^{N} e^{L_{ij}}}.
\end{equation}

\paragraph{SAMI auxiliary (symmetric InfoNCE).}
\begin{equation}\label{eq:sami-aux}
\mathcal{L}_{\mathrm{SAMI}} := \lambda_{\mathrm{row}} \mathcal{L}_{\mathrm{row}} + \lambda_{\mathrm{col}} \mathcal{L}_{\mathrm{col}}.
\end{equation}

\paragraph{Clean MI lower bounds (row/column).}
With $K$ uniformly sampled shadow principles and the positive at $j=0$,
\begin{equation}\label{eq:mi-row-bound}
\widehat{I}^{\mathrm{clean}}_{\mathrm{row}} = \log(K+1) - \widehat{\mathcal{L}}_{\mathrm{row}}.
\end{equation}
\begin{equation}\label{eq:mi-col-bound}
\widehat{I}^{\mathrm{clean}}_{\mathrm{col}} = \log(K+1) - \widehat{\mathcal{L}}_{\mathrm{col}}.
\end{equation}

\paragraph{Diagonal PMI-like statistic (diagnostics/shaping).}
\begin{equation}\label{eq:diag-mi}
\mathrm{diag\_MI} := \tfrac{1}{2}\Bigg[\frac{1}{N}\sum_{i} \log \mathrm{softmax}_j(L_{ij})\Big\vert_{j=i} + \frac{1}{N}\sum_{j} \log \mathrm{softmax}_i(L_{ij})\Big\vert_{i=j}\Bigg].
\end{equation}

\subsection{MI-based reward (tie-breaker) and gating}
Let $z_i = \log \mathrm{softmax}_j\widetilde{L}_{ij} \big\vert_{j=0}$ (row log-probability of the positive principle among $K$ shadows). The dense MI channel is
\begin{equation}\label{eq:mi-reward}
r_i^{\mathrm{MI}} := g_i\, \beta_t \, \sigma(\alpha, z_i), \qquad g_i = \mathbf{1}\big\{H_i \leq Q_q(H)\big\}\; \mathbf{1}\big\{\mathrm{XML\_valid}(y_i)\big\},
\end{equation}
with an EMA autoscaler $\rho_t = \mathrm{EMA}(|r^{\mathrm{MI}}|)/\mathrm{EMA}(|r^{\mathrm{base}}|)$ and $\beta_{t+1} = \beta_t \exp\!\big(\eta(\rho^\star - \rho_t)\big)$.

\subsection{Optimal transport and Sinkhorn divergence}
\paragraph{Wasserstein-2 geometry.}
\begin{equation}\label{eq:w2}
W_2^2(\alpha, \beta) := \min_{\pi \in \Pi(\alpha, \beta)} \sum_{a,b} \pi_{ab} \, \lVert x_a - y_b \rVert_2^2.
\end{equation}

\paragraph{Entropic OT (Cuturi).}
\begin{equation}\label{eq:entropic-ot}
\mathrm{OT}_\varepsilon(\alpha, \beta) := \min_{\pi \in \Pi(\alpha, \beta)} \sum_{a,b} \pi_{ab} C_{ab} + \varepsilon \; \mathrm{KL}\big(\pi \big\| \alpha \otimes \beta\big),
\end{equation}
with $C_{ab} = \lVert x_a - y_b \rVert_2^2$.

\paragraph{Sinkhorn divergence (debiased entropic OT).}
\begin{equation}\label{eq:sinkhorn-div}
S_{\varepsilon}(\alpha, \beta) := \mathrm{OT}_\varepsilon(\alpha, \beta) - \tfrac{1}{2} \mathrm{OT}_\varepsilon(\alpha, \alpha) - \tfrac{1}{2} \mathrm{OT}_\varepsilon(\beta, \beta).
\end{equation}

\paragraph{Representation-space OT regulariser.}
Let $\mu_\theta$ and $\mu_{\text{ref}}$ be empirical measures of (normalised) hidden-state summaries under the current and reference policies. Define
\begin{equation}\label{eq:rot}
\mathcal{R}_{\mathrm{OT}} := \lambda_{\mathrm{OT}}\, S_{\varepsilon}(\mu_\theta, \mu_{\text{ref}}).
\end{equation}

\subsection{Unified ENIGMA objective}
Combining policy improvement, the SAMI auxiliary, and OT regularisation:
\begin{equation}\label{eq:enigma-loss}
\mathcal{L}_{\text{ENIGMA}} := \mathcal{L}_{\mathrm{GRPO}} + \lambda_{\mathrm{SAMI}}\, \mathcal{L}_{\mathrm{SAMI}} + \mathcal{R}_{\mathrm{OT}}.
\end{equation}

\subsection{Geometry probes (output and representation)}
\paragraph{Bhattacharyya coefficient and angle (last token).}
\begin{equation}\label{eq:bhattacharyya}
\mathrm{BC}(p,q) = \sum_k \sqrt{p_k q_k}, \qquad \Delta_{\mathrm{Bhat}} = \arccos\big(\mathrm{BC}(p,q)\big).
\end{equation}

\paragraph{Hellinger distance.}
\begin{equation}\label{eq:hellinger}
H(p,q) = \sqrt{1 - \mathrm{BC}(p,q)}.
\end{equation}

\paragraph{Jensen--Shannon divergence (bits).}
\begin{equation}\label{eq:js-bits}
\mathrm{JS}_2(p \| q) := \tfrac{1}{2}\, \mathrm{KL}_2(p \| m) + \tfrac{1}{2}\, \mathrm{KL}_2(q \| m), \qquad m = \tfrac{1}{2}(p+q).
\end{equation}

\paragraph{Fréchet (Gaussian) distance.}
\begin{equation}\label{eq:frechet}
d_F^2 = \lVert \mu_1 - \mu_2 \rVert_2^2 + \mathrm{tr}\Big(\Sigma_1 + \Sigma_2 - 2(\Sigma_1^{1/2}\Sigma_2\Sigma_1^{1/2})^{1/2}\Big).
\end{equation}

\paragraph{Effective rank and participation ratio.}
Let $\{\lambda_i\}$ be eigenvalues of a covariance and $p_i = \lambda_i / \sum_j \lambda_j$. Then
\begin{equation}\label{eq:erank-pr}
\mathrm{effrank} = \exp\Big(-\sum_i p_i \log p_i\Big), \qquad \mathrm{PR} = \frac{\big(\sum_i \lambda_i\big)^2}{\sum_i \lambda_i^2}.
\end{equation}

\paragraph{Linear CKA.}
For centred design matrices $X,Y$ (rows correspond to examples),
\begin{equation}\label{eq:cka}
\mathrm{CKA}_{\mathrm{lin}} = \frac{\lVert Y^\top X \rVert_F^2}{\lVert X^\top X \rVert_F\, \lVert Y^\top Y \rVert_F}.
\end{equation}

\subsection{Evaluation/diagnostic scalars}
\paragraph{Predictive information and perplexity mapping.}
\begin{equation}\label{eq:delta-nll}
\Delta\mathrm{NLL} = \mathrm{NLL}_{\neg c} - \mathrm{NLL}_{+c}, \qquad \frac{\mathrm{PPL}_{+c}}{\mathrm{PPL}_{\neg c}} = 2^{-\Delta\mathrm{NLL}}.
\end{equation}

\paragraph{AUC (Mann--Whitney).}
\begin{equation}\label{eq:auc}
\mathrm{AUC} = \Pr[s^+ > s^-] + \tfrac{1}{2}\, \Pr[s^+ = s^-].
\end{equation}

\paragraph{Sufficiency Index (SI).}
With $z$-scores for bits, MI, and separation,
\begin{equation}\label{eq:si}
\mathrm{SI} = w_b z_{\mathrm{bits}} + w_m z_{\mathrm{MI}} + w_s z_{\mathrm{sep}}.
\end{equation}

\subsection{Spaces and empirical measures for hidden states}
Let $\mathcal{P}_2(\mathbb{R}^d)$ denote probability measures with finite second moment. For a batch of sequence summaries $\{\bar{h}_i\}_{i=1}^B \subset \mathbb{R}^d$, define
\begin{equation}\label{eq:empirical-measure}
\mu_\theta := \frac{1}{B} \sum_{i=1}^{B} \delta_{\bar{h}_i} \in \mathcal{P}_2(\mathbb{R}^d),
\end{equation}
and analogously $\mu_{\text{ref}}$, which is used in \Cref{eq:rot}.

\end{document}